\documentclass[final,1p,times]{elsarticle}
\usepackage{graphicx}
\usepackage{dcolumn}
\usepackage{bm}
\usepackage{xcolor}
\usepackage{subcaption}
\usepackage{amsmath}
\usepackage{hyperref}
\usepackage{natbib} \setcitestyle{comma,sort&compress,numbers,square}
\hypersetup{hypertex=true,
            colorlinks=true,
            linkcolor=blue,
            anchorcolor=blue,
            citecolor=blue}
\usepackage{amsthm}
\usepackage{geometry}
\theoremstyle{definition}

\begin{document}
\begin{frontmatter}
\title{Fast Simulation of Particulate Suspensions Enabled by Graph Neural Network}
\date{}
\author[WISC]{Zhan Ma}
\author[WISC]{Zisheng Ye}
\author[WISC]{Wenxiao Pan\corref{cor}}
\ead{wpan9@wisc.edu}
\cortext[cor]{Corresponding author}
\address[WISC]{Department of Mechanical Engineering, University of Wisconsin-Madison, Madison, WI 53706, USA}

\begin{abstract}
Predicting the dynamic behaviors of particles in suspension subject to hydrodynamic interaction (HI) and external drive can be critical for many applications. By harvesting advanced deep learning techniques, the present work introduces a new framework, hydrodynamic interaction graph neural network (HIGNN), for inferring and predicting the particles' dynamics in Stokes suspensions. It overcomes the limitations of traditional approaches in computational efficiency, accuracy, and/or transferability. In particular, by uniting the data structure represented by a graph and the neural networks with learnable parameters, the HIGNN constructs surrogate modeling for the mobility tensor of particles which is the key to predicting the dynamics of particles subject to HI and external forces. To account for the many-body nature of HI, we generalize the state-of-the-art GNN by introducing higher-order connectivity into the graph and the corresponding convolutional operation. For training the HIGNN, we only need the data for a small number of particles in the domain of interest, and hence the training cost can be maintained low. Once constructed, the HIGNN permits fast predictions of the particles' velocities and is transferable to suspensions of different numbers/concentrations of particles in the same domain and to any external forcing. It has the ability to accurately capture both the long-range HI and short-range lubrication effects. We demonstrate the accuracy, efficiency, and transferability of the proposed HIGNN framework in a variety of systems. The requirement on computing resource is minimum: most simulations only require a desktop with one GPU; the simulations for a large suspension of 100,000 particles call for up to 6 GPUs.    
\end{abstract}


\end{frontmatter}

\section{Introduction}



Particulate suspensions are ubiquitous and important in a wide range of applications, including materials science, biological systems, soft robotics, advanced manufacturing, and food processing \cite{ParticleHI_PRL2019,DissipAssembUni_NP2020a,BiofilmGMLS_PanNC2021,Marchetti2013Active3,Banani2017Biomatter,ColloidsReview_COCIS2019,Yang2020Reconfigurable_microbots,Zohdi2017,KIM2020Multiphase}, to name a few. Due to the small sizes of particles, the Reynolds numbers for suspension flows are typically low and can fall within the Stokes limit. To elucidate how particles are spatio-temporally coupled and how external forces control the dynamic behaviors of particles, theoretical modeling and computer simulations can play a key role. However, simulating particulate Stokes suspensions is a challenging task, because the dynamics of particles is not only driven by external forcing, but also strongly influenced by hydrodynamic interaction (HI) that is at play in both short and long ranges \cite{Brady1988_SDreview,Wyss2006Difficulties2,ParticulateReview_Maxey2017,ColloidNS_NCM2019,Zohdi_CMAME2007}. As a result, the particles' trajectories can be highly correlated over long distances compared with the particle size, in the meanwhile affected by the lubrication effects over short distances when in near contact with other particles. In addition, HI consists of many-body contributions, and hence cannot be simply modeled as a pairwise interaction \cite{Brady1988_SDreview,ParticulateReview_Maxey2017,Zohdi_CMAME2007}. 

In literature, there are mainly two categories of approaches for modeling and simulating particulate suspensions. In the first category, suspension flows are regarded as a fluid-solid interaction problem, i.e., solids freely moving in a flow subject to bidirectional hydrodynamic couplings. Thus, one can solve the Stokes equations with no-slip boundary conditions imposed on the particles’ surfaces. To this end, a numerical solver is required for solving the associated boundary-valued partial differential equations (PDEs). Along this line, some representative numerical methods (not a thorough list) include the immersed boundary method \cite{IBM_CMAME2007,IBM_CMAME2016}, boundary integral method \cite{ParticulateBI_JCP2020}, force-coupling method \cite{YEO2010FCM3}, and generalized moving least squares method \cite{GMLS_StokesColloid_HuCMAME2019}. All these methods require to handle numerical discretization and to solve large-scale linear systems at each time step as particles move and the configuration of particles evolves with time. As a result, the simulations can be computationally demanding and prohibitive for very large scale applications. To relieve the computation expense to some extent, approximations are introduced, leading to the approaches of the second category. The matrix that relates the particles' velocities to external forces is called the mobility tensor. The approximation to it by the Rotne-Prager-Yamakawa tensor \cite{RPY_JFM2014} is valid for dilute suspensions. A more accurate approximation is by the Stokesian dynamics (SD) \cite{Brady1988_SDreview}, which splits HI into the far-field and near-field parts. SD firstly constructs the mobility matrix by pair-wisely adding the far-field HI through multipole expansions truncated at the stresslet level, and the many-body nature of HI is implicitly captured for the far field by inverting the mobility matrix \cite{Durlofsky1987SD}. The near-field lubrication effects are then added to the obtained far-field resistance matrix as a pairwise interaction between any pair of particles through asymptotic lubrication resistance functions \cite{ARP1977exact_resistance1,Jeffrey1984exact_resistance2, Kim1985exact_resistance3,Jeffrey1992exact_resistance4,Kim1991exact_resistance5}. 
By inverting the resultant resistance matrix the velocities of particles can be obtained. Thus, SD is accurate up to the stresslet level for the far-field HI and up to the two-body contributions for the near-field lubrication effects, beyond which it is difficult to further improve the accuracy of SD. By avoiding to solve PDEs, SD can be computationally more efficient. However, it still requires to compute the inverse of matrices at each time step. For suspensions of a large number of particles, inversion of each large, dense matrix can still be expensive and in turn slow down the simulations. Thus, the accelerated SD \cite{sierou2001ASD,WANG2016ASD2,Ouaknin2021ParaASD} and fast SD \cite{Fiore2019FSD} were developed to address this issue. By introducing Ewald sum splitting \cite{Ewald1921, hasimoto_1959_ewald}, the many-body far-field interaction is split into a real space contribution (with near-neighbor interactions only) and a wave space contribution, which in turn allows for using fast Fourier transform \cite{Cooley1965FFT} to accelerate the inversion of matrix.
 
In recent years, the rapid development of machine learning techniques has provided us new perspectives on modeling and simulating physical problems. In particular, graph neural networks (GNNs) \cite{Scarselli2009GNN,Gilmer2017MPNN, battaglia2018relational}, a type of deep neural networks directly operating on graphs, have been applied to learning and inferring dynamics that involves interactions between particles \cite{Battaglia2016Interaction,Li2019Propagation,Sanchez-Gonzalez2020GNS,Bapst2020GNN}. In those efforts, particles and pairwise (two-body) interactions are modeled through the vertices and edges of a graph. To involve many-body interactions beyond pairwise, multi-step message passing was introduced in GNN, which generates a sequence of latent graphs with the updated features of vertices. {At each step of message-passing, GNN first computes the message of every vertex, which is a function of its own feature vector with learnable parameters. The feature vector is then updated for every vertex by aggregating the message from each of its neighbors, and the updated feature can be used to calculate the messages in the next step of message passing. By such,} this multi-step passing ensures the propagation of information between more than two particles and hence implicitly accounts for many-body interactions. One salient feature of GNN for modeling interacting particles is that a well trained GNN can be applied to inferring the dynamics of arbitrary numbers of particles.
However, for achieving that, the previous GNN models require the interactions between particles within a cutoff distance \cite{Sanchez-Gonzalez2020GNS}, which means all long-range interactions beyond the cutoff distance cannot be considered. 
The reason for this limitation lies in the fact that ignoring long-range messages at each passing step can keep the updated feature vectors from being out of distribution when making predictions for any number of particles. As a result, one can avoid extrapolation at each message passing step and thereby apply the trained GNN model to arbitrary numbers of particles; otherwise, the trained GNN model is not transferable to systems with more particles than those used in training. In regard to training cost, even the long-range interactions can be ignored, one must include a sufficiently large number of particles for training to ensure the training data thoroughly distributed over the entire range of the updated feature vectors in each passing step, which in turn significantly increases the cost of generating data and training the GNN. Due to the high training cost and the incapability to account for long-range interactions, the previously reported strategies and methodologies utilizing GNNs \cite{Battaglia2016Interaction,Li2019Propagation,Sanchez-Gonzalez2020GNS,Bapst2020GNN}, although showing a promising potential, cannot be applied to modeling particulate suspensions, where the long-range HI cannot be completely neglected, and generating adequate training data with a large number of particles is intractable.

Therefore, the present work introduces a new GNN framework, named as hydrodynamic interaction graph neural network (HIGNN), to enable fast simulations of particulate suspensions. The many-body HI is decomposed into different $m$-body contributions, e.g., single-body, two-body, three-body, etc. For each $m$-body contribution, we seek a surrogate model as a neural network with learnable parameters. The HIGNN is built on the edge convolutional GNN \cite{wang2019edge_conv} but generalizes it to include higher-order connectivity in graph and the corresponding convolutional operation to account for many-body interactions. For example, a face connectivity between three vertices is introduced into the graph and used to describe the three-body connections and thereby to capture the three-body HI effects. Note that this strategy is fundamentally different from the previous work \cite{Sanchez-Gonzalez2020GNS} that relies on multi-step message passing to involve many-body interactions, and hence is able to account for long-range interactions. Furthermore, if the desired accuracy requires to retain up to $m$-body contributions in HI, e.g., $m=3$, the training would need the data for three particles sampled with different spatial configurations in the domain. Thus, the training cost can be maintained very low. Moreover, since how each $m$-body contributions in HI depend on the associated particles’ relative positions is invariant regardless of the number of particles in the suspension and external forcing, the trained HIGNN can be applied to simulating suspensions with different numbers of particles in the same domain and is transferable across different external forces. Compared with the traditional approaches such as SD or solving the Stokes equations, applying the trained HIGNN for inferring and simulating particles' dynamics can be computationally more efficient and does not need additional efforts in developing linear solvers, preconditioners, and/or parallel implementation. In addition, as long as the training data include the sample configurations of $m$ particles in near contact, the HIGNN can accurately capture the near-field $m$-body lubrication effects. Finally, the accuracy of the HIGNN framework is controllable by {truncation of higher-order force moments, i.e., retaining up to different $m$-body contributions in HI.} 

We demonstrate the accuracy, transferability, and efficiency attributes of the HIGNN framework for predicting the particles’ dynamics in a variety of suspensions and problem setups. 
As an additional benefit, the present work also paves a way to advance deep learning techniques in regard to GNNs by introducing higher-order connectivity in graph and the corresponding convolutional operation.

The rest of the paper is organized as follows. In \S\ref{sec:method}, we explain the proposed methodology, provide theoretical analysis, describe the main components of the HIGNN framework, and lay out the training procedure. Some detailed derivations of equations are provided in \ref{sec:append}. In \S\ref{sec:results}, we present our results about the training of the HIGNN and applying the trained HIGNN for simulating different particulate suspensions, including sedimenting particles, self-assembling particles, suspensions in a periodic or unbounded domain, and suspensions with small to moderate to large numbers of particles. We examine the accuracy of the HIGNN’s predictions by comparison with the ground truth. The computer times spent for training the HIGNN and for applying the trained HIGNN in simulations are evaluated to assess the training cost and computational efficiency of the HIGNN. We finally conclude in \S\ref{sec:conclusion} with a summary of our contributions and discuss about the potential future extensions of the HIGNN following this work.

\section{Methodology} \label{sec:method}

\subsection{Problem Statement}
Consider a system of $N$ rigid spherical particles of radius $a$ suspended in an incompressible Newtonian fluid of density $\rho$ and viscosity $\mu$. Denote the subdomain occupied by the fluid as $\Omega_f$, the subdomain occupied by the particles as $\Omega_i$, and the boundary of each particle as $\Gamma_i$ with $i = 1,2,\dots,N$. Assuming the Reynolds number $Re=\rho Ua/\mu \ll 1$ with $U$ the characteristic velocity of particles,
the particles' motion is governed by the over-damped dynamics that neglects the inertia effect, as given by:
\begin{equation}\label{equ:BD_mobility}
\begin{bmatrix}
    \dot{\textbf{X}} \\
   \dot{\boldsymbol{\theta}}
\end{bmatrix}=
\begin{bmatrix}
    \mathbf{U} \\
    \boldsymbol{\omega}
\end{bmatrix}
= \textbf{M} (\textbf{X})\cdot 
\begin{bmatrix}
    \textbf{F} \\
    \textbf{T}
\end{bmatrix}\;,
\end{equation}
where the vectors $\textbf{X} \in \mathbb{R}^{3N}$ and $\boldsymbol{\theta} \in \mathbb{R}^{3N}$ represent the spatial positions and orientations, respectively, of all particles; $\mathbf{U}\in \mathbb{R}^{3N}$ and $\boldsymbol{\omega}\in \mathbb{R}^{3N}$ denote the translational and rotational velocities, respectively; $\textbf{M}\in\mathbb{R}^{6N\times 6N}$ is the mobility tensor; $\textbf{F} \in \mathbb{R}^{3N}$ and $\textbf{T} \in \mathbb{R}^{3N}$ includes all external forces and torques applied on each particle and can be of various origins, such as gravity, external fields, and interparticle potentials. Each entity $\mathbf{M}_{ij} \in \mathbb{R}^{3 \times 3} $ in the mobility tensor $\textbf{M}$ corresponds to the translational or rotational velocity of particle $i$ induced per unit force or torque exerted on particle $j$. Since each particle's mobility is affected by the presence of other particles due to HI, $\textbf{M}$ must account for many-body contributions and depends on the configuration $\textbf{X}$ of all particles in suspension. The key to solving Eq. \eqref{equ:BD_mobility} in simulations is to obtain $\textbf{M}$ at each instant time. 

Due to the bidirectional hydrodynamic couplings between the particle kinematics and Stokes flow, one way to obtain $\textbf{M} (\textbf{X})$ is to directly solve the Stokes equations subject to prescribed boundary conditions on the particles’ surfaces  \cite{ParticulateReview_Maxey2017,YEO2010FCM3,GMLS_StokesColloid_HuCMAME2019,GMLS_ScalableSolver_Ye2022}: 
\begin{equation}
    \left\{
    \begin{aligned}
         & -\nabla p + \mu \nabla^2 \mathbf{u} = 0                                                     & \quad & \forall \mathbf{x} \in \Omega_f                    \\
         & \nabla \cdot \mathbf{u} = \mathbf{0}                                                                                       & \quad & \forall \mathbf{x} \in \Omega_f                    \\
         & \mathbf{u}  = \mathbf{U}_i + \boldsymbol{\omega}_i \times (\mathbf{x} - \mathbf{X}_i)                   & \quad & \forall \mathbf{x} \in \Gamma_i, ~ i = 1, 2, \dots, N \\
         & -\mathbf{n} \cdot \nabla p + \mu \mathbf{n} \cdot \nabla ^2 \mathbf{u} = 0 & \quad & \forall \mathbf{x} \in \Gamma_1 \cup \Gamma_2 \dots \cup \Gamma_{N}  \\
         & \mathbf{F}_{i} + \int_{\Gamma_i} \boldsymbol \sigma \cdot \mathbf{n} dS = \mathbf{0}  & \quad &  i = 1,2, \dots, N \\
         & \mathbf{T}_{i} + \int_{\Gamma_i}  (\mathbf{x} - \mathbf{X_i}) \times (\boldsymbol{\sigma} \cdot \mathbf{n}) dS = \mathbf{0}  & \quad &  i = 1,2, \dots, N 
    \end{aligned}
    \right.
    \label{eq:governing_eq}
\end{equation}
where $\mathbf{u}$ and $p$ denote the fluid velocity and pressure, respectively; $\mu$ is the viscosity of fluid; $\mathbf{n}$ is the unit normal vector outward facing at boundary $\Gamma_i$; $\boldsymbol{\sigma} = -p\mathbf{I} + \mu [\nabla \mathbf{u} + (\nabla \mathbf{u})^T]$ is the stress exerted by the fluid on the particles. Alternatively, following SD and its variants \cite{Brady1988_SDreview,sierou2001ASD,Ouaknin2021ParaASD,Fiore2019FSD}, the mobility tensor can be approximately obtained through multipole expansions truncated at the stresslet level. Either way can be computationally demanding and requires significant efforts in handling near-field lubrication effects \cite{Durlofsky1987SD,YEO2010FCM3,GMLS_StokesColloid_HuCMAME2019}, accelerating inversion of dense matrices \cite{Ouaknin2021ParaASD,Fiore2019FSD}, designing scalable iterative linear solvers \cite{GMLS_ScalableSolver_Ye2022}, and/or implementing parallel computing \cite{ParticulateBI_JCP2020,Ouaknin2021ParaASD,GMLS_ScalableSolver_Ye2022}. In addition, following either way, the computed $\textbf{M}$ cannot be transferred to other suspensions of the same type of particles but different numbers or concentrations. 

In the present work, we introduce a paradigm-shifting new way to compute for $\textbf{M}$ utilizing advanced GNN techniques, which permits fast evaluation of $\textbf{M}$ in simulations and also enables transferability across systems with different numbers of particles in the same domain. We next provide theoretical analysis and technical details about the proposed methodology.

\subsection{Theoretical Analysis}


From the many-body nature of HI, the mobility tensor $\textbf{M}$ may be decomposed as single-body and different many-body contributions, e.g., two-body, three-body, etc. Starting from the single-body contribution, i.e., HI between particles is neglected, $\textbf{M}$ reduces to a diagonal matrix, for example, in unbounded domain $\textbf{M}=(6\pi \mu a)^{-1} \mathbf{I}$. Accounting for HI, the velocity of a particle $i$ is perturbed by the presence of another particle $j$, the so-called two-body interaction. Thus, the diagonal $\textbf{M}$ needs to be corrected by considering the two-body HI effects that can be divided into: $\boldsymbol{\alpha}^{(s)}_2$ and $\boldsymbol{\alpha}^{(t)}_2$, where $\boldsymbol{\alpha}^{(s)}_2(\mathbf{X}_i,\mathbf{X}_j)$ corresponds to the contribution of two-body HI to $\mathbf{M}_{ii}$ due to the presence of particle $j$, and $\boldsymbol{\alpha}^{(t)}_2(\mathbf{X}_i,\mathbf{X}_j)$ represents the contribution of two-body HI to $\mathbf{M}_{ij}$ ($i\neq j$). Note that both $\boldsymbol{\alpha}^{(s)}_2$ and $\boldsymbol{\alpha}^{(t)}_2$ depend on only the positions of particles $i$ and $j$ and hence are functions of $\mathbf{X}_i$ and $\mathbf{X}_j$. In the presence of the third particle $k$, the three-body HI also contributes to  $\mathbf{M}_{ii}$ and $\mathbf{M}_{ij}$ ($i\neq j$), represented by $\boldsymbol{\alpha}^{(s)}_3(\mathbf{X}_i, \mathbf{X}_k, \mathbf{X}_j)$ and $\boldsymbol{\alpha}^{(t)}_3(\mathbf{X}_i, \mathbf{X}_k, \mathbf{X}_j)$, respectively, depending on the positions of particles $i$, $j$, and $k$. 
Along this line, $\boldsymbol{\alpha}^{(s)}_m$ and $\boldsymbol{\alpha}^{(t)}_m$ can be considered as the correction terms required for $\mathbf{M}_{ii}$ and $\mathbf{M}_{ij}$ ($i\neq j$), respectively, to account for the $m$-body HI. 


Therefore, $\textbf{M} (\textbf{X})$ can be expressed as an expansion of additive terms arising from different $m$-body contributions. As $m$ increases, the $m$-body HI decays at a faster rate with respect to the characteristic distance $r$ between particles. In particular, for spherical particles in unbounded domain, the leading orders of $\boldsymbol{\alpha}^{(t)}_2$, $\boldsymbol{\alpha}^{(s)}_2$, $\boldsymbol{\alpha}^{(t)}_3$, and $\boldsymbol{\alpha}^{(s)}_3$ are $O(r^{-1}) $, $O(r^{-4}) $, $O(r^{-4}) $, and $O(r^{-7}) $, respectively; the dominant orders of the contributions of four-body and above are  $ O(r^{-10})$ to $\mathbf{M}_{ii}$ and $O(r^{-7})$ to $\mathbf{M}_{ij}$ ($i\neq j$) ~\cite{Mazur1982ManysphereHI}. 

From the above analysis, we can make systematic approximations by truncation. For instance, we retain up to three-body contributions, i.e., both far-field HI and near-field lubrication effect are truncated to three-body, 
by 1) considering the fact that the contributions of four-body and above to the far-field HI are much smaller than the two-body and three-body contributions and 2) noting that it is of small probability that a particle is very closely surrounded by three or more other particles simultaneously.
Further, we can set a cutoff distance for $\boldsymbol{\alpha}^{(s)}_2$, $\boldsymbol{\alpha}^{(t)}_3$, and $\boldsymbol{\alpha}^{(s)}_3$ since they decay much faster than $\boldsymbol{\alpha}^{(t)}_2$. If higher-order accuracy is desired, we can retain more $m$-body contributions, with cutoff distances set for $\boldsymbol{\alpha}^{(s)}_2$ and the contributions of three-body and above (but not for $\boldsymbol{\alpha}^{(t)}_2$ due to its slow decay with respect to $r$). 

With the above-discussed approximation made for the mobility tensor, we can derive (details in \ref{sec:append}):
\begin{equation}\label{Eq:theo_basis_trun_final}
\begin{split}
    \mathbf{U}_i  & \approx  \boldsymbol{\alpha}_1 (\mathbf{X}_i) \cdot \mathbf{F}_i + \sum\limits_{\substack{{j = 1} \\ j\neq i}}^N \boldsymbol{\alpha}_2 (\mathbf{X}_i, \mathbf{X}_j)  \cdot \mathbf{F}_{i,j}  + \sum\limits_{\substack{ j, k: j \neq i,k \neq i \\i,j \in \mathcal{N}(k) }}^N \boldsymbol{\alpha}_3 (\mathbf{X}_i, \mathbf{X}_k, \mathbf{X}_j ) \cdot \mathbf{F}_{i,j} \;,
\end{split}
\end{equation}
where $\boldsymbol{\alpha}_1(\mathbf{X}_i)$ denotes the general single-particle mobility in the absence of HI between particles ($\boldsymbol{\alpha}_1(\mathbf{X}_i)=(6\pi \mu a)^{-1} \mathbf{I}$ for spherical particles in unbounded domain); $\boldsymbol{\alpha}_2 = [ \boldsymbol{\alpha}_2^{(s)},  \boldsymbol{\alpha}_2^{(t)}] \in \mathbb{R}^{3 \times 6}$ and $\boldsymbol{\alpha}_2^{(s)} (\mathbf{X}_i, \mathbf{X}_j) =\mathbf{0}$ for $j \not \in \mathcal{N}(i)$; $\boldsymbol{\alpha}_3 = [ \boldsymbol{\alpha}_3^{(s)},  \boldsymbol{\alpha}_3^{(t)}] \in \mathbb{R}^{3 \times 6} $; $\mathbf{F}_{i,j} = [\mathbf{F}_{i}^T , \mathbf{F}_{j}^T ]^T \in \mathbb{R}^{6 \times 1}$; and $\mathcal{N}(*)$ denotes the neighbors of particle $*$ within the cutoff distance $R_\text{cut}$. It is worth noting that Eq. \eqref{Eq:theo_basis_trun_final} can be applied to \textit{general} scenarios where the particulate suspensions are in \textit{unbounded}, \textit{periodic}, or \textit{bounded} domains. For different domains, the functions for $\boldsymbol{\alpha}_1$, $\boldsymbol{\alpha}_2$, and $\boldsymbol{\alpha}_3$ would be different.

We limit our discussion herein to translational motions of spherical particles without torques. However, the proposed methodology can be readily extended to include rotational velocities and torques.

\subsection{Hydrodynamic Interaction Graph Neural Network}

The next task is to construct each $\boldsymbol{\alpha}_m$ retained in the approximation (as in Eq. \eqref{Eq:theo_basis_trun_final}), which depends on the positions of associated $m$ particles and is essentially a nonlinear mapping from $m$-body connection to its contribution in the target particle's mobility. For that, we need to address the following questions: \textit{i}) how to characterize each $m$-body connection; \textit{ii}) how to model each $\boldsymbol{\alpha}_m$; and \textit{iii}) how to ensure the constructed $\boldsymbol{\alpha}_m$ transferable across suspensions of different numbers of particles. The key idea of our method is to use a graph data structure to characterize many-body connections and to build the nonlinear mapping $\boldsymbol{\alpha}_m$ as a neural network with learnable parameters. The graph data structure and nonlinear mappings can be united in a GNN. We hence introduce a novel GNN architecture, named as Hydrodynamic Interaction Graph Neural Network (HIGNN).

\subsubsection{Framework}\label{subsubsec:HIGNN_framework}
Considering a suspension of $N$ particles, we build a graph of $N$ vertices, denoted as $\boldsymbol{\mathcal{V}} = [1,2,\dots,N]^T$, with each particle regarded as a vertex. Each vertex is associated with a \textit{feature vector} whose elements are $\mathbf{X}_i$. The \textit{output vector} of each vertex is the particle's velocity $\mathbf{U}_i$. A \textit{directed} edge between two vertices, from the source vertex to the target vertex, defines a two-body connection. The total number of edges is $N_e = N(N-1)$ to encompass all two-body connections and their HIs. The $N_e$ edges are denoted by $\boldsymbol{\mathcal{E}} \in \mathbb{Z}^{2 \times N_e}$ with the two entities of each column corresponding to the indices of the target and source vertices of a directed edge. The \textit{edge feature} is defined by the forces on the target and source vertices, i.e., $\mathbf{F}_{i,j}$ in Eq.~\eqref{Eq:theo_basis_trun_final}. 

\begin{figure}[htbp]
\centering
\includegraphics[width=0.5\linewidth]{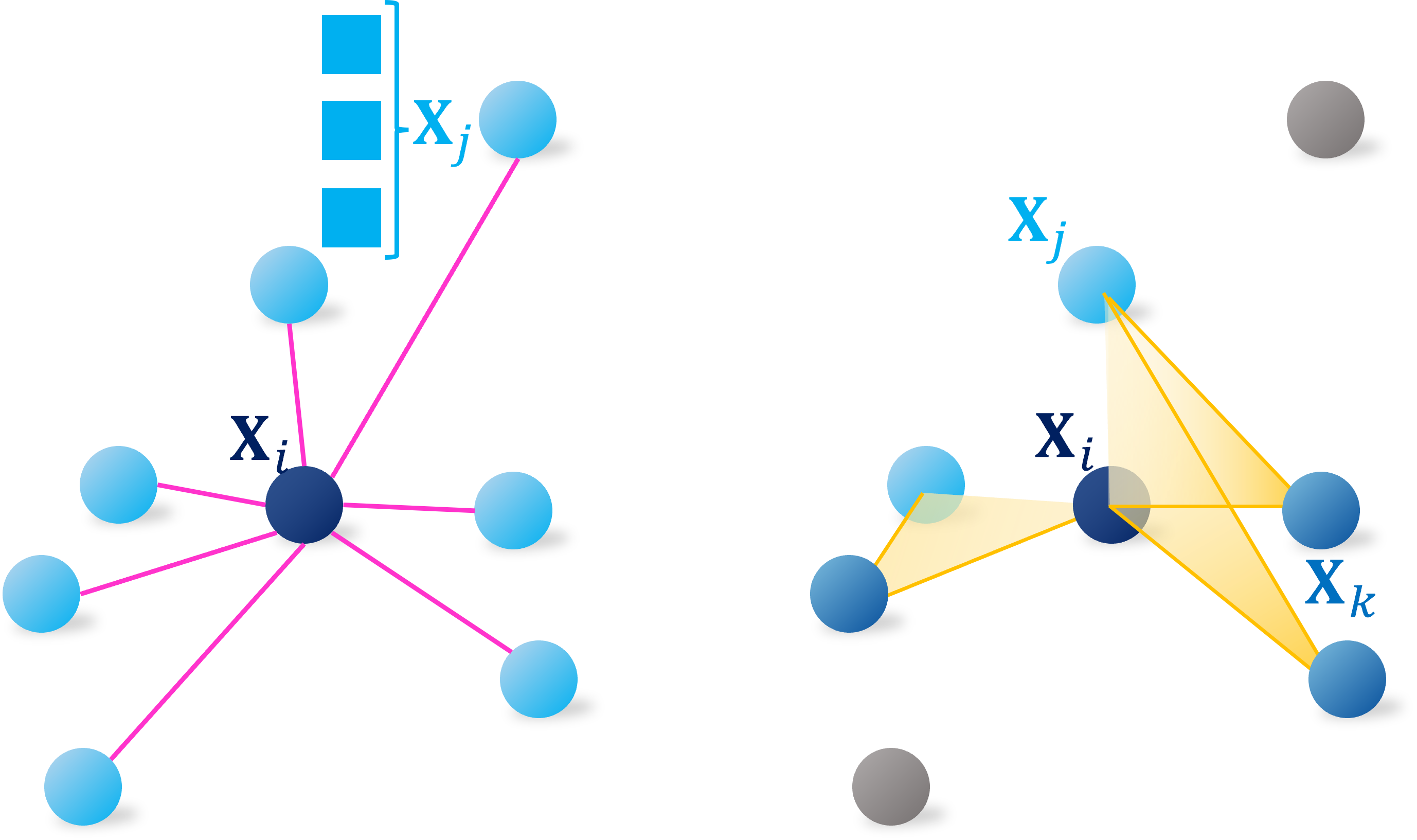}
\caption{A graph consisting of vertices, edges, and faces. Here, only the edges emanating from the target vertex $i$ are displayed; the faces built are associated with the target vertex $i$ and satisfy $i, j \in \mathcal{N}(k)$.}
\label{fig:graph_edge_face}
\end{figure}

The state-of-the-art GNN methods such as the edge convolutional GNN \cite{Scarselli2009GNN,wang2019edge_conv} only consider simple graphs consisting of vertices and edges. In this work, we go beyond and generalize the edge convolutional GNN to deal with more complex graphs that include higher-order connectivity in addition to vertices and edges. By introducing ``faces" into the graph, each comprising three vertices, we can define three-body connections. Given three vertices $(i,k,j)$, we define a \textit{directed} face for describing three-body HI, by which the velocity of the target vertex $i$ induced by the force exerted on the source vertex $j$ is affected by the presence of the third, named as ``passing vertex" $k$. A face is built only when the three vertices are within the preset $R_\text{cut}$, i.e., face $(i,k,j)$ exists if $i, j \in \mathcal{N}(k)$. The $N_f$ such defined faces are denoted by $\boldsymbol{\mathcal{F}} \in \mathbb{Z}^{3 \times N_f}$ with the entities in each column the indices of the three vertices forming a face. The \textit{face feature} is also defined by the forces on the target and source vertices, i.e., $\mathbf{F}_{i,j}$ in Eq.~\eqref{Eq:theo_basis_trun_final}. FIG. \ref{fig:graph_edge_face} depicts a graph consisting of vertices, edges, and faces, denoted as: $ \boldsymbol{\mathcal{G} = (\mathcal{V}, \mathcal{E}, \mathcal{F})}$. By further introducing ``tetrahedra" into the graph, each comprising four vertices, we can define four-body connections and thereby include the contributions of four-body HI in the mobility; and so on so forth, as needed for achieving desired accuracy.

Without loss of generality, assuming that the approximation as in Eq. \eqref{Eq:theo_basis_trun_final} satisfies the desired accuracy, any configuration $\mathbf{X}$ of $N$ particles is represented as a graph with edges and faces characterizing the two-body and three-body connections, respectively. Taking the graph as the input, GNN constructs $\boldsymbol{\alpha}_2$ and $\boldsymbol{\alpha}_3$ as the nonlinear mappings $\mathbf{h}_{\boldsymbol{\Theta}_2}$ and $ \mathbf{g}_{\boldsymbol{\Theta}_3}$ on those connections, respectively, with $\boldsymbol{\Theta}_2$ and $\boldsymbol{\Theta}_3$ the network parameters. In this work, we adopt the multilayer perceptron (MLP) neural networks for building the nonlinear mappings. The resulting surrogate modeling for the velocity of a particle can be expressed as:
\begin{equation}\label{Eq:U_surrogate_3body}
\begin{split}
    \mathbf{U}_i^{\text{HIGNN}}  & = \boldsymbol{\alpha}_1 (\mathbf{X}_i) \cdot \mathbf{F}_i + \sum\limits_{\substack{j \\(i,j)\in \boldsymbol{\mathcal{E}}}}
    \mathbf{h}_{\boldsymbol{\Theta}_2} (\mathbf{X}_i, \mathbf{X}_j) \cdot \mathbf{F}_{i,j}  +
    \sum \limits_{\substack{j,k \\(j,k,i) \in \boldsymbol{\mathcal{F}}}}
    \mathbf{g}_{\boldsymbol{\Theta}_3} (\mathbf{X}_i, \mathbf{X}_k, \mathbf{X}_j) \cdot \mathbf{F}_{i,j} \;,
\end{split}
\end{equation}
where $\mathbf{h}_{\boldsymbol{\Theta}_2}$ and $ \mathbf{g}_{\boldsymbol{\Theta}_3}$ are the surrogate models in the form of MLP neural networks for $\boldsymbol{\alpha}_2$ and $\boldsymbol{\alpha}_3$ in Eq. \eqref{Eq:theo_basis_trun_final}, respectively.

\begin{figure}
\centering
\begin{subfigure}{1.0\textwidth}
\centering
\includegraphics[width=\textwidth]{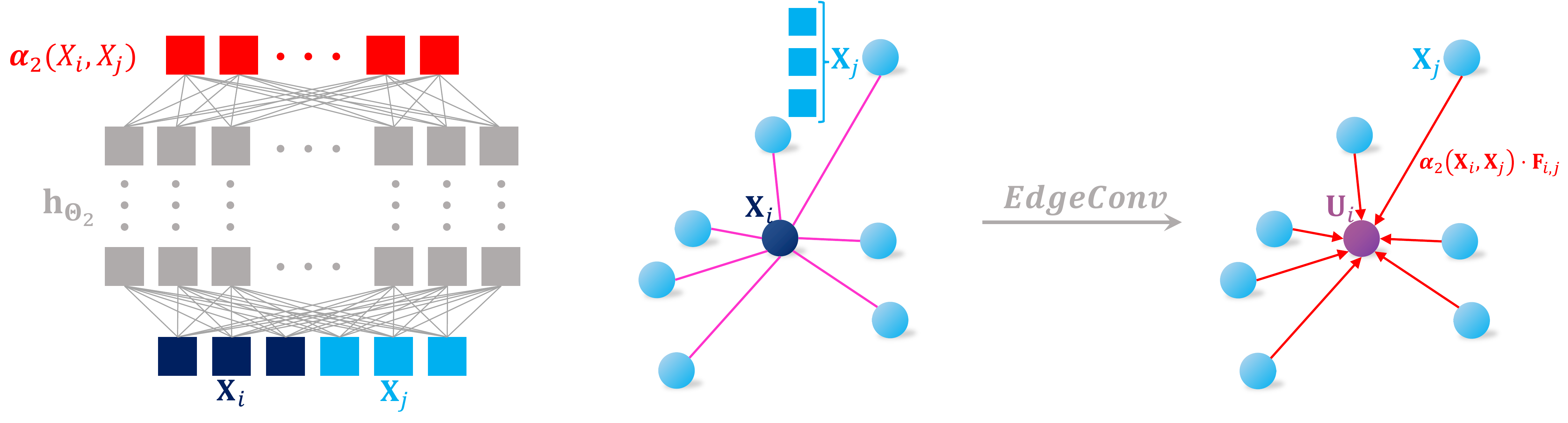}
\caption{The surrogate model of $\boldsymbol{\alpha}_2(\mathbf{X}_j, \mathbf{X}_i)$ is the MLP neural network $\mathbf{h}_{\boldsymbol{\Theta}_2}$ obtained through EdgeConv.} 
\label{sfig:edge_conv}
\end{subfigure}

\vspace{0.8cm}
\begin{subfigure}{1.0\textwidth}
\centering
\includegraphics[width=\textwidth]{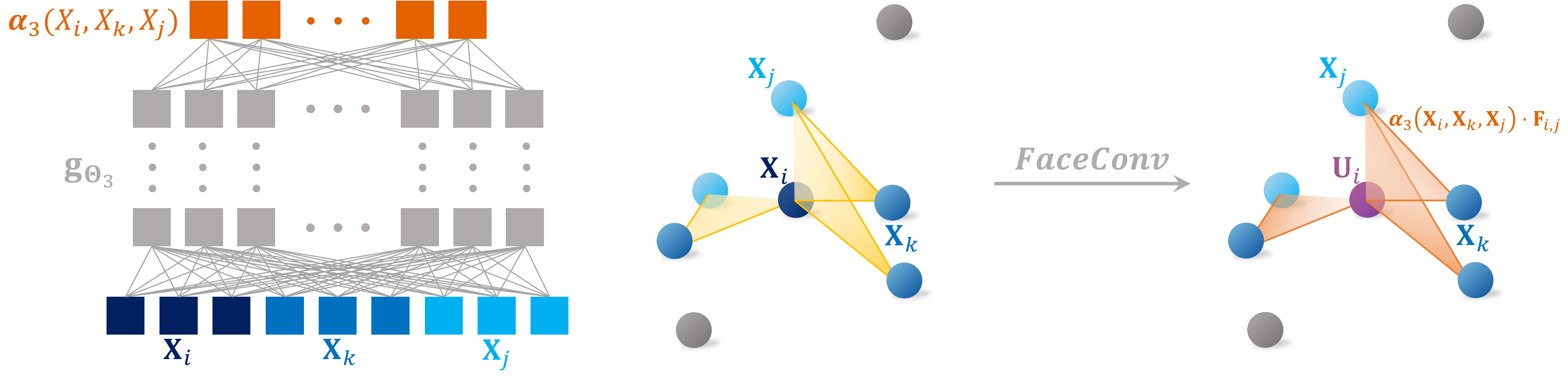}
\caption{The surrogate model of $\boldsymbol{\alpha}_3(\mathbf{X}_j, \mathbf{X}_k, \mathbf{X}_i)$ is the MLP neural network  $\mathbf{g}_{\boldsymbol{\Theta}_3}$ obtained through FaceConv.} 
\label{sfig:face_conv}
\end{subfigure}
\caption{Main components of the HIGNN framework.} 
\label{fig:graph_conv}
\end{figure}

To obtain $\mathbf{h}_{\boldsymbol{\Theta}_2}$, the edge convolutional operation (EdgeConv) \cite{wang2019edge_conv} aggregates the edge information associated with all edges pointing to each target vertex, i.e., $\mathbf{o}^{\text{EdgeConv}}_i = \sum\limits_{j: (j,i) \in \mathcal{E}}  \mathbf{e}_{ji} $ with $\mathbf{e}_{ji} = \boldsymbol{\alpha}_{2} (\mathbf{X}_i, \mathbf{X}_j) \cdot \mathbf{F}_{i,j}$. In analogy to edge convolution, we introduce a new \textit{face convolution} (FaceConv) in GNN to aggregate the information of faces associated with the same target vertex as: $\mathbf{o}^{\text{FaceConv}}_i = \sum\limits_{j, k: (j, k, i) \in \boldsymbol{\mathcal{F}}}  \mathbf{f}_{jki} $, where the face information $\mathbf{f}_{jki} = \boldsymbol{\alpha}_3 (\mathbf{X}_i, \mathbf{X}_k, \mathbf{X}_j ) \cdot \mathbf{F}_{i,j} $. The main components of the HIGNN framework are summarized in FIG. \ref{fig:graph_conv}.

For any unbounded or periodic domain, $\boldsymbol{\alpha}_1$ has its analytical form. For a bounded domain, we also need to construct the surrogate model of $\boldsymbol{\alpha}_1 (\mathbf{X}_i)$, which can be another nonlinear mapping obtained by the vertex convolutional operation in GNN.

\subsubsection{Training of HIGNN}\label{subsubsec:HIGNN_training}

To obtain the HIGNN-based surrogate modeling as in Eq. \eqref{Eq:U_surrogate_3body}, the training requires only the data for \emph{three particles} sampled with different spatial configurations in the domain of interest. In general, if the desired accuracy requires to retain up to the contributions of $m$-body HI, the training would need the data for $m$ particles sampled with different spatial configurations in the domain. For each configuration, we can randomly choose one particle to apply a \textit{unit} force in $x$, $y$, or $z$ direction. Each training data can be generated by solving only a small problem involving $m$ particles, for one step (no temporal integration is needed since we only need the data of $\textbf{U}$ at each configuration of $m$ particles). The solution of each such small problem can be readily obtained, e.g., by numerically solving the boundary-valued Stokes equations. If we need $N_\text{train}$ data, we just solve $N_\text{train}$ such small problems. All $N_\text{train}$ data can be produced simultaneously in a \emph{high-throughput} fashion, which is efficient and does not require any parallel implementation. The loss function used for training the HIGNN is: 
\begin{equation}\label{Eq:Loss}
    \mathcal{L} = \frac{1}{N_\text{train}} \left. \sum\limits_{i=1}^{N_\text{train}} \left( \|  \mathbf{U}_i - \mathbf{U}^\text{HIGNN}_i \|_2^2  \middle /  \| \mathbf{U}_i \|_2^2\right) \right.\;.
\end{equation}
Here, choosing the \textit{relative} mean-squared-error loss with $L_2$ norm of training data in the denominator is to ensure consistent accuracy regardless of velocity magnitudes.

\subsubsection{Transferability of HIGNN}\label{subsubsec:HIGNN_transferable}
Note that the surrogate models for $\boldsymbol{\alpha}_1$, $\boldsymbol{\alpha}_2$, and $\boldsymbol{\alpha}_3$ in HIGNN are invariant regardless of the number of particles in suspension. Therefore, the trained HIGNN can be applied to suspensions of any concentration or number of particles in the same domain. Furthermore, the surrogate modeling for the mobility tensor $\textbf{M}$ is independent on external forcing exerted on the particles. Thus, in addition to being transferable across suspensions of different numbers of particles, the trained HIGNN can also be transferable to any type and magnitude of external forcing. 

\section{Results}\label{sec:results}
To assess the accuracy, efficiency, and transferability of the proposed HIGNN, we have applied it for simulating particulate suspensions of different numbers of particles and subject to different types of external forcing. For demonstration purposes and due to the code accessibility for generating training and validation data, we consider particulate suspensions in unbounded and periodic domains in this work. The particles are identical with the radius of $a = 1$. All numeric values are non-dimensional. The computational cost for training the HIGNN or for applying the trained HIGNN to prediction was evaluated on a desktop using one NVIDIA RTX 3070 GPU and one AMD 3900XT CPU @ 3.79 GHz, except in \S\ref{subsec:unbounded_largescale} where multiple GPUs were used for simulating a large-scale suspension.

In an unbounded or periodic domain, the HIs between particles are only related to their relative positions. Thus, the HIGNN model can be rewritten as:
\begin{equation}\label{Eq:3body_GNN_relative}
\begin{split}
    \mathbf{U}^{\text{HIGNN}}_i & = \boldsymbol{\alpha}_1 \cdot \mathbf{F}_i + \sum\limits_{\substack{j :(i,j)\in \boldsymbol{\mathcal{E}}}}  \mathbf{h}_{\boldsymbol{\Theta}_2} ( \mathbf{X}_j - \mathbf{X}_i) \cdot \mathbf{F}_{i,j} + \sum\limits_{j,k: (j,k,i) \in \boldsymbol{\mathcal{F}}}  \mathbf{g}_{\boldsymbol{\Theta}_3} (\mathbf{X}_j - \mathbf{X}_i, \mathbf{X}_k- \mathbf{X}_i) \cdot \mathbf{F}_{i,j} \;.
\end{split} 
\end{equation}
{Note that in a bounded domain, the HI between particles or the mobility tensor of particles will depend on the relative positions not only between the particles but also to the boundaries, for which the more general HIGNN model as in Eq. \eqref{Eq:U_surrogate_3body} must be followed.}

\subsection{Periodic domain}\label{subsec:periodic}
In this section, we consider particulate suspensions in a three-dimensional periodic box of length $32$. To construct the HIGNN, we only need to invoke the training process once. Due to its transferability, the trained HIGNN can be directly applied in all the test cases considered herein.

\subsubsection{Training of the HIGNN}
The model in Eq. \eqref{Eq:3body_GNN_relative} allows the input dimensions of $\mathbf{h}_{\boldsymbol{\Theta}_2} $ and $\mathbf{g}_{\boldsymbol{\Theta}_3} $ reduced to $3$ and $6$, respectively, while their output dimensions are both $3\times 6$. For the particles of $a = 1$ and the periodic box of length $32$, $\boldsymbol{\alpha}_1 = {0.982}/{( 6 \pi \mu)} \mathbf{I}$. Both $\mathbf{h}_{\boldsymbol{\Theta}_2} $ and $\mathbf{g}_{\boldsymbol{\Theta}_3}$ are modeled by MLP neural networks. Each has 4 hidden layers with 64, 256, 128, and 64 nodes on each layer, respectively. The hyperbolic tangent (tanh) is chosen as the activation function. 

We generated 80,000 different configurations of \textit{three} particles in the periodic box of length $32$. For each configuration, each of the three particles was assigned a force randomly sampled from: $\mathbf{F} = [1, 0, 0]^T$, $[0,1,0]^T$, or $[0,0, 1]^T$; and the resultant velocity of each particle was computed. To achieve higher accuracy, we could compute the velocities by numerically solving the boundary-valued Stokes equations. However, due to the code accessibility \cite{SD_opensource}, we used SD to generate data for training and validation purposes. The data was generated in a high-throughput way. The generated data were then used to train the HIGNN by minimizing the loss given in Eq. \eqref{Eq:Loss}. When generating data and training the HIGNN, without assuming any prior knowledge we set a sufficiently long cutoff $R_\text{cut} = 20.0$ for the three-body contributions and thereby for building the face connectivity in the graph. 
The stochastic gradient optimizer, Adam \cite{Kingma2015Adam}, was used for the optimization, with the batch size of 512. The learning rate ($lr$) was set by the scheduler: $lr = 0.001\times 0.5^{\lfloor \text{epoch}/100 \rfloor }$ (see Table \ref{tab:lr}). 
\begin{table}[htbp] 
	\centering
	\caption{Learning rate scheduler}
	\label{tab:lr}  
	\begin{tabular}{ccccc} 
		\hline\hline\noalign{\smallskip}	
		Epoch & 0 $\sim$ 100 & 100 $\sim$ 200 & 200 $\sim$ 300 & 300 $\sim$ 400  \\
		\noalign{\smallskip}\hline\noalign{\smallskip}
		Learning rate & 0.001 & 0.0005 & 0.00025 & 0.000125 \\
		\noalign{\smallskip}\hline
	\end{tabular}
\end{table}

\begin{figure}[htbp]
\centering
\includegraphics[width = 0.6\linewidth ]{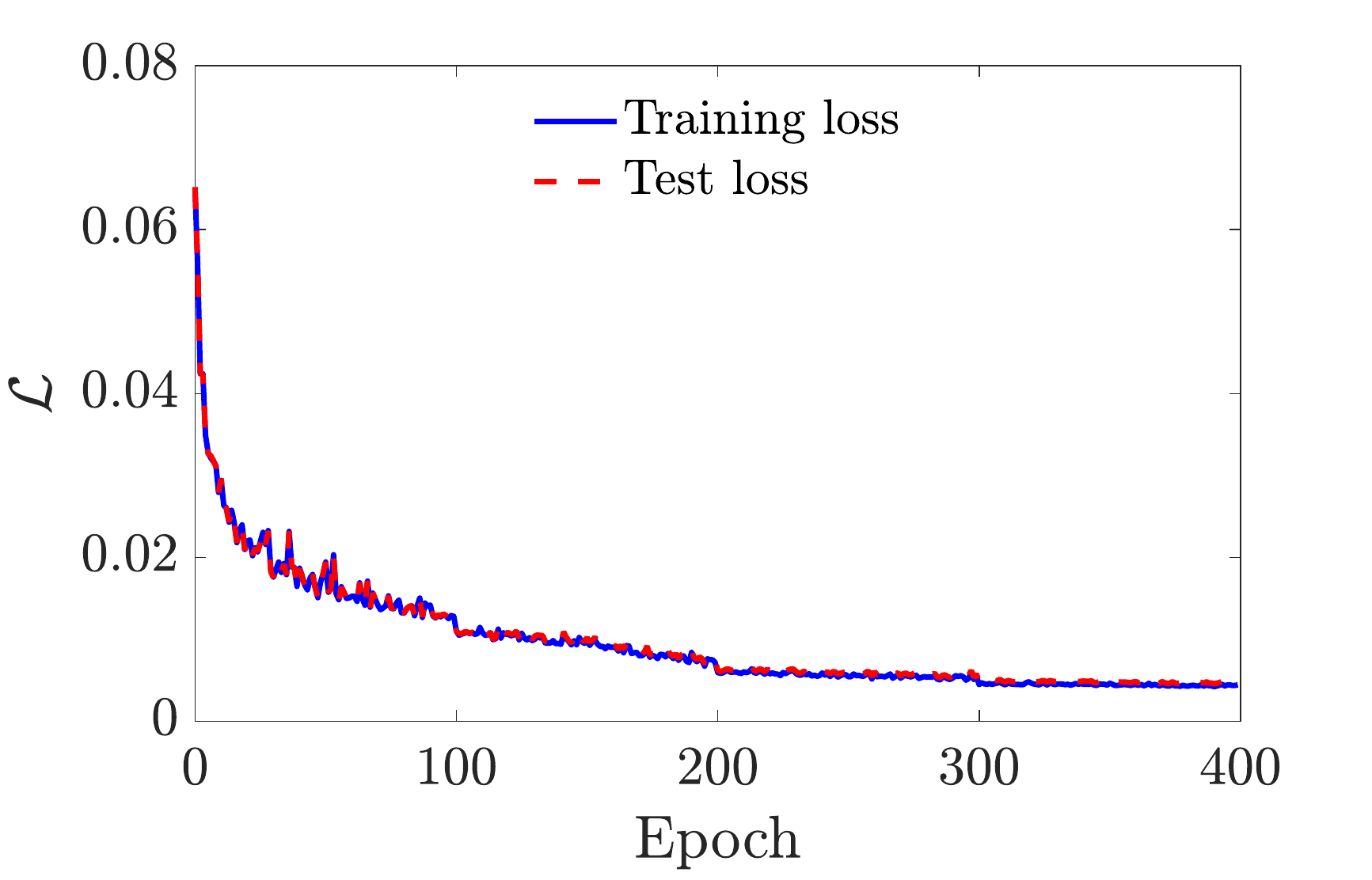}
\caption{Periodic domain: Convergence of the training and test losses for training the HIGNN.}
\label{fig:learning_curve}
\end{figure}

Fig.~\ref{fig:learning_curve} depicts the training performance by examining the decrease and convergence of training and test losses, where $N_\text{train}=60,000$ data are used for training and 20,000 data are used for test. The training process took $\sim$49 min on a desktop using one NVIDIA RTX 3070 GPU and one AMD 3900XT CPU @ 3.79 GHz. Once the HIGNN is trained, the surrogate modeling given in Eq. \eqref{Eq:3body_GNN_relative} is determined and can be applied to simulating different particulate suspensions in the same periodic domain. 


\subsubsection{A benchmark problem}\label{subsec:periodic_benchmark}

\begin{figure}[htbp]
\centering
\begin{subfigure}{0.45\textwidth}
\centering
\includegraphics[width=0.8\textwidth]{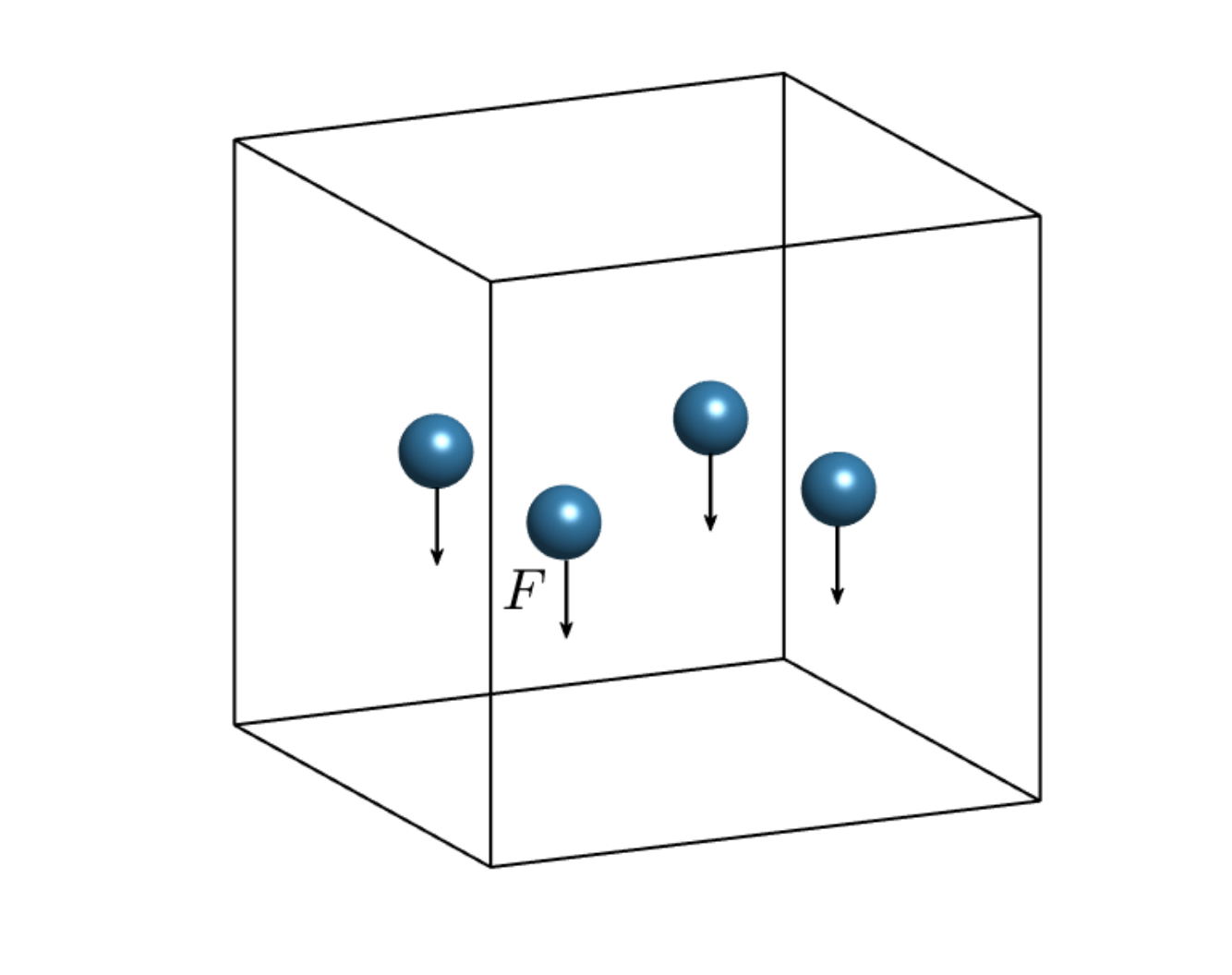}
\caption{Problem setup with the force perpendicular to the lattice plane.}
\label{sfig:setup_square_perp}
\end{subfigure}
\quad
\begin{subfigure}{0.45\textwidth}
\centering
\includegraphics[width=0.8\textwidth]{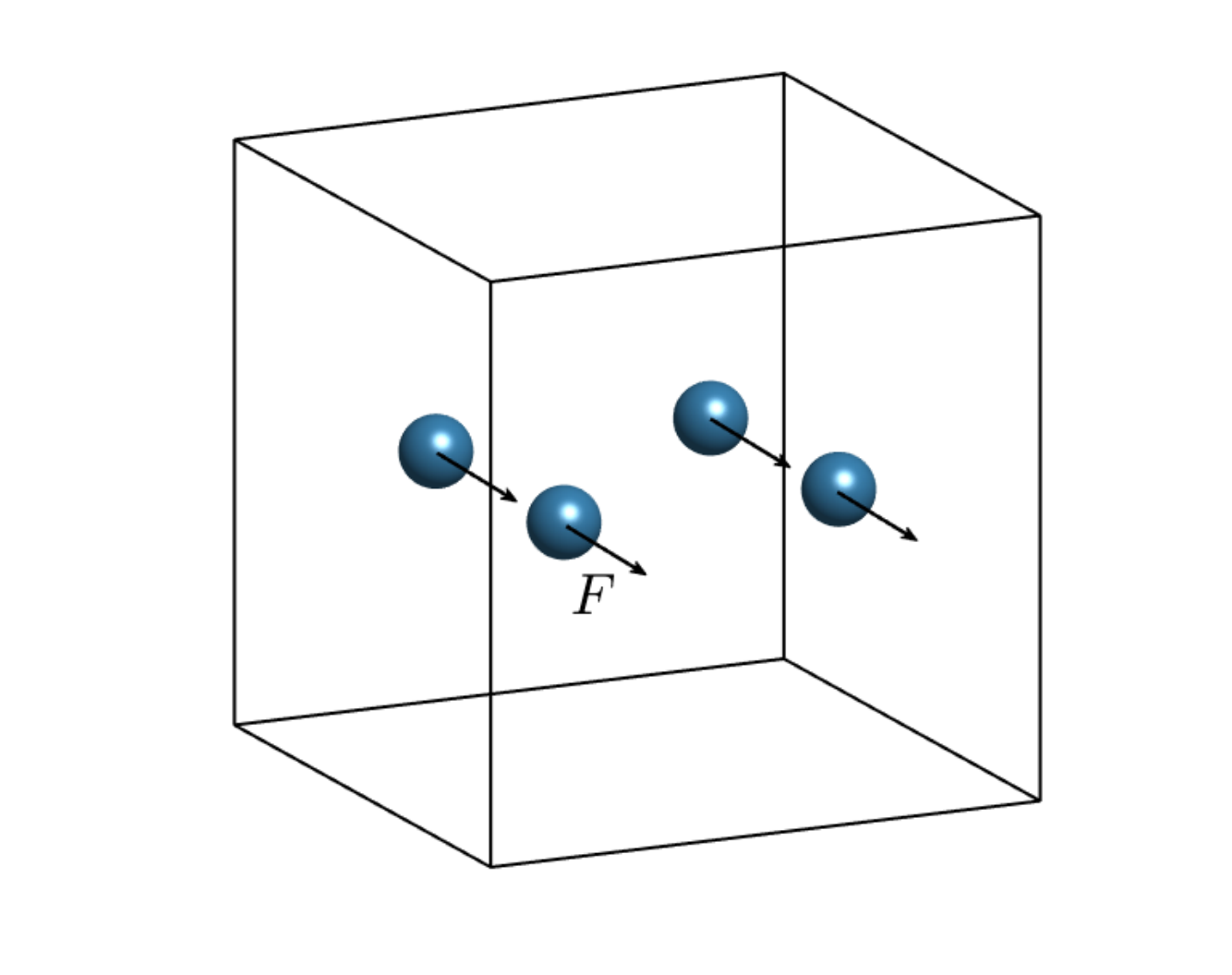}
\caption{Problem setup with the force parallel to a side of the lattice.}
\label{sfig:setup_square_para}
\end{subfigure}
\quad
\begin{subfigure}{0.45\textwidth}
\centering
\includegraphics[width=1.0\textwidth]{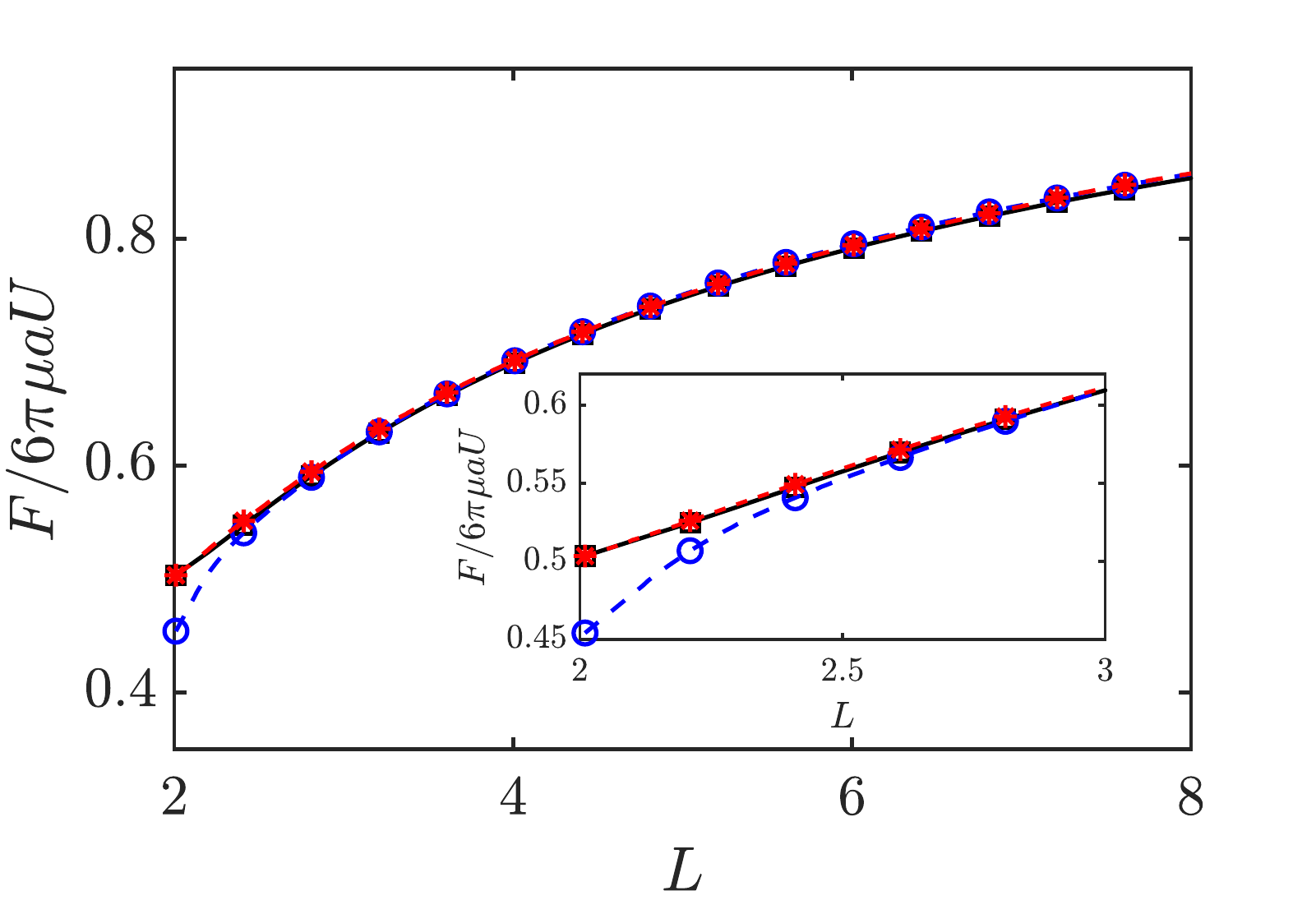}
\caption{Drag coefficient for the setup in $(a)$.}
\label{sfig:tri_comparison_perp}
\end{subfigure}
\quad
\begin{subfigure}{0.45\textwidth}
\centering
\includegraphics[width=1.0\textwidth]{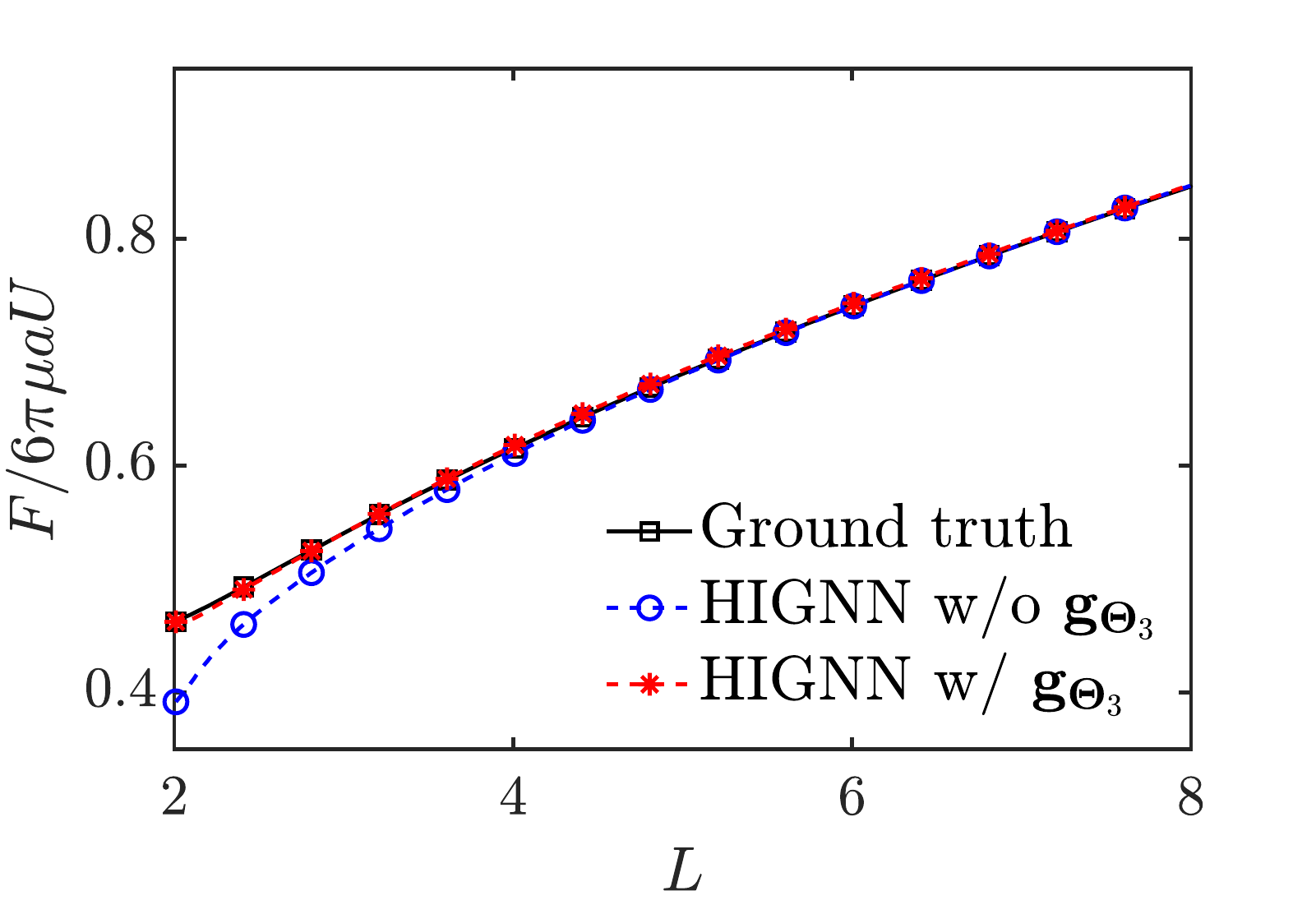}
\caption{Drag coefficient for the setup in $(b)$.}
\label{sfig:tri_comparison_para}
\end{subfigure}
\caption{Drag coefficient ($ F/6\pi \mu a U$) predicted for one of the four particles positioned in a square lattice of length $L$.}
\label{fig:tri_comparison}
\end{figure}

The trained HIGNN was first applied to simulating a benchmark problem with four particles, where the particles are positioned in a square lattice of length $L$ and subject to uniform forces perpendicular or parallel to the lattice, as illustrated in Figs. \ref{sfig:setup_square_perp} and \ref{sfig:setup_square_para}. This problem was adapted from \cite{Durlofsky1987SD}, where an unbounded domain was considered, while in this work the four particles are in a periodic box. This problem is appropriate for validating the accuracy of the HIGNN and demonstrating the importance of three-body HI effects.

Given different $L$ (distance between the particles' centers), the particles' velocities predicted by the HIGNN are plotted in Figs.~\ref{sfig:tri_comparison_perp} and \ref{sfig:tri_comparison_para} and compared with the ground truth (SD simulation results). We find that for a wide range of separation distances between the particles, the predictions by the HIGNN consistently show good accuracy. Note that when $L = 2.01 $, the particles are in close contact with a separation of only $0.01 a$. We hence validate that the HIGNN is accurate in capturing both the long-range HI and near-field lubrication effects.

To examine the effect of three-body HI, we further compare the predictions with or without the last term in Eq. \eqref{Eq:3body_GNN_relative}. When the particles' distance $L > 5.0$, the contribution of three-body HI can be negligible. However, as the particles' separation is smaller, the contribution of three-body HI to the particles' velocities becomes more significant and must be included in the model to ensure accuracy. From these findings, the cutoff for three-body contributions is set as $R_\text{cut} = 5.0$ henceforth in \S\ref{subsec:periodic} for building the face connectivity in the graph when applying the HIGNN for predictions.  

 

\subsubsection{Assessment of the prediction cost}
\begin{figure}[htbp]
\centering
\includegraphics[width = 0.5\linewidth ]{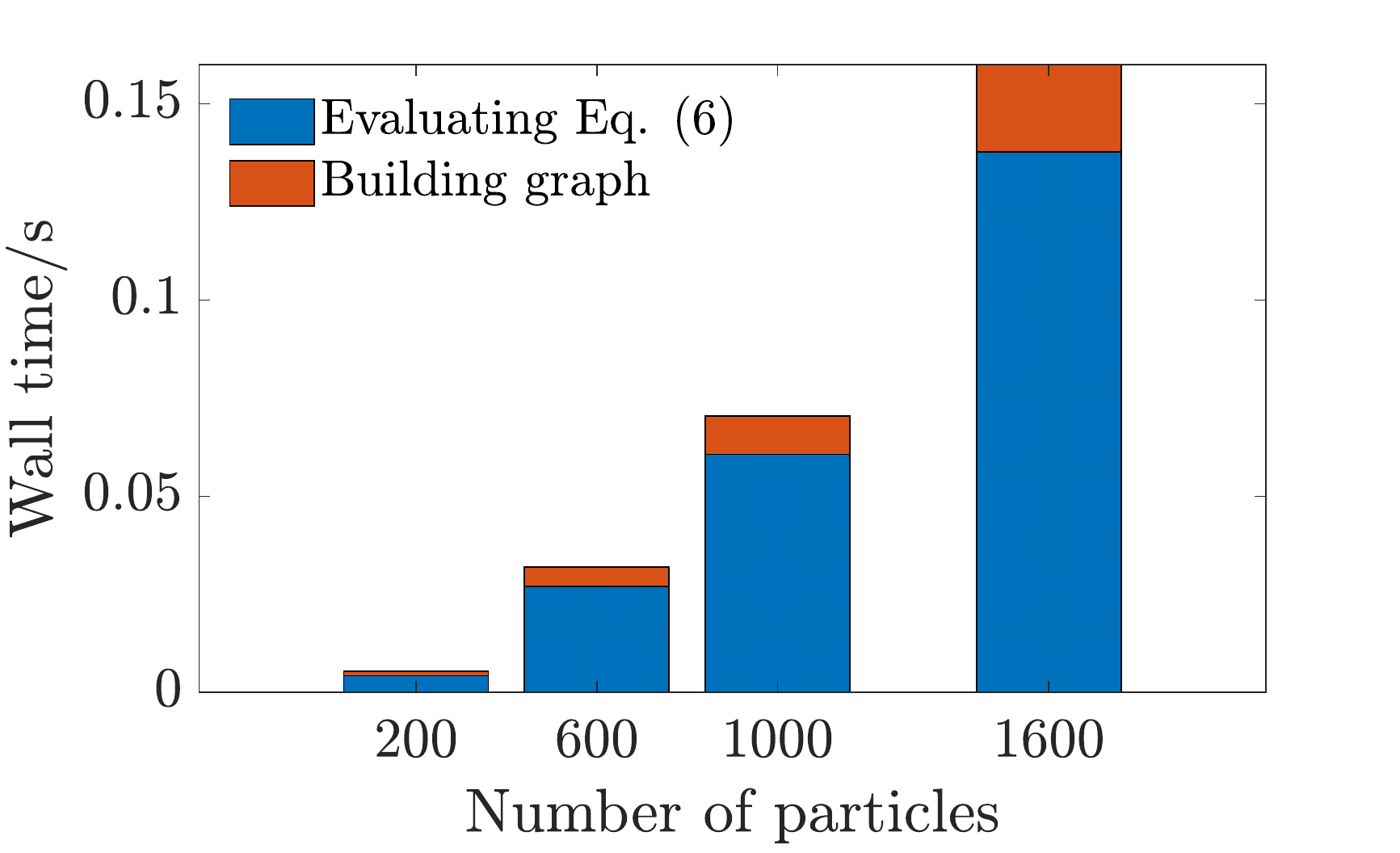}
\caption{Computer time cost by the HIGNN for inferring all particles' velocities at one time step for different numbers of particles.}
\label{fig:wall_time}
\end{figure}

Next, we assess the cost of using the trained HIGNN for prediction. Varying numbers (200-1,600) of particles are placed to form a primitive cubic lattice structure \cite{birkhoff1940lattice} at the center of the periodic domain and subject to the gravitational force as in the previous case. We predict the velocities of all particles by the HIGNN. The prediction cost comprises two parts: 1) building the graph for the particle system, i.e., determining all entities for $\boldsymbol{\mathcal{G} = (\mathcal{V}, \mathcal{E}, \mathcal{F})}$; 2) evaluating Eq. \eqref{Eq:3body_GNN_relative}. The wall time for inferring all particles' velocities by the HIGNN for one time step is summarized in Fig.~\ref{fig:wall_time}, where the blue bar corresponds to the time of evaluating Eq. \eqref{Eq:3body_GNN_relative}, and the orange bar denotes the time spent for building the graph. As can be seen, while building the graph only contributes up to $1/7$ of the total cost, evaluating Eq. \eqref{Eq:3body_GNN_relative} dominates the prediction cost. 

Using one GPU, the HIGNN costs only 0.0054 sec to predict 200 particles' velocities; if the number of particles increases to 1,600, the wall time spent by the HIGNN is 0.16 sec. 
As a reference, the same tests need 0.4 and 5.0 secs to compute the velocities for 200 and 1,600 particles, respectively, for one time step, using the most recent parallel accelerated Stokesian dynamics (ASD) \cite{Ouaknin2021ParaASD} and 8 CPUs of one Sandy Bridge processor on the supercomputer Stampede \cite{SuperComputer}. Through these assessments, we demonstrate the superior computational efficiency of the HIGNN for simulating particulate suspensions.

\subsubsection{Suspensions of self-assembling particles}
\begin{figure}[htbp]
\centering
\includegraphics[width = 0.5\linewidth]{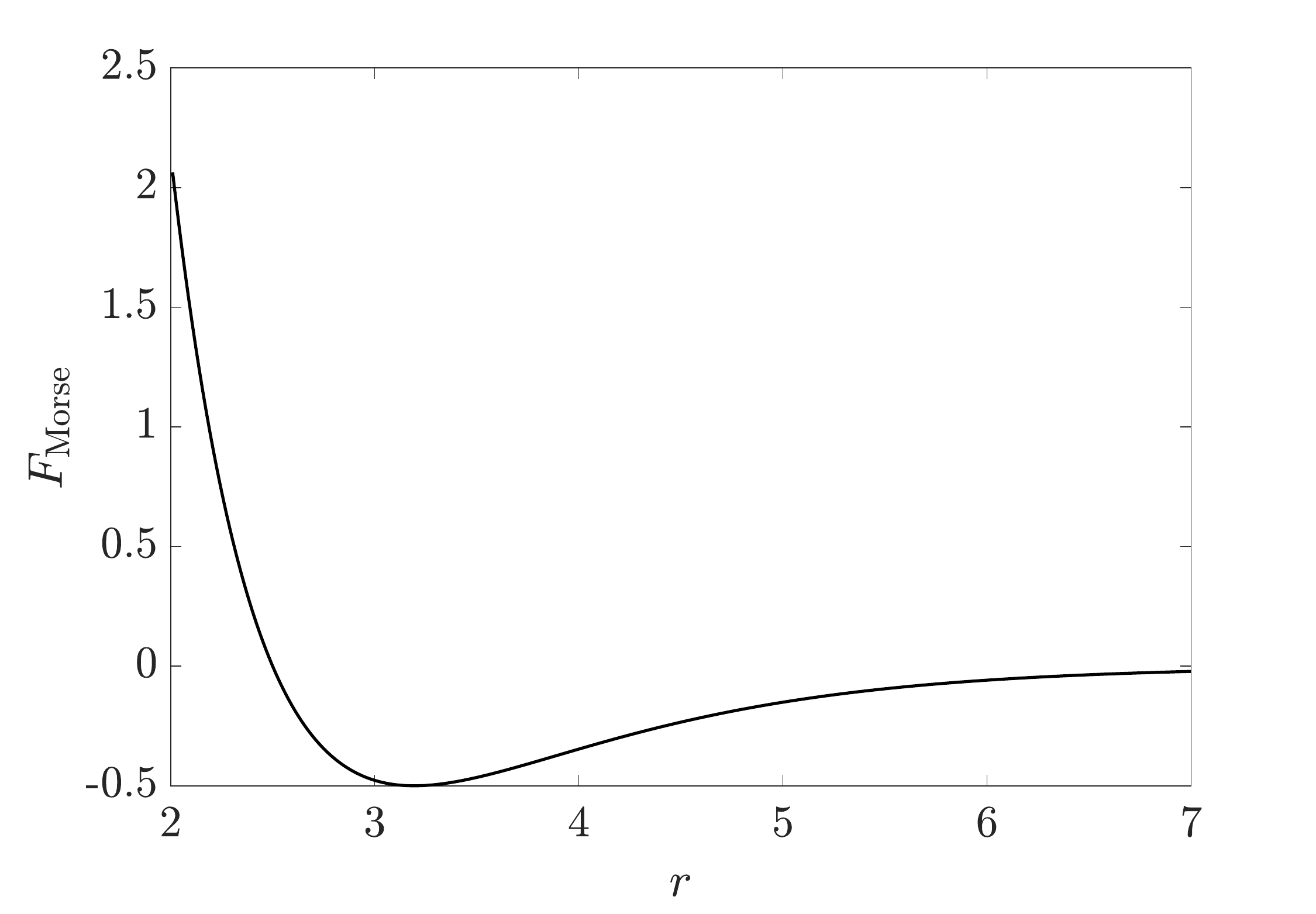}
\caption{The interparticle Morse potential force defined in Eq. \eqref{eq:Morse}. Here, the negative value means the force is attractive; the positive value implies the force is repulsive.}
\label{fig:F_morse}
\end{figure}

Finally, to demonstrate its transferability across different external forcing, the trained HIGNN is applied for simulating suspensions of self-assembling particles. In particular, the particles, initially uniformly distributed in the periodic box, are subject to the interparticle Morse potential: 
\begin{equation}\label{eq:Morse}
    F_{\text{Morse}} (r) = -2 \rho_M D_e ( e^{-\rho_M(r_e - r)} - e^{-2\rho_M(r_e - r)} ) \;,
\end{equation}
where $r$ is the center-to-center distance between particles; $D_e$ is the depth of the attractive potential; $ r_e $ is the equilibrium distance; and $\rho_M $ is a parameter controlling the interaction range. Following \cite{ColloidNS_NCM2019}, the parameters are set as: $ \rho_M = 1 $, $D_e = 1$, and $r_e = 2.5$. As depicted in Fig.~\ref{fig:F_morse}, the interparticle Morse potential is attractive in long range and repulsive in short range, which drives the particles assembling into clusters.
Thus, the total external force on each particle is given by:
\begin{equation}\label{eq:Morse_totalforce}
    \mathbf{F}_{i} = \sum\limits_{j \neq i} F_{\text{Morse}} (r_{ij}) \hat{\mathbf{r}}_{ij} \;,
\end{equation}
where $\hat{\mathbf{r}}_{ij} = \frac{\mathbf{X}_i - \mathbf{X}_j}{| {\mathbf{X}_i - \mathbf{X}_j}|}$ is the unit vector pointing from particle $j$ to particle $i$. 

\begin{figure}
\centering
\begin{subfigure}{0.24\textwidth}
\centering
\includegraphics[width=1.1\textwidth]{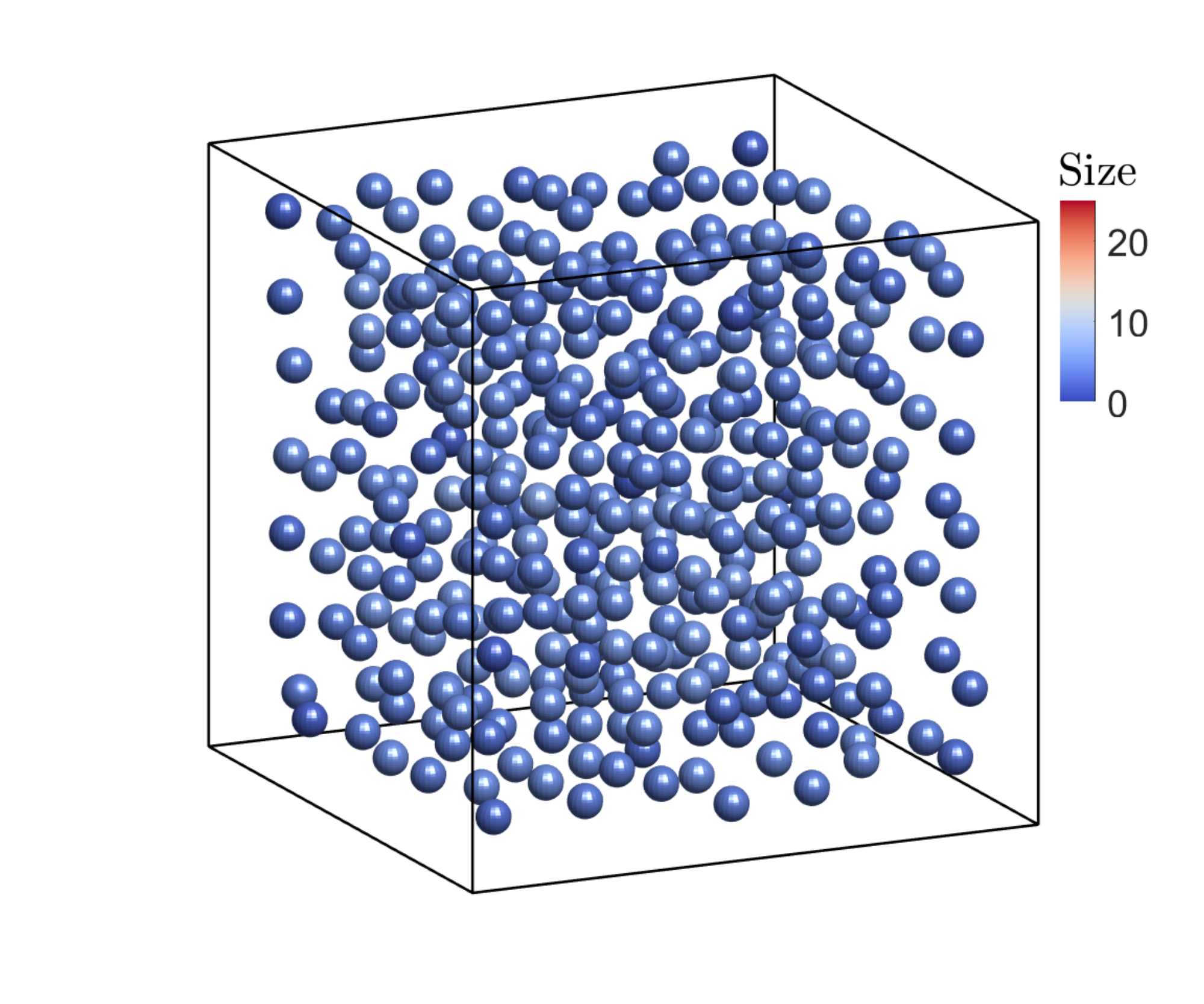}
\caption*{$t = 0 $}
\label{sfig:phi2_t0}
\end{subfigure}
\hfill
\begin{subfigure}{0.24\textwidth}
\centering
\includegraphics[width=1.1\textwidth]{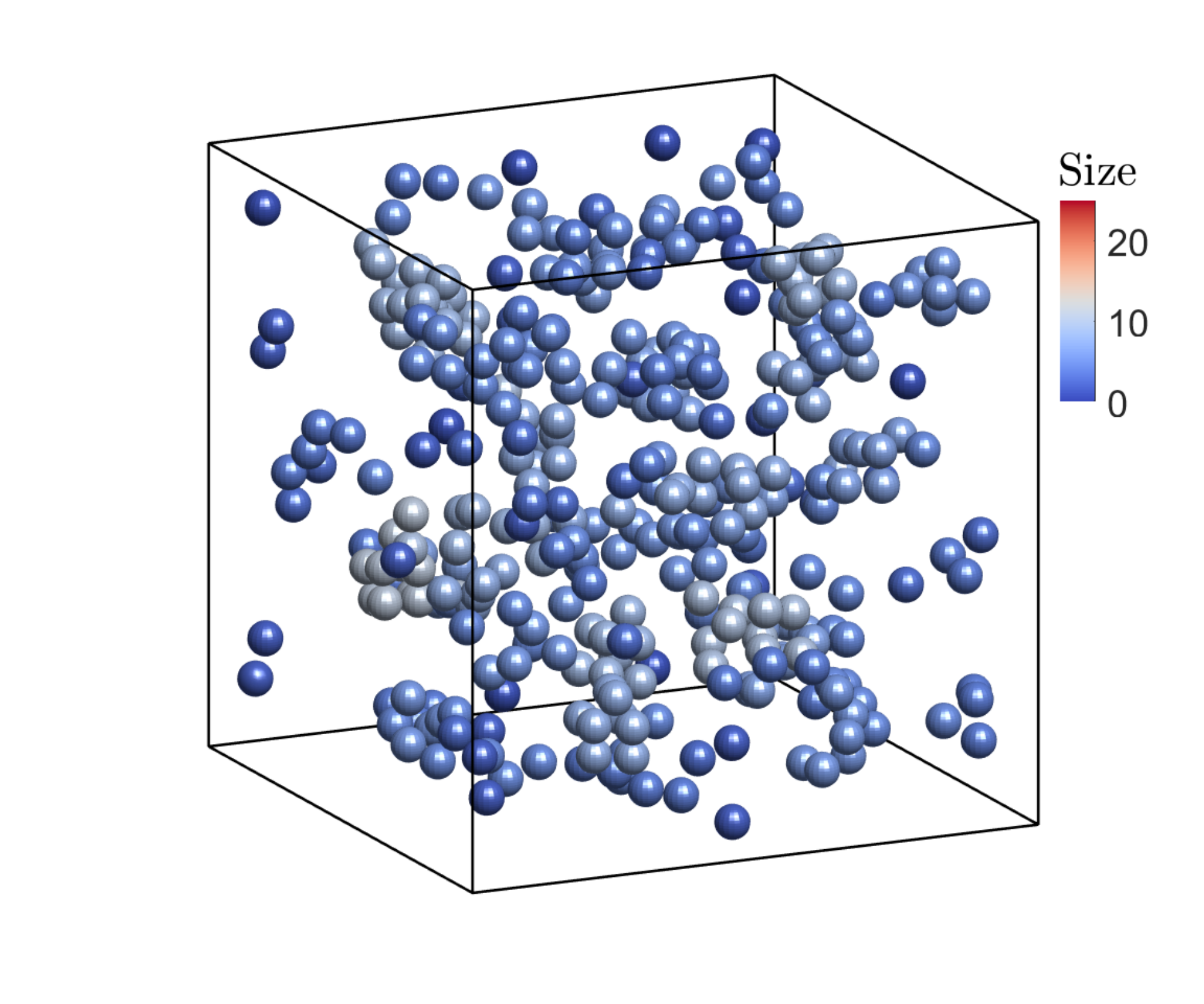}
\caption*{$t = 10 $}
\label{sfig:phi2_t1}
\end{subfigure}
\hfill
\begin{subfigure}{0.24\textwidth}
\centering
\includegraphics[width=1.1\textwidth]{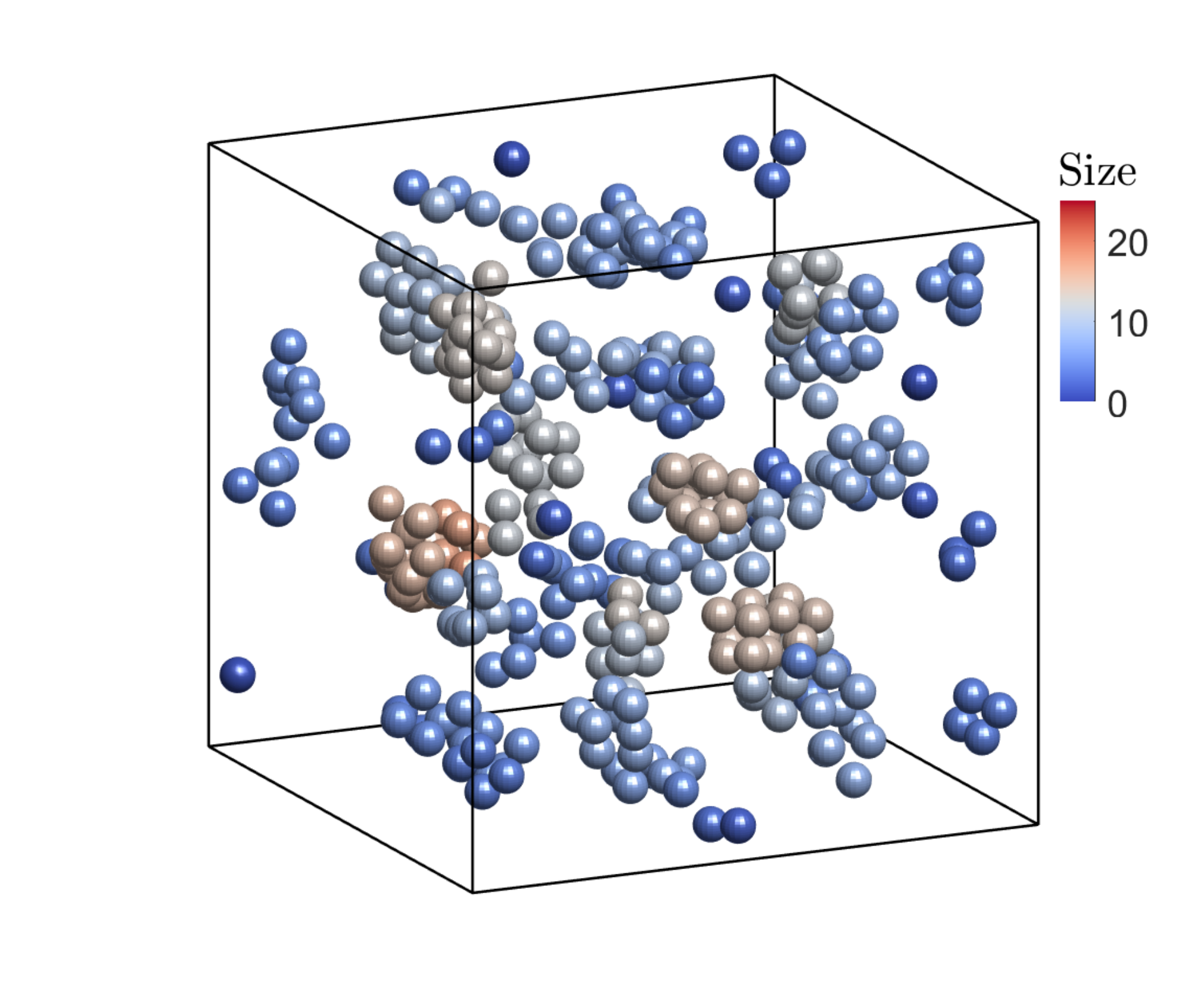}
\caption*{$t = 20 $}
\label{sfig:phi2_t2}
\end{subfigure}
\hfill
\begin{subfigure}{0.24\textwidth}
\centering
\includegraphics[width=1.1\textwidth]{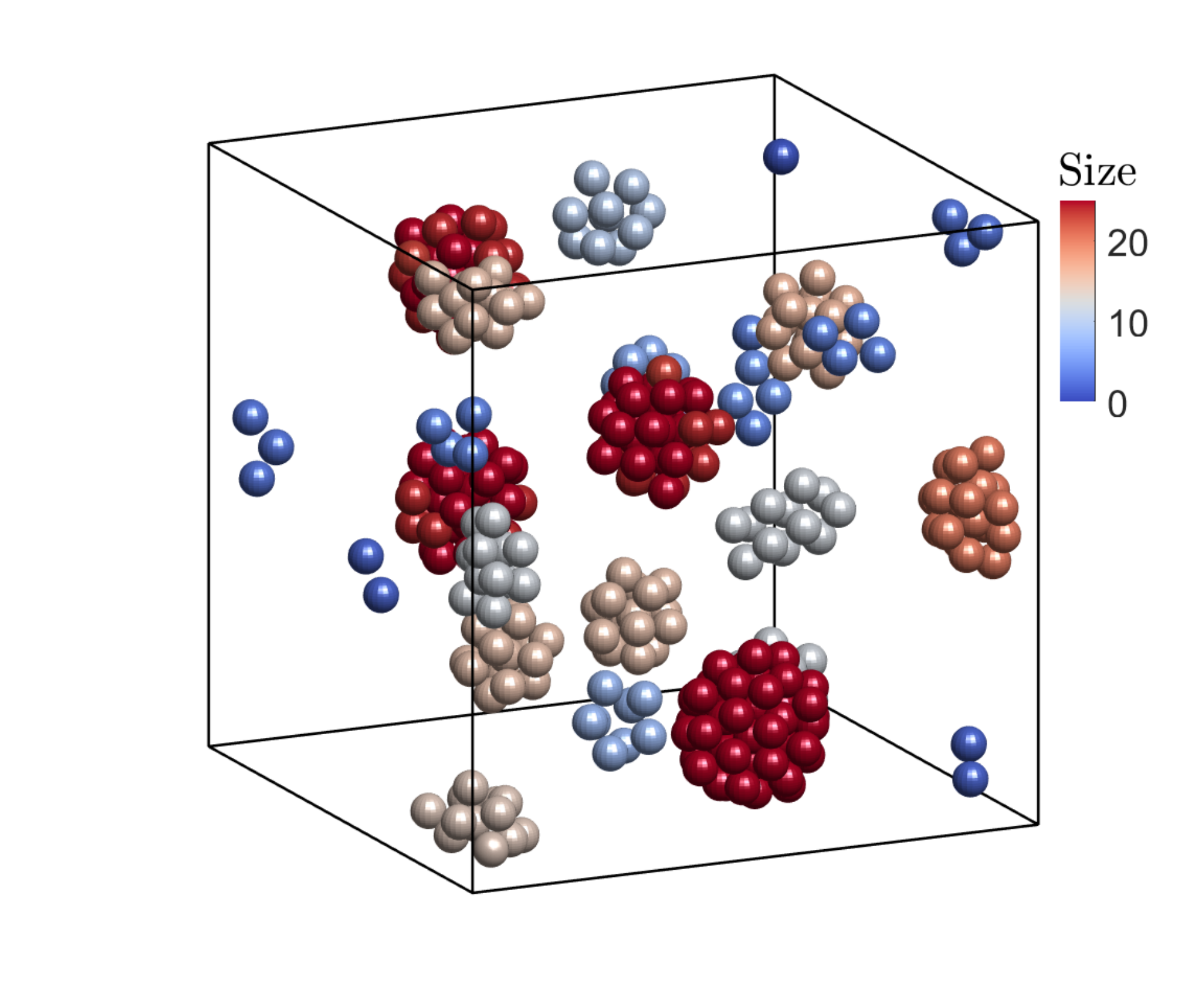}
\caption*{$t = 40 $}
\label{sfig:phi2_t4}
\end{subfigure}
\vfill
\begin{subfigure}{0.24\textwidth}
\centering
\includegraphics[width=1.1\textwidth]{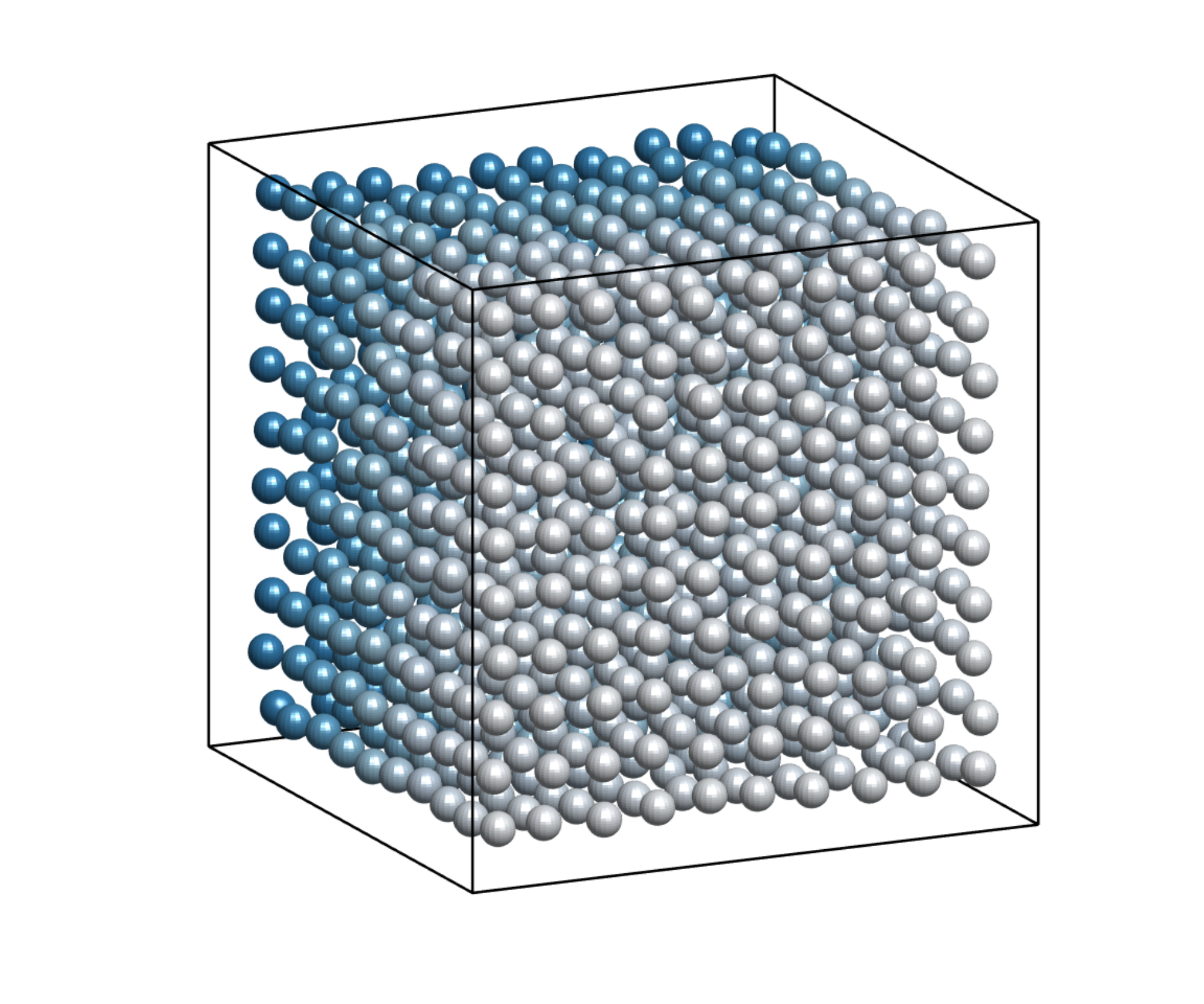}
\caption*{$t = 0$}
\label{sfig:phi12_t0}
\end{subfigure}
\hfill
\begin{subfigure}{0.24\textwidth}
\centering
\includegraphics[width=1.1\textwidth]{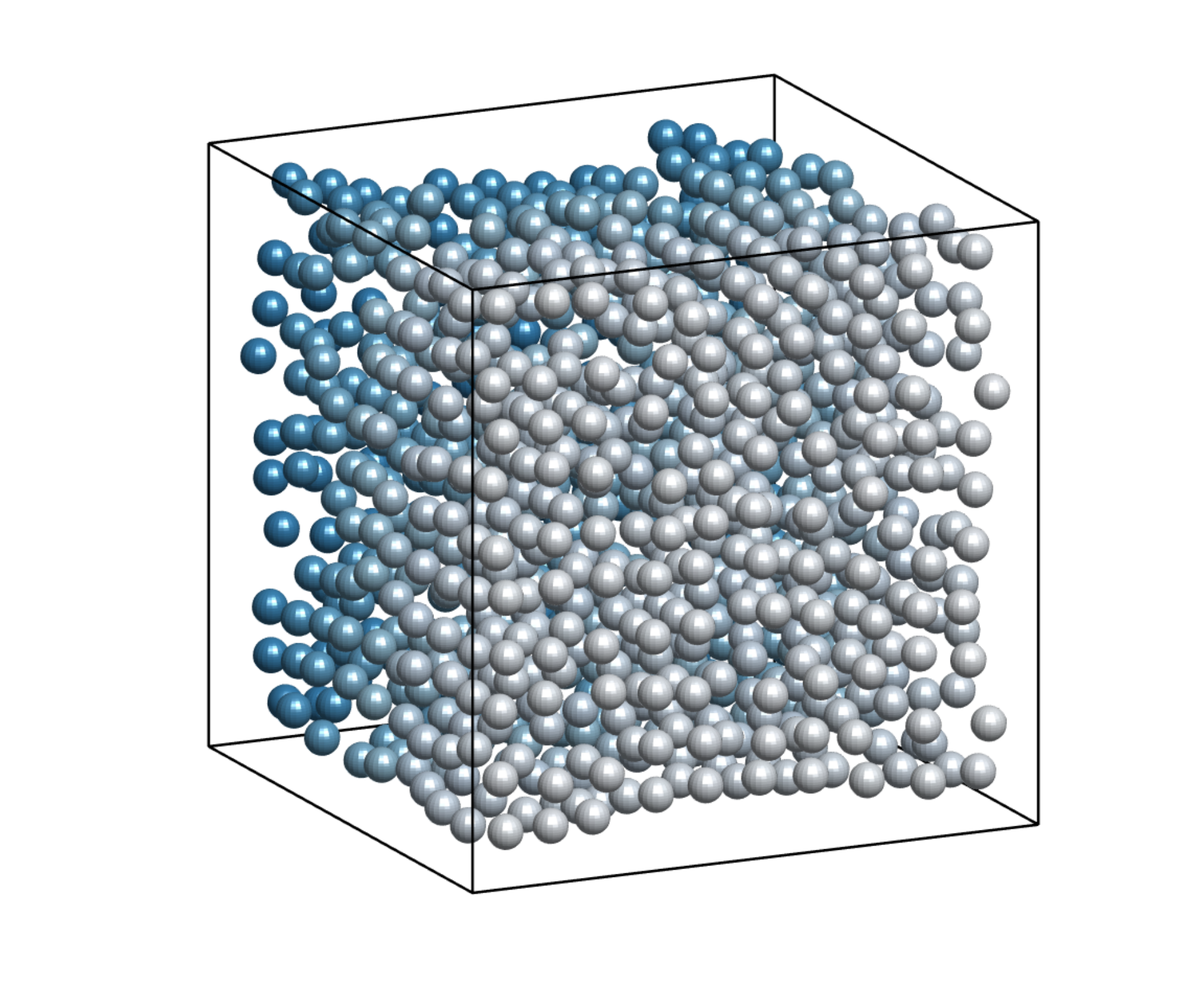}
\caption*{$t = 0$}
\label{sfig:phi12_t1}
\end{subfigure}
\hfill
\begin{subfigure}{0.24\textwidth}
\centering
\includegraphics[width=1.1\textwidth]{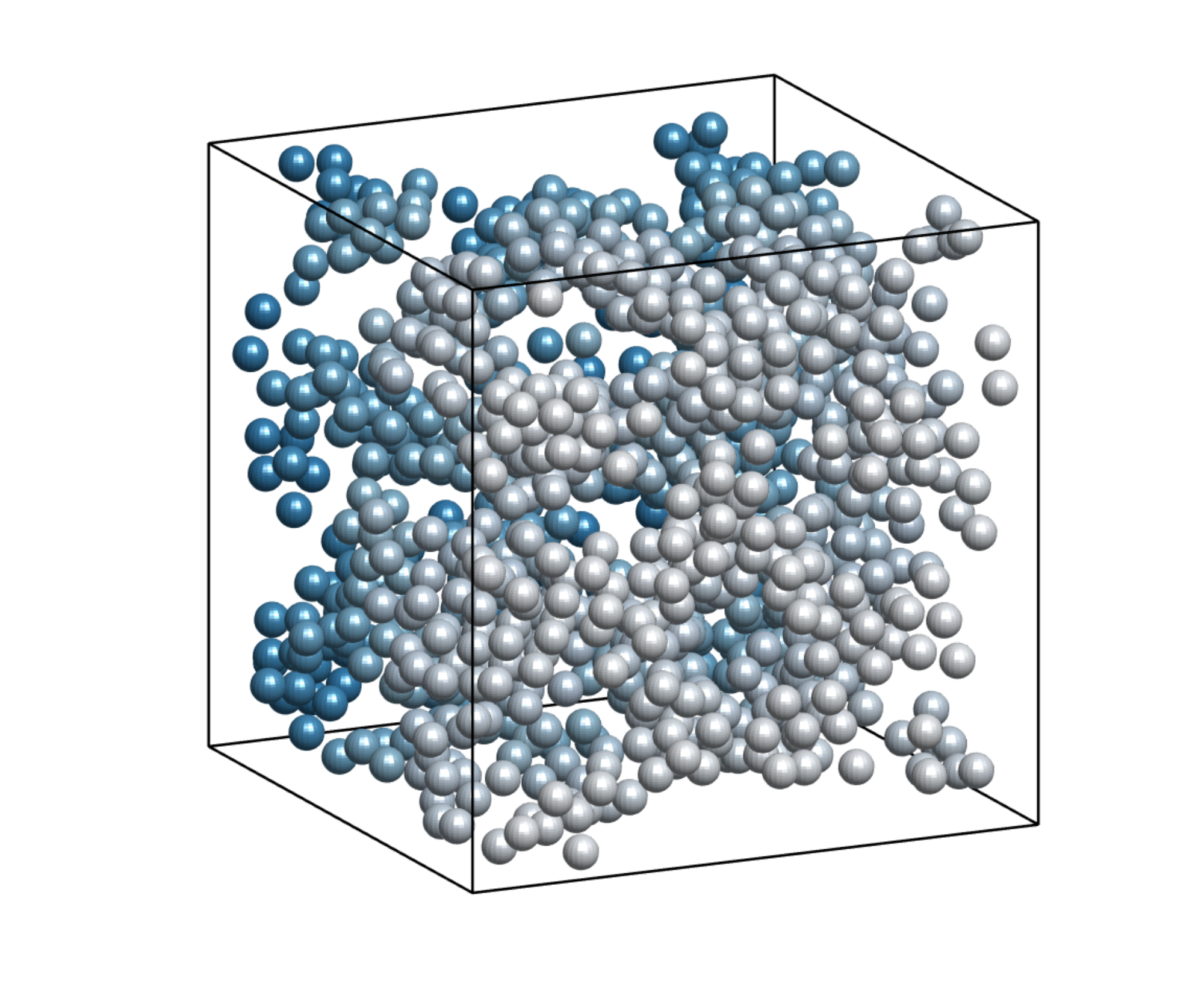}
\caption*{$t = 3 $}
\label{sfig:phi12_t2}
\end{subfigure}
\hfill
\begin{subfigure}{0.24\textwidth}
\centering
\includegraphics[width=1.1\textwidth]{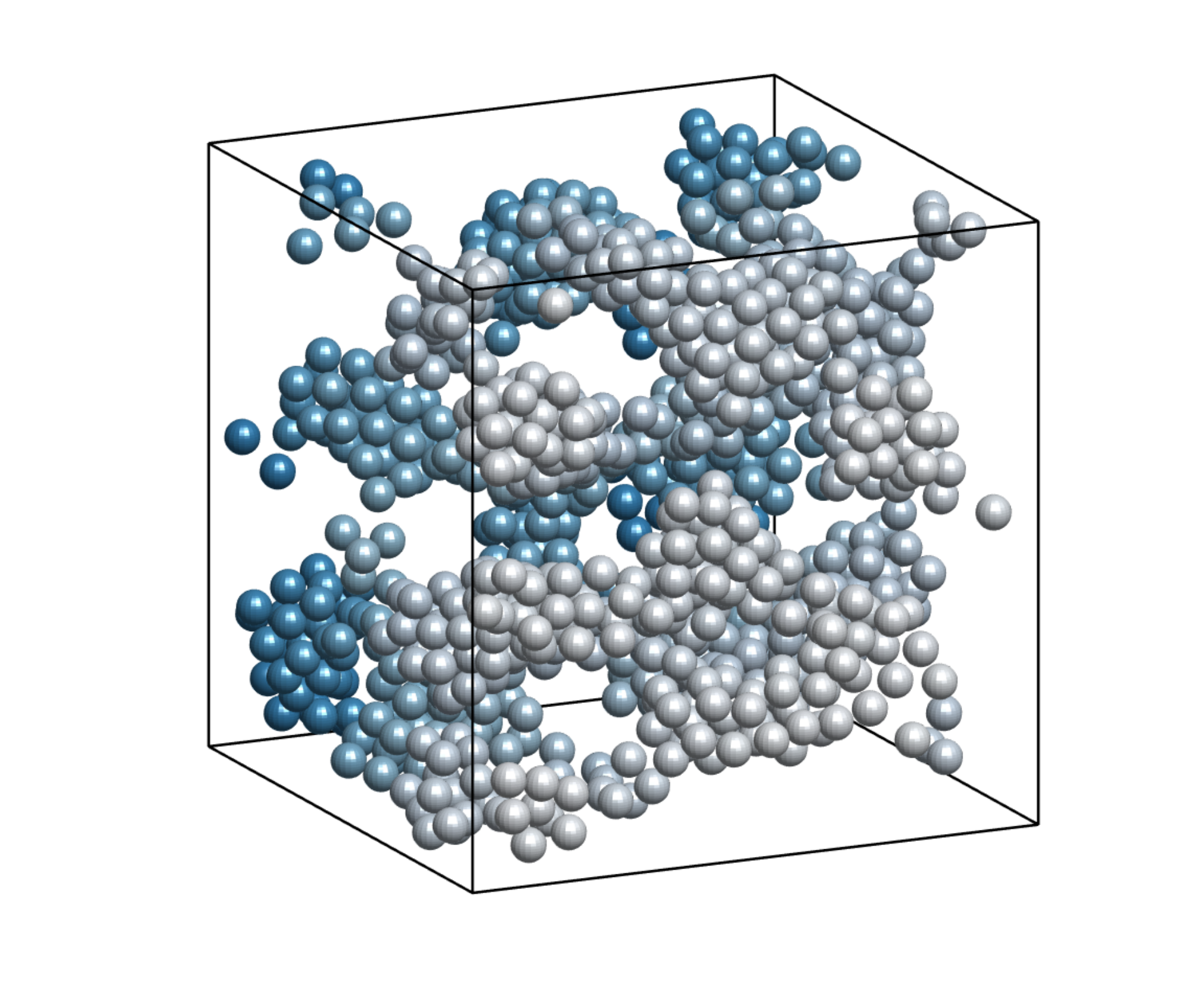}
\caption*{$t = 6 $}
\label{sfig:phi12_t4}
\end{subfigure}
\caption{Self-assembly of 343 (top) and 1000 (bottom) particles in suspension with the volume fraction of 4.4\% and 12.7\%, respectively. In the top panel, the particle color represents the number of particles in each cluster (see the color bars). In the bottom panel, particles are colored to distinguish the particles in the front from the back.}
\label{fig:phi2}
\end{figure}

Since the HIGNN is transferable across different numbers of particles and different types of forces, without retraining we used the previously trained HIGNN and the external force calculated from Eq. \eqref{eq:Morse_totalforce} to predict the velocities of all particles at each time step, and employed the explicit Euler temporal integrator to update the particles' positions with the time step $\Delta t = 0.0005$. The assemblies predicted at two different volume fractions are illustrated in Fig.~\ref{fig:phi2}. In either case, the HIGNN consistently reproduces all stages of phase separation observed in \cite{ColloidNS_NCM2019}. 
The computer time spent for each case using the same hardware as in other cases is reported below: in the case of 343 particles, for each time step it took {0.0081 sec} to predict the velocities of all particles, and the entire simulation ran for {80,000} time steps and cost $\sim$10.8 min; in the case of 1,000 particles, for each time step it took {0.054 sec} to predict the velocities of all particles, and the entire simulation ran for {12,000} time steps and cost $\sim$10.5 min.

\subsection{Unbounded domain}\label{subsec:unbounded}
In this section, we consider a series of particulate suspensions in an unbounded domain. Same as in \S\ref{subsec:periodic} for a periodic domain, we only need to train the HIGNN once and then can directly apply the trained HIGNN in all the following cases, due to its transferability. 

\subsubsection{Training of the HIGNN}
We still employ the model in Eq. \eqref{Eq:3body_GNN_relative}. For the particles of $a = 1$ in an unbounded domain, $\boldsymbol{\alpha}_1 = {1}/{( 6 \pi \mu)} \mathbf{I}$. Both $\mathbf{h}_{\boldsymbol{\Theta}_2} $ and $\mathbf{g}_{\boldsymbol{\Theta}_3}$ are modeled by MLP neural networks. Each has 4 hidden layers with 64, 256, 128, and 64 nodes on each layer, respectively. The hyperbolic tangent (tanh) is chosen as the activation function. 

\begin{figure}[htbp]
\centering
\includegraphics[width = 0.6\linewidth ]{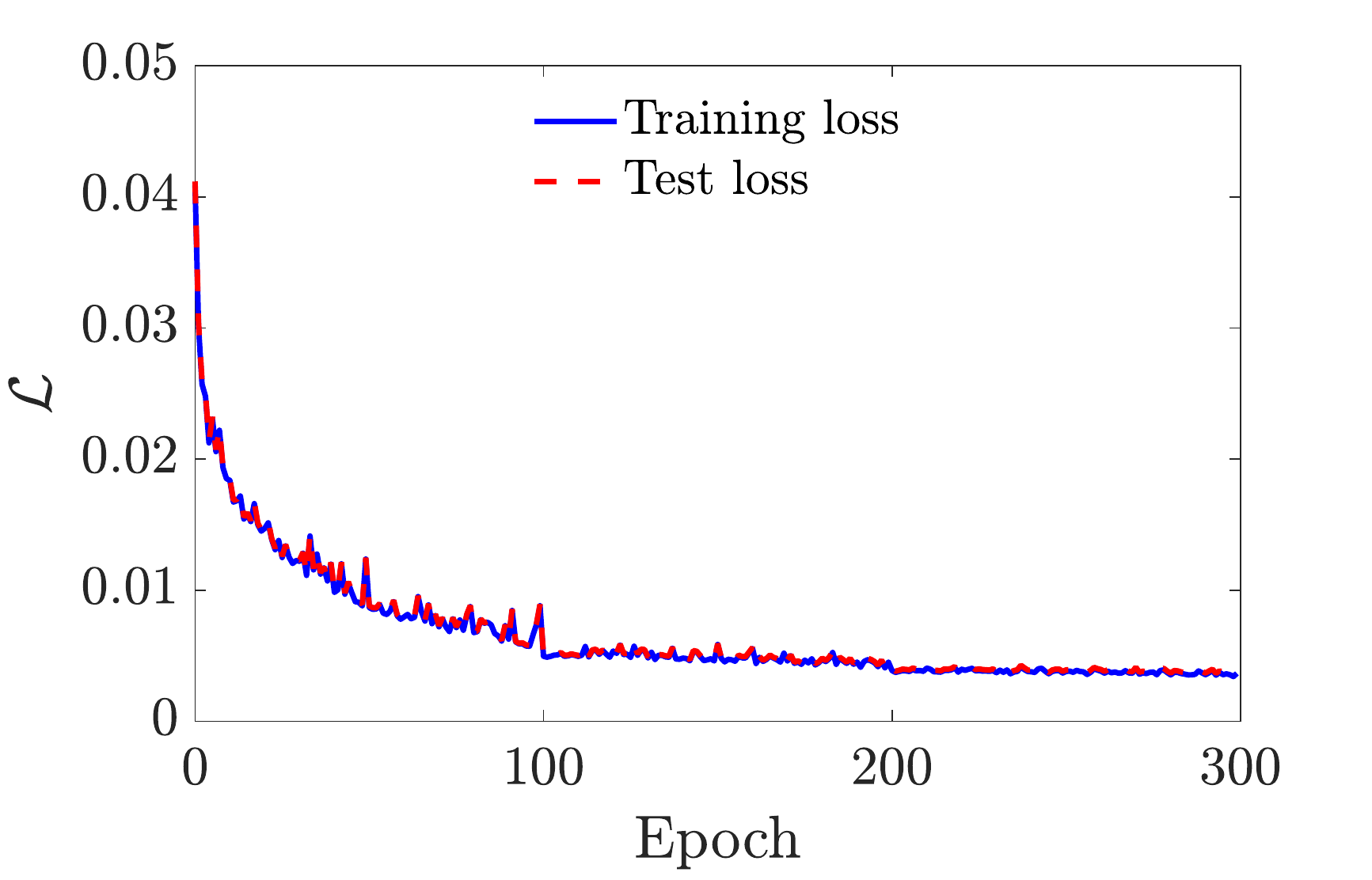}
\caption{Unbounded domain: Convergence of the training and test losses for training the HIGNN.}
\label{fig:learning_curve_unbounded}
\end{figure}

We generated 60,000 different configurations of \textit{three} particles in the unbounded domain with the largest distance of $ 500 $. For each configuration, each of the three particles was assigned a force randomly sampled from: $\mathbf{F} = [1, 0, 0]^T$, $[0,1,0]^T$, or $[0,0, 1]^T$; and the resultant velocity of each particle was computed. The data was generated by SD in a high-throughput way for training and validation purposes. Based on our findings in \S\ref{subsec:periodic_benchmark}, this time we set the three-body cutoff as $R_\text{cut} = 5$ in the training process. We further confirmed it in \S\ref{subsec:unbounded_benchmark}. The batch size for training the neural networks was set as 512, and the learning rate ($lr$) was still set by the scheduler: $lr = 0.001\times 0.5^{\lfloor \text{epoch}/100 \rfloor }$ (see Table \ref{tab:lr}). The training loss and the test loss are depicted in Fig.~\ref{fig:learning_curve_unbounded}, where the ratio between the training data and test data is $5:1$. It took $\sim$25 min to train the HIGNN. 

\subsubsection{A benchmark problem}\label{subsec:unbounded_benchmark}
\begin{figure}
\centering
\begin{subfigure}{0.45\textwidth}
\centering
\includegraphics[width=0.8\textwidth]{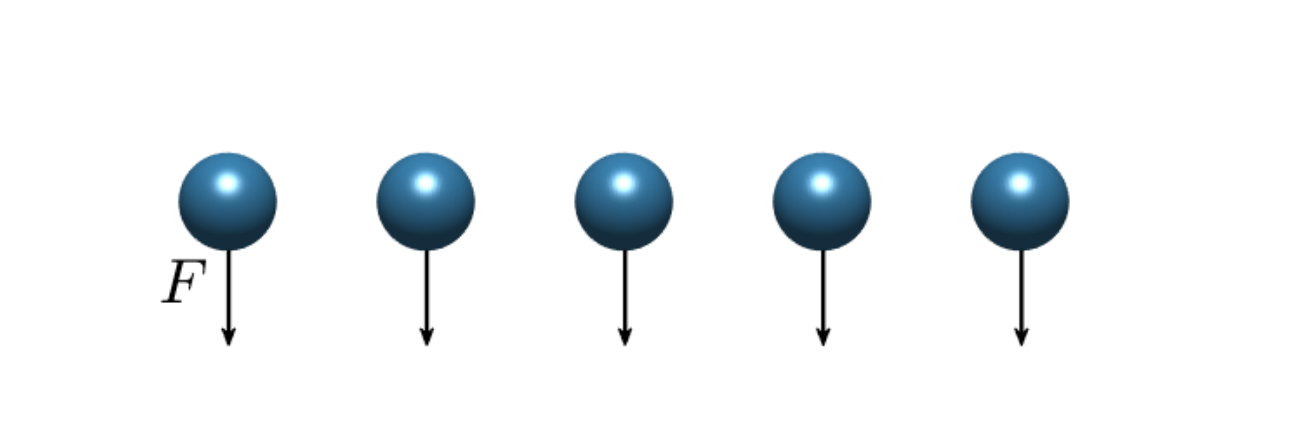}
\caption{Problem setup with the force perpendicular to the chain.}
\label{sfig:setup_perp}
\end{subfigure}
\quad
\begin{subfigure}{0.45\textwidth}
\centering
\includegraphics[width=0.8\textwidth]{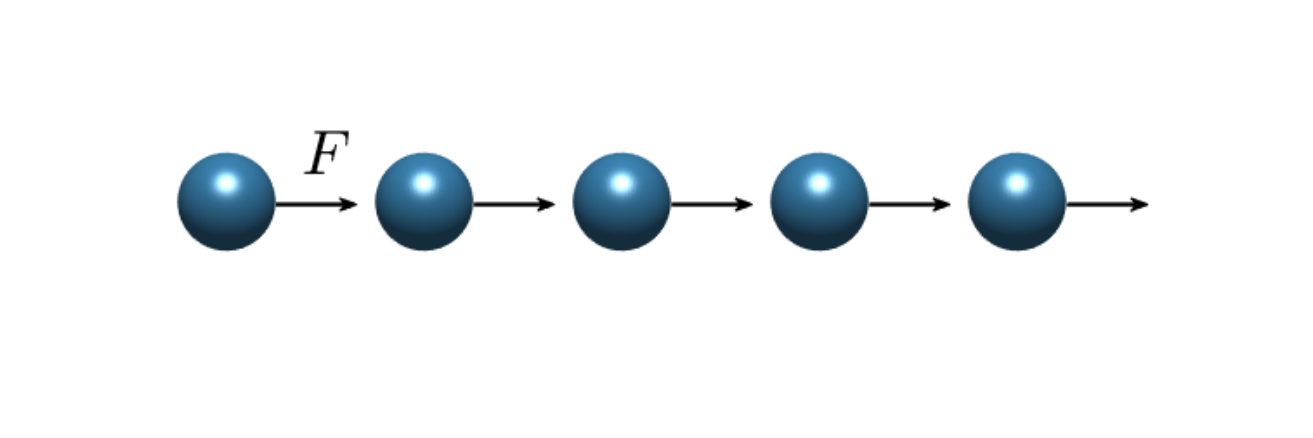}
\caption{Problem setup with the force parallel to the chain.}
\label{sfig:setup_para}
\end{subfigure}
\quad
\begin{subfigure}{0.45\textwidth}
\centering
\includegraphics[width=1.0\textwidth]{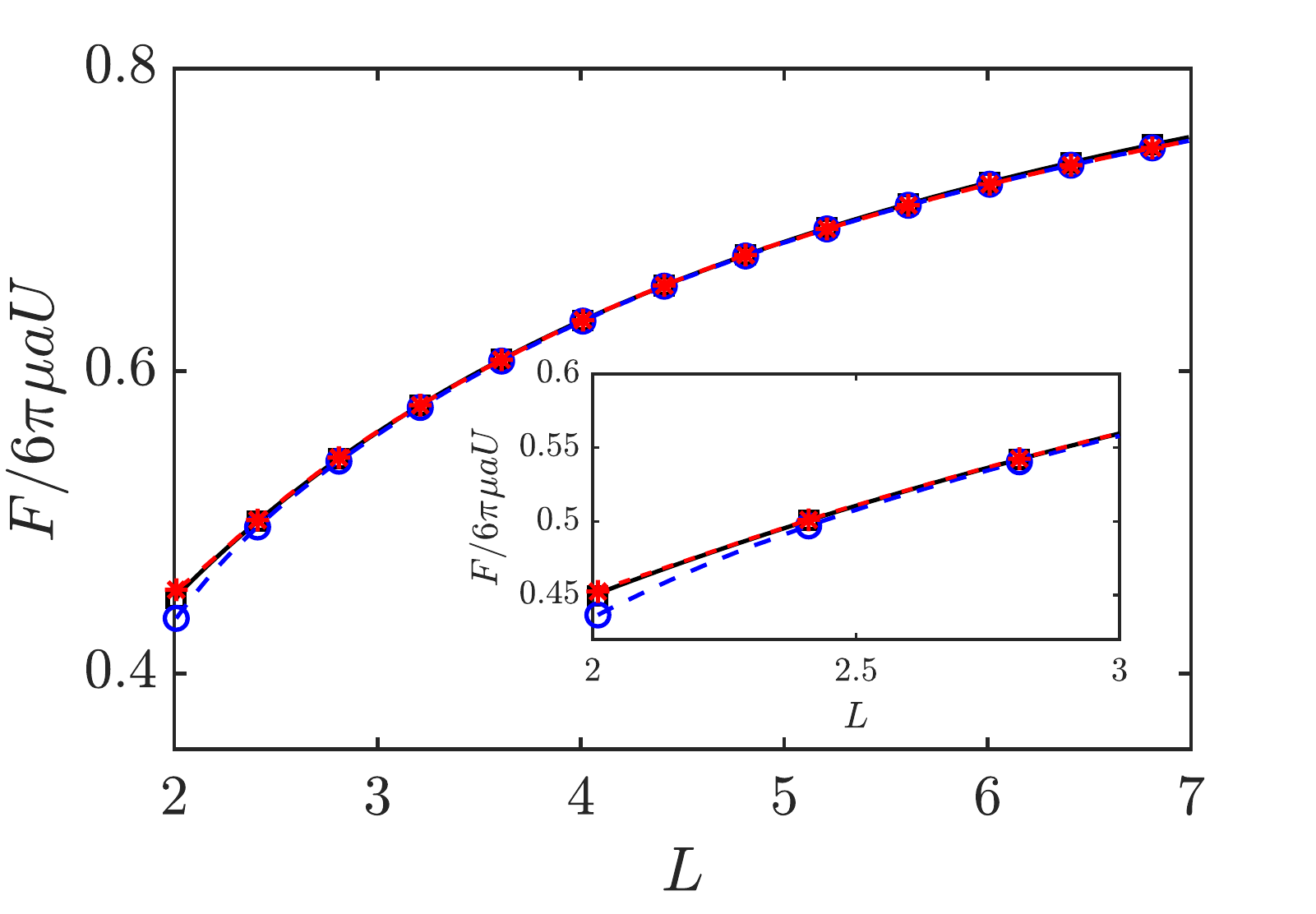}
\caption{Drag coefficient for the setup in $(a)$.}
\label{sfig:line_perp}
\end{subfigure}
\quad
\begin{subfigure}{0.45\textwidth}
\centering
\includegraphics[width=1.0\textwidth]{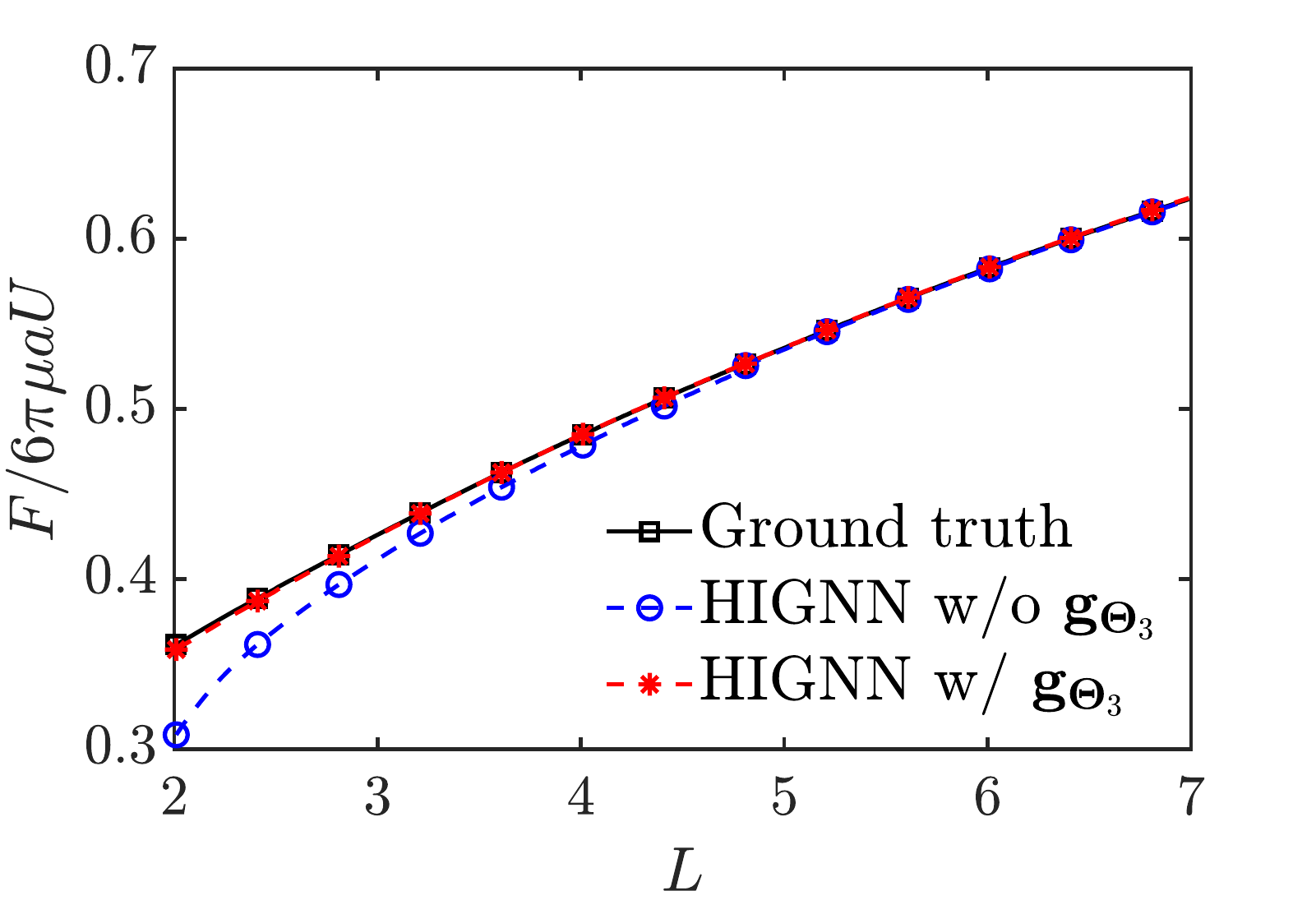}
\caption{Drag coefficient for the setup in $(b)$.}
\label{sfig:line_para}
\end{subfigure}
\caption{Drag coefficient ($ F/6\pi \mu a U$) predicted for the central particle in a five-particle chain with a uniform  center-to-center distance $L$.}
\label{fig:line}
\end{figure}

The trained HIGNN was first applied to simulating a benchmark problem \cite{Ganatos1978HI}, where a chain of particles evenly spaced with a center-to-center distance of $L$, subject to a uniform force perpendicular or parallel to the line connecting the particles' centers, as illustrated in Figs.~\ref{sfig:setup_perp} and \ref{sfig:setup_para}, respectively. Through this benchmark problem, we can validate the accuracy of the HIGNN predictions and examine the importance of three-body HI effects. 

Varying the distance between particles ($L$), the particle velocities predicted by the HIGNN are plotted in Figs.~\ref{sfig:line_perp} and \ref{sfig:line_para} and compared with the ground truth (SD simulation results). Without loss of generality, the velocity of the particle in the center was computed for comparison. When the three-body HI is included, i.e., the last term in Eq. \eqref{Eq:3body_GNN_relative}, the predictions by the HIGNN are found satisfactorily accurate for a wide range of distances between the particles, including when $L = 2.01 $, the particles are in close contact with a separation of only $0.01 a$. We hence demonstrate that the HIGNN is able to accurately account for both the long-range HI and near-field lubrication effects. When $L>5.0$, the contributions of three-body HI are found negligible, confirming the choice of $R_\text{cut} = 5$ set for the three-body cutoff as in the training and used henceforth in \S\ref{subsec:unbounded} for building the face connectivity in the graph when applying the HIGNN for predictions.

\begin{figure}
\centering
\centering
\includegraphics[width=0.6\textwidth]{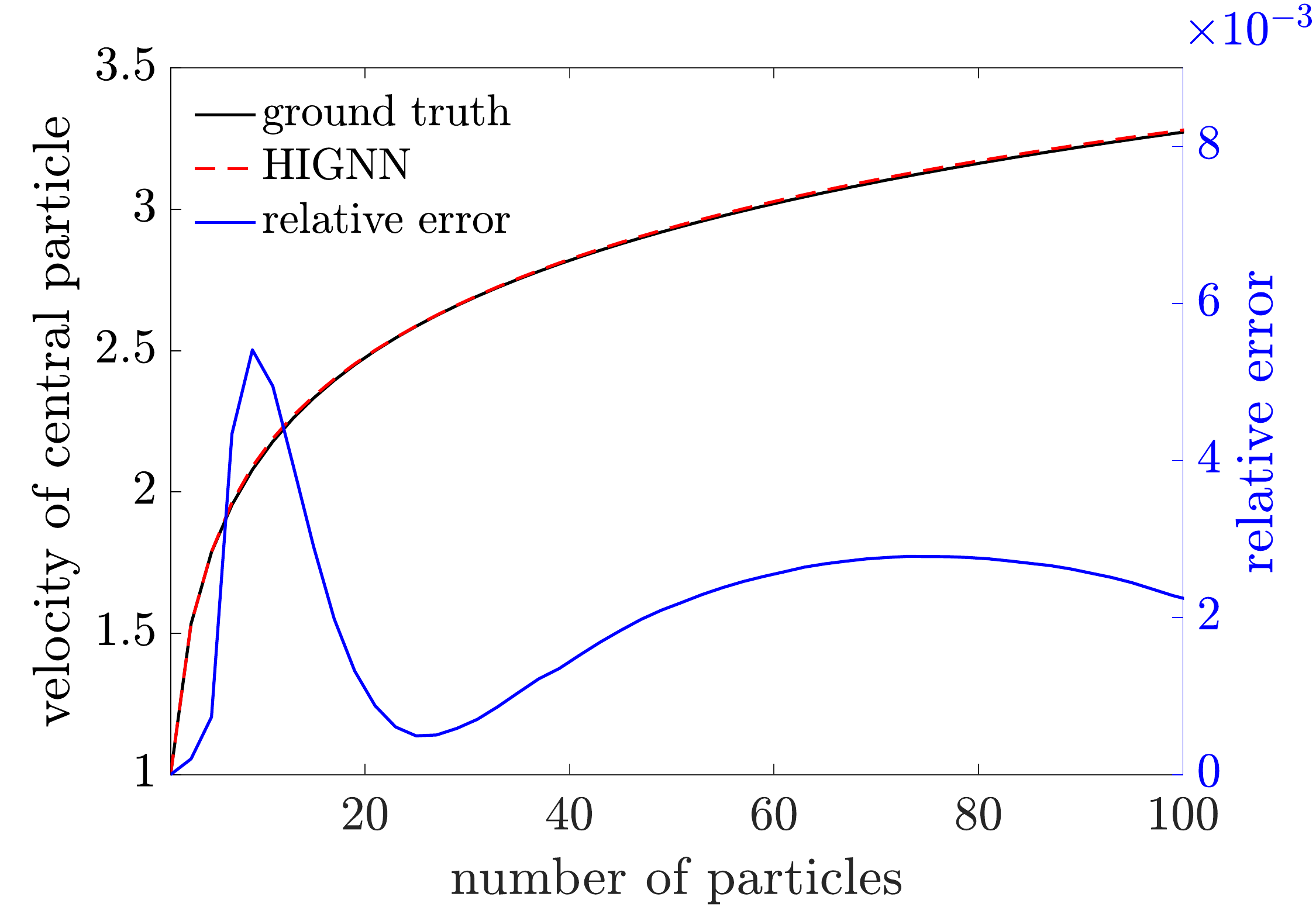}
\caption{The HIGNN's predictions on the velocity of the central particle in a $N$-particle chain with a uniform  center-to-center distance $L=3$ for different $N$ (total number of particles in the chain). Left $y$-axis: Velocity of the central particle. Right $y$-axis: Relative error compared with the ground truth.}
\label{fig:err_n}
\end{figure}

Furthermore, through this problem we also examined if the HIGNN's accuracy could deteriorate when applying it to simulating systems of larger numbers of particles. Following the setup in Fig.~\ref{sfig:setup_perp} with $L = 3$, we gradually increased the total number of particles. Fig.~\ref{fig:err_n} plots the velocity of the central particle predicted by the HIGNN and the relative error compared with the ground truth. Up to 100 particles, the relative errors lie below 0.25\% and do not indicate any systematic increase with increasing numbers of particles. 

\subsubsection{Long-time dynamic simulation} \label{subsec:long_time}
\begin{figure}
\centering
\begin{subfigure}{0.175\textwidth}
\centering
\includegraphics[width=1.0\textwidth]{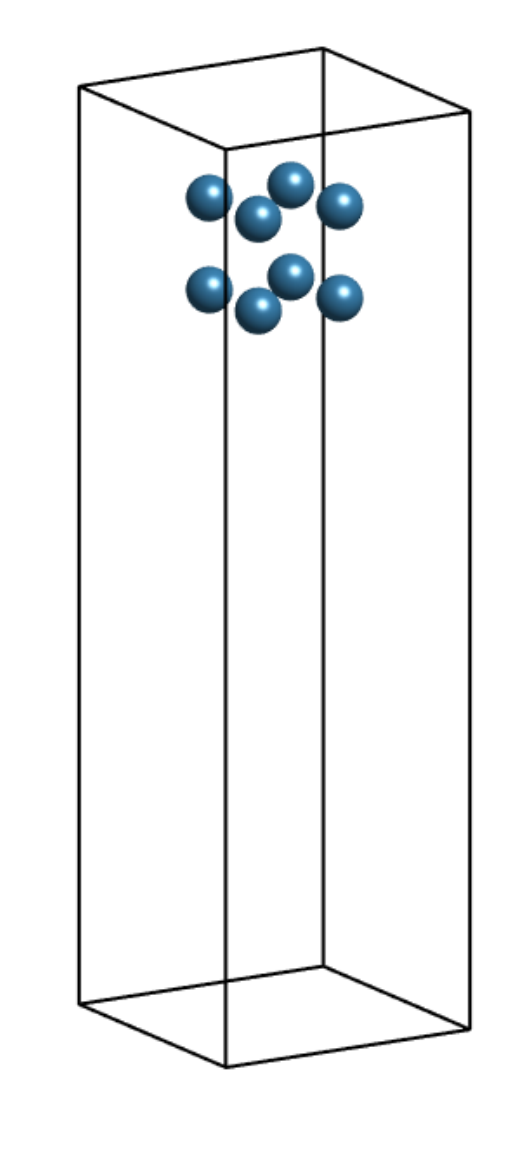}
\caption*{$t = 0$}
\label{sfig:cube_snap_L4_t0}
\end{subfigure}
\quad
\begin{subfigure}{0.175\textwidth}
\centering
\includegraphics[width=1.0\textwidth]{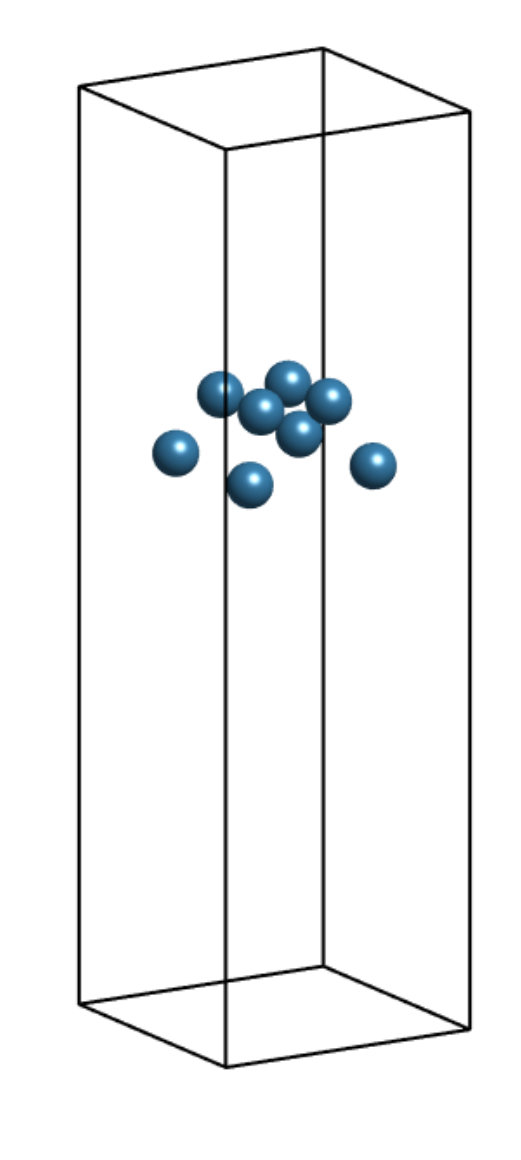}
\caption*{$t = 8.5$}
\label{sfig:cube_snap_L4_t1}
\end{subfigure}
\quad
\begin{subfigure}{0.175\textwidth}
\centering
\includegraphics[width=1.0\textwidth]{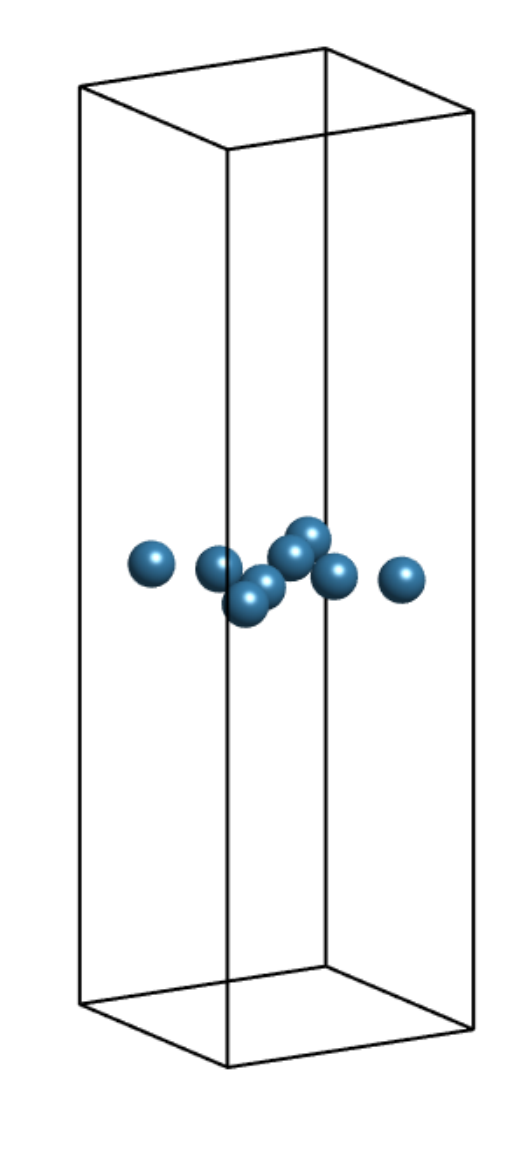}
\caption*{$t = 17$}
\label{sfig:cube_snap_L4_t2}
\end{subfigure}
\quad
\begin{subfigure}{0.175\textwidth}
\centering
\includegraphics[width=1.0\textwidth]{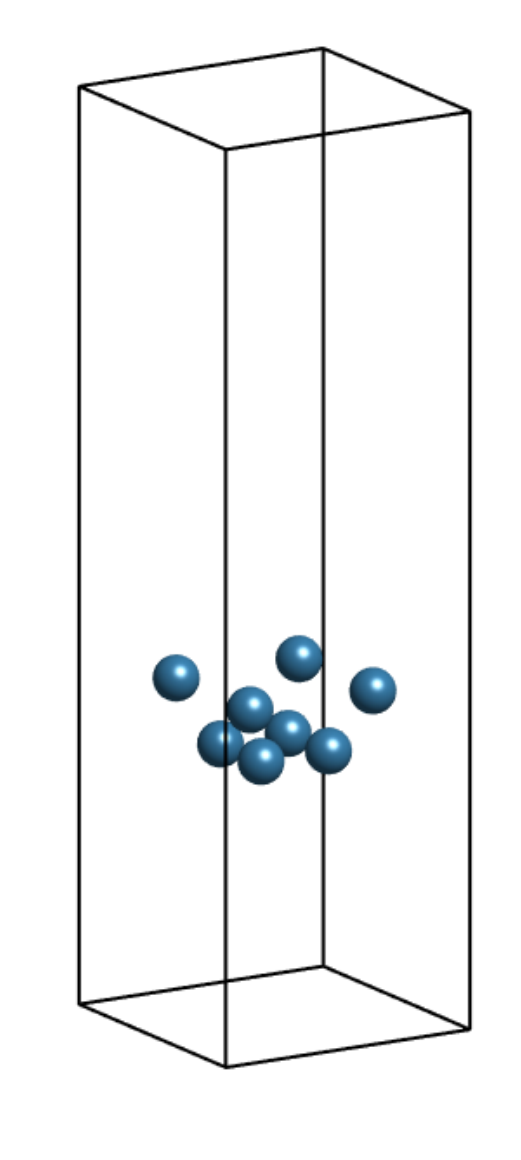}
\caption*{$t = 25.5 $}
\label{sfig:cube_snap_L4_t3}
\end{subfigure}
\quad
\begin{subfigure}{0.175\textwidth}
\centering
\includegraphics[width=1.0\textwidth]{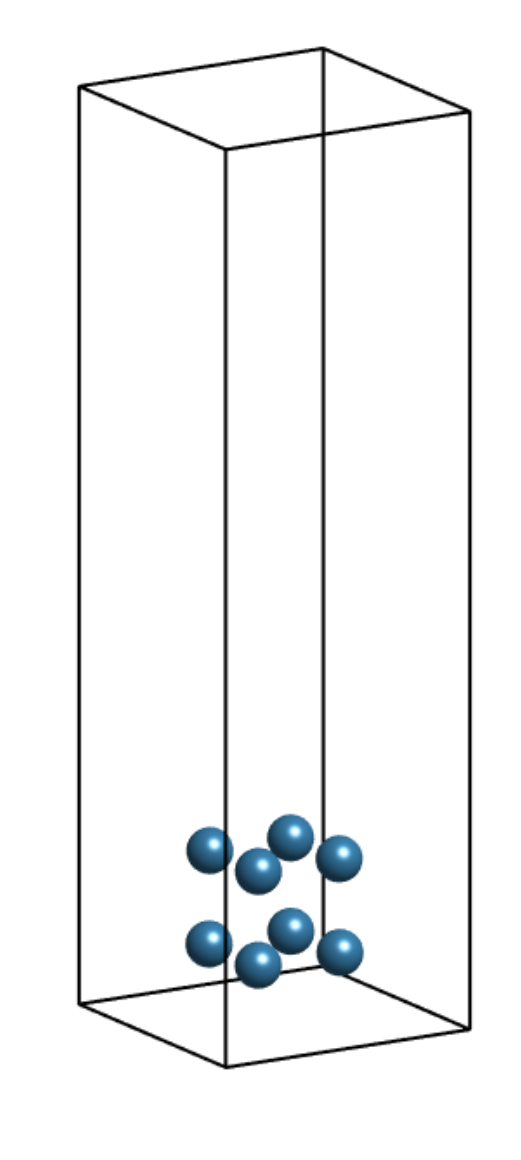}
\caption*{$t = 34 $}
\label{sfig:cube_snap_L4_t4}
\end{subfigure}
\quad
\begin{subfigure}{0.175\textwidth}
\centering
\includegraphics[width=1.0\textwidth]{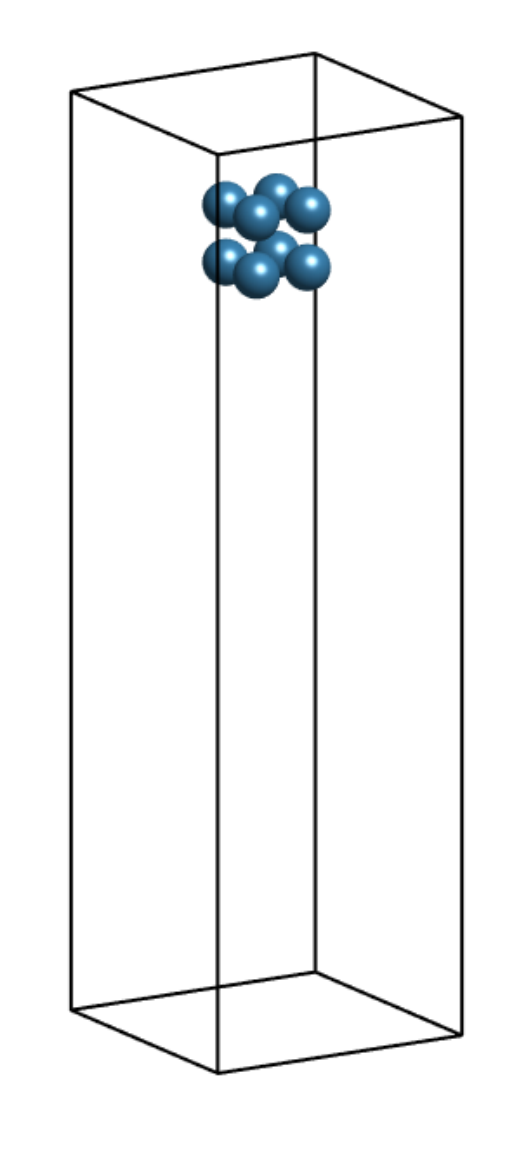}
\caption*{$t = 0$}
\label{sfig:cube_snap_L2_t0}
\end{subfigure}
\quad
\begin{subfigure}{0.175\textwidth}
\centering
\includegraphics[width=1.0\textwidth]{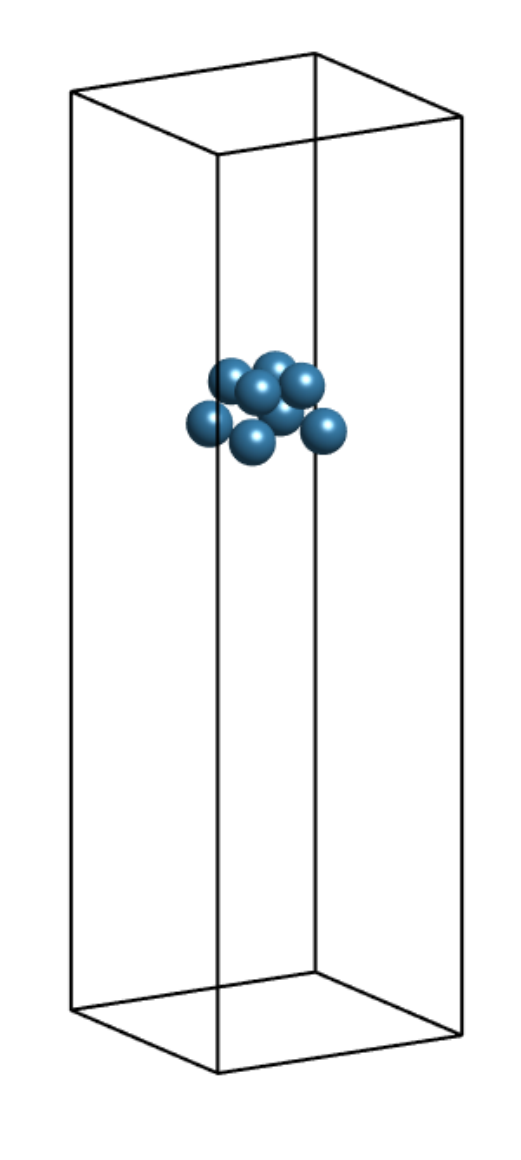}
\caption*{$t = 4.5$}
\label{sfig:cube_snap_L2_t1}
\end{subfigure}
\quad
\begin{subfigure}{0.175\textwidth}
\centering
\includegraphics[width=1.0\textwidth]{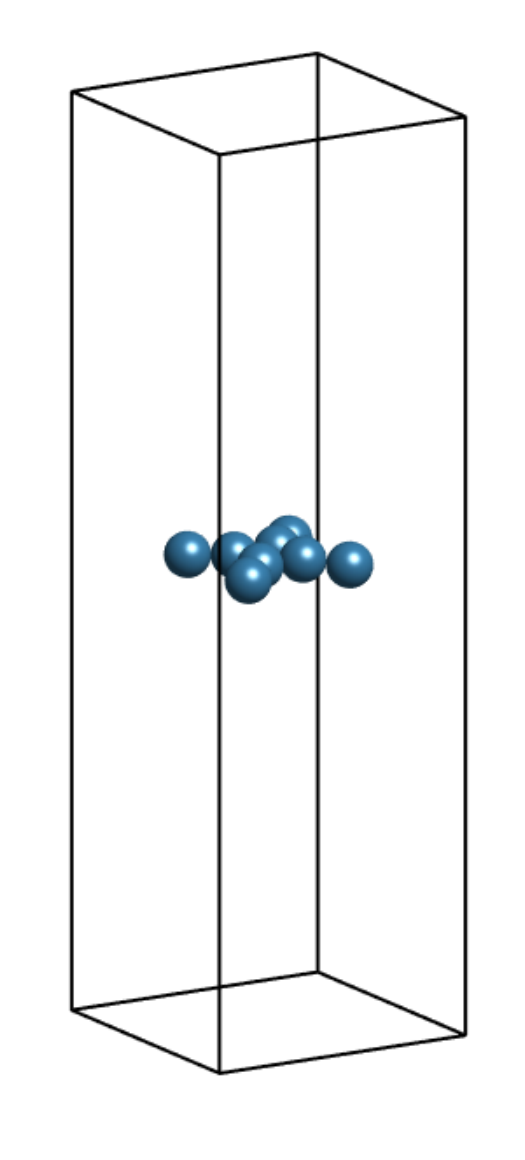}
\caption*{$t = 9$}
\label{sfig:cube_snap_L2_t2}
\end{subfigure}
\quad
\begin{subfigure}{0.175\textwidth}
\centering
\includegraphics[width=1.0\textwidth]{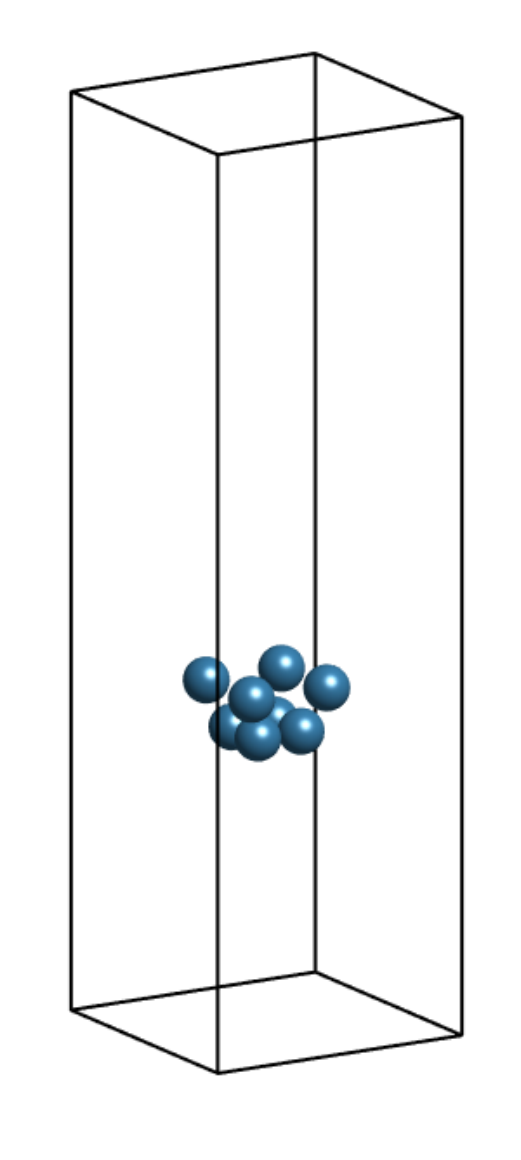}
\caption*{$t = 13.5 $}
\label{sfig:cube_snap_L2_t3}
\end{subfigure}
\quad
\begin{subfigure}{0.175\textwidth}
\centering
\includegraphics[width=1.0\textwidth]{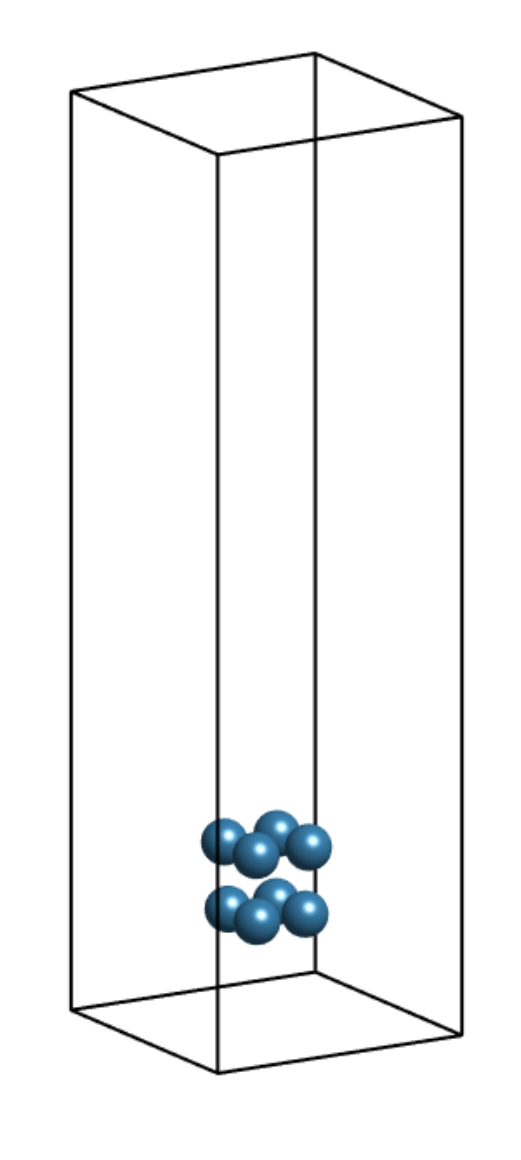}
\caption*{$t = 18 $}
\label{sfig:cube_snap_L2_t4}
\end{subfigure}
\caption{Snapshots of 8 particles sedimenting in an unbounded domain, predicted by the HIGNN for two different initial configurations with $L = 4$ (top) and $L = 2.5$ (bottom), respectively. The boxes are drawn as a guide to the eye.}
\label{fig:cube_snap}
\end{figure}

The trained HIGNN was next applied for a long-time dynamic simulation to predict the trajectories of sedimenting particles. Initially, eight particles are placed in a cubic lattice of length $L = 4$ and $L = 2.5$, as illustrated by the snapshots at $t = 0$ in Fig.~\ref{fig:cube_snap}, respectively, and each particle is exerted the same gravitational force along $-z$ direction, $\mathbf{F} = [0,0, -1]$. A temporal integrator, explicit Euler, was used for updating the particles' positions from their velocities with $\Delta t = 0.001$. At each time step, we applied the HIGNN for predicting all the particles' velocities. By such, we simulated the dynamics and predicted the trajectories of all eight particles. 

\begin{figure}
\centering
\begin{subfigure}{0.45\textwidth}
\centering
\includegraphics[width=1.0\textwidth]{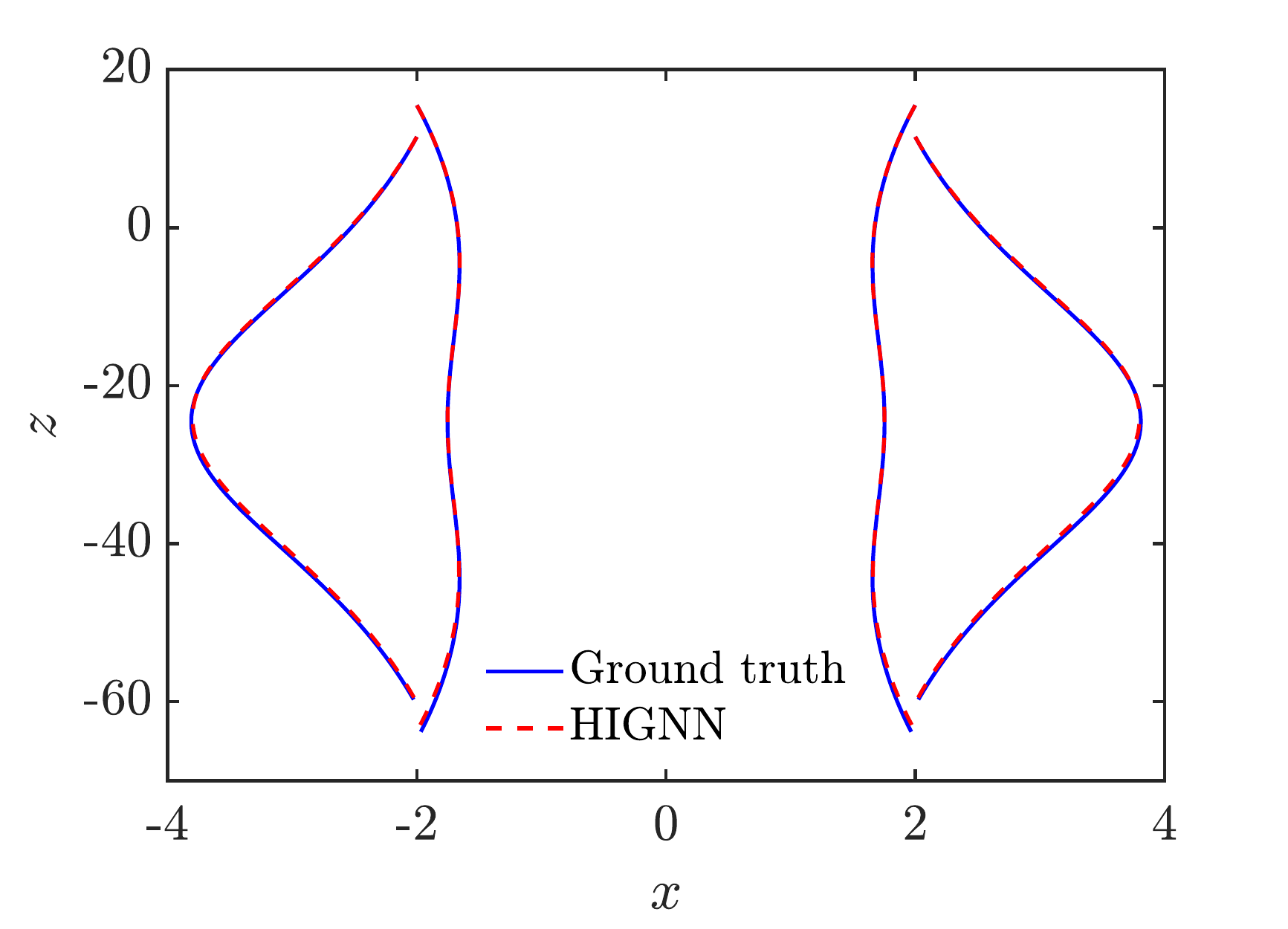}
\caption{}
\label{sfig:cube_traj1}
\end{subfigure}
\quad
\begin{subfigure}{0.45\textwidth}
\centering
\includegraphics[width=1.0\textwidth]{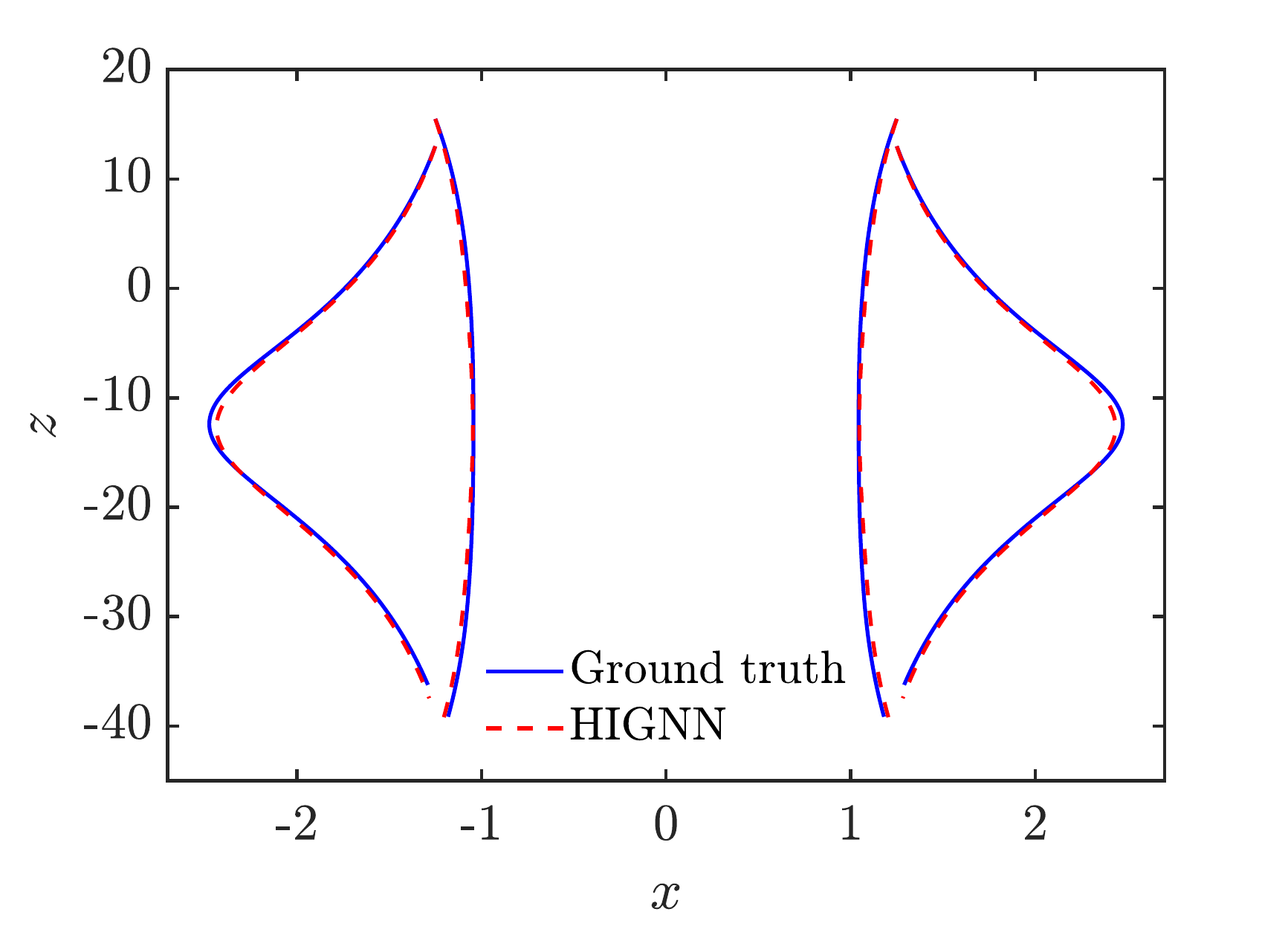}
\caption{}
\label{sfig:cube_traj2}
\end{subfigure}
\caption{Trajectories for 4 out of 8 particles sedimenting in an unbounded domain, predicted by the HIGNN for two different initial configurations with (a) $L = 4$ and (b) $L = 2.5$, respectively.}
\label{fig:cube}
\end{figure}

Fig.~\ref{fig:cube_snap} shows some snapshots of the eight sedimenting particles from our simulations. The predicted trajectories are presented in Figs.~\ref{sfig:cube_traj1} and \ref{sfig:cube_traj2}, where we only show the trajectories of four particles due to the symmetry on the $xz$- and $yz$-plane. The trajectories predicted by the HIGNN agree well with the ground truth in both cases. Note that in the case with $L = 2.5$, the particles' initial separation is only $0.5a$. We hence demonstrate the stability and accuracy of the HIGNN in long-time dynamic simulations.

\begin{figure}
\centering
\begin{subfigure}{0.27\textwidth}
\centering
\includegraphics[width=1.0\textwidth]{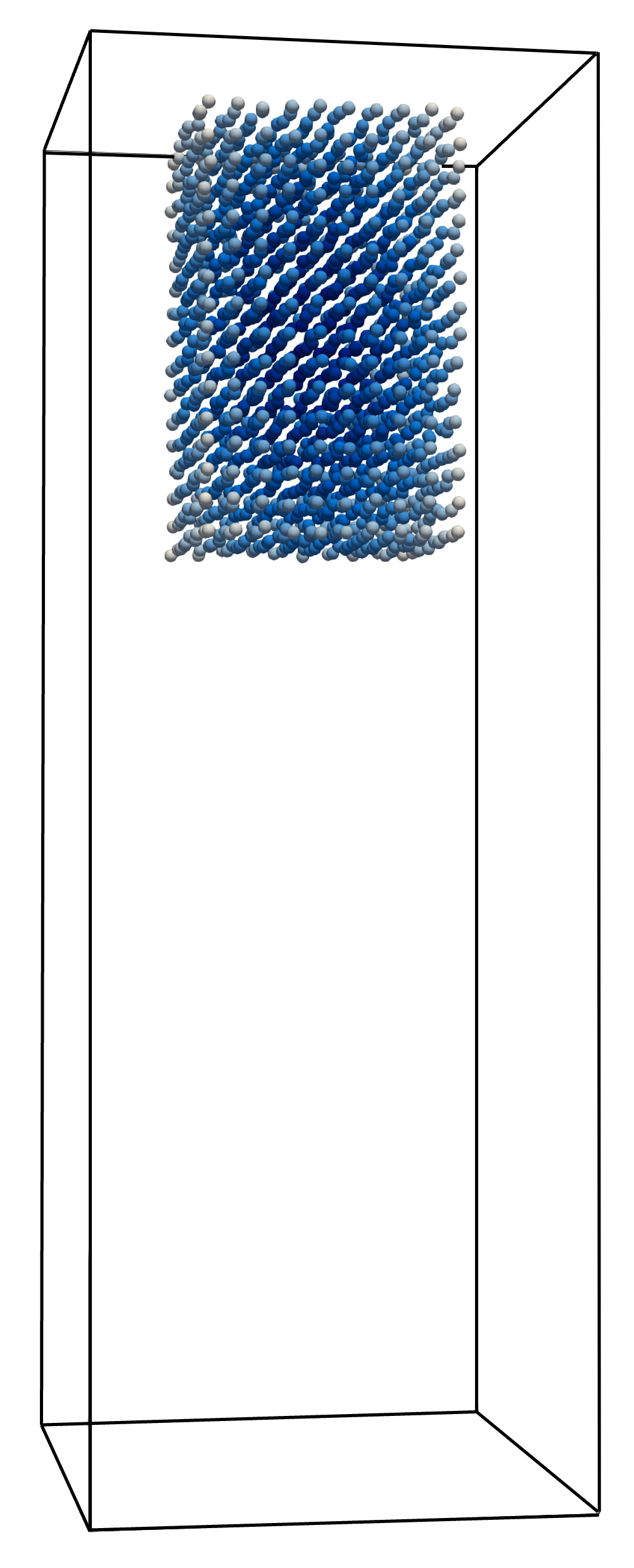}
\caption{$t = 0 $}
\label{sfig:1600_t0}
\end{subfigure}
\quad
\begin{subfigure}{0.27\textwidth}
\centering
\includegraphics[width=1.0\textwidth]{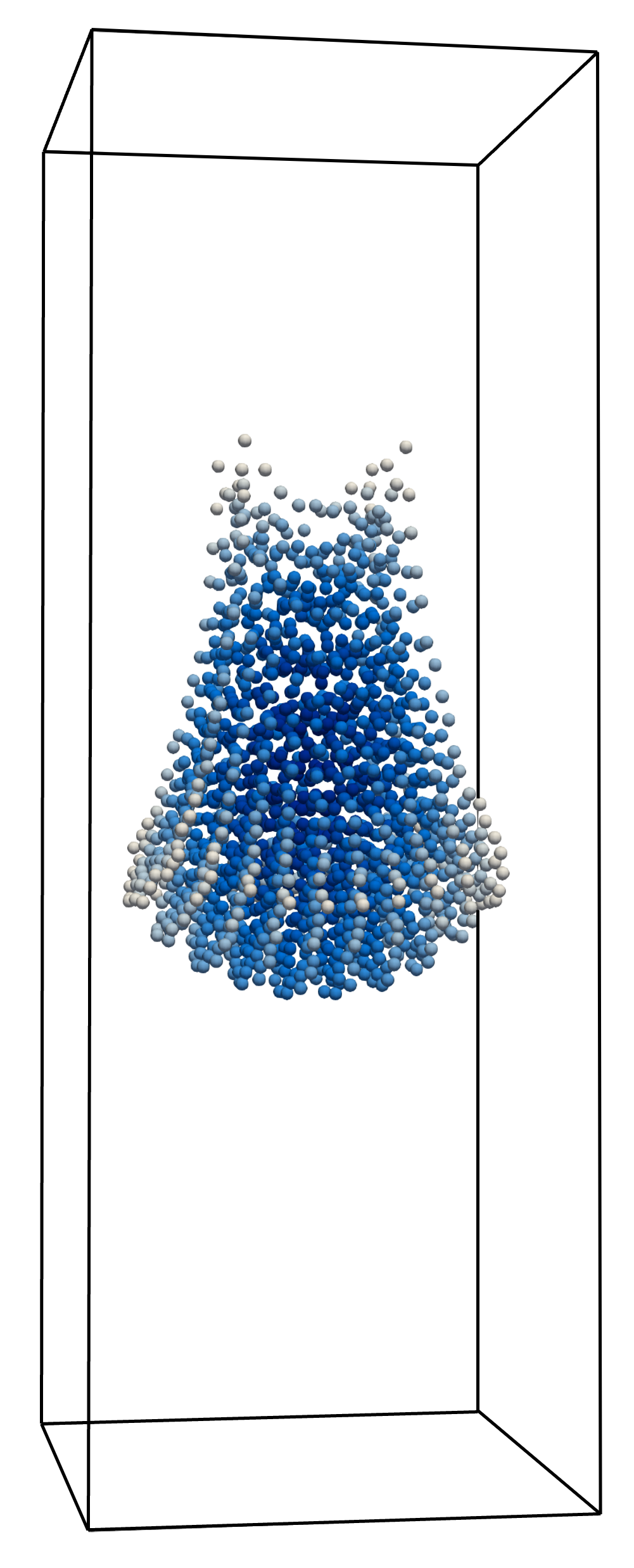}
\caption{$t = 1 $}
\label{sfig:1600_t1}
\end{subfigure}
\quad
\begin{subfigure}{0.34\textwidth}
\centering
\includegraphics[width=1.0\textwidth]{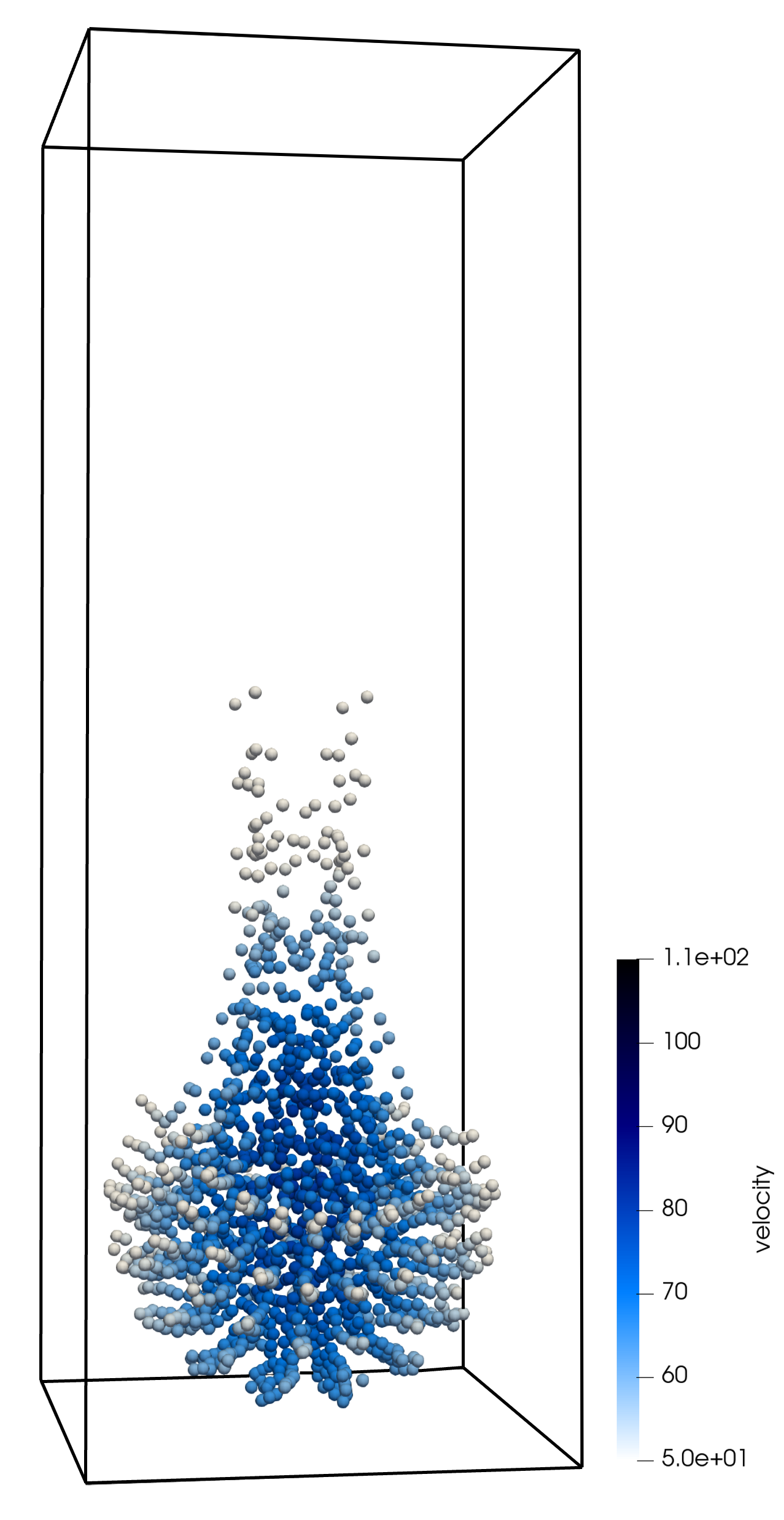}
\caption{$t = 2 $}
\label{sfig:1600_t2}
\end{subfigure}
\caption{Snapshots of 1,600 particles sedimenting in an unbounded domain, predicted by the HIGNN. The boxes are drawn as a guide to the eye. The color on the particles is correlated to the magnitudes of the particles' instantaneous velocities.}
\label{fig:1600_gravity}
\end{figure}

Further, we increased the number of particles to 1,600 and simulated their sedimenting dynamics. Initially, the particles are placed in a cubic lattice of length $L=3$, as illustrated in Fig.~\ref{sfig:1600_t0}. Some snapshots from the simulation are presented in Fig.~\ref{fig:1600_gravity}. Here, we outline the computational cost for this simulation: each time step spent 0.16 sec (wall time), and the simulation ran for 2000 time steps and cost $\sim$5.3 min, using the hardware stated in the beginning of \S\ref{sec:results}.

\subsubsection{Large-scale suspension} \label{subsec:unbounded_largescale}
\begin{figure}
\centering
\includegraphics[width=1.0\textwidth]{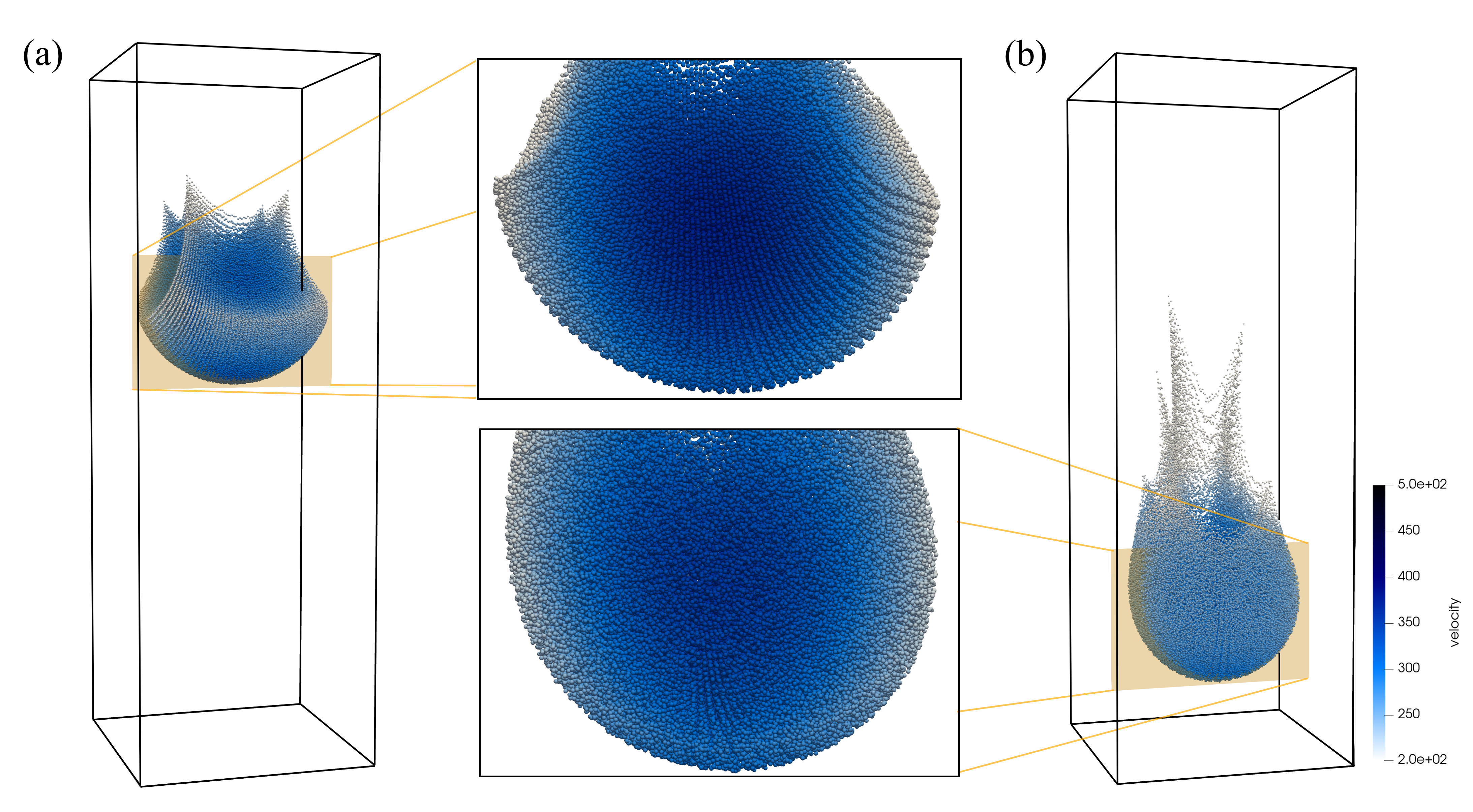}
\caption{Snapshots of 100,000 particles sedimenting in an unbounded domain, predicted by the HIGNN at two different times (a) $t=0.2$ and (b) $t=0.4$. The boxes are drawn as a guide to the eye. The color on the particles is correlated to the magnitudes of the particles' instantaneous velocities.}
\label{fig:100000_gravity}
\end{figure}

Finally, the trained HIGNN was used to simulate a very large-scale suspension consisting of 100,000 particles. As in the prior section (\S\ref{subsec:long_time}), the particles in suspension are subject to sedimentation, with the initial configuration same as in the case of 1,600 particles. For this simulation, the time step was set as $ \Delta t = 0.01 $ in the explicit Euler time integrator. In Fig.~\ref{fig:100000_gravity}, we present some snapshots of 100,000 particles sedimenting in an unbounded domain, predicted by the HIGNN. We evaluated the computational cost using two hardware settings. First, the simulation was performed on a workstation with 4 NVIDIA RTX 2080 Ti GPUs and one Intel(R) Xeon(R) Silver 4110 CPU @ 2.10GHz. It spent 130.3 sec to predict the velocities of all 100,000 particles at each time step, and the entire simulation ran for 40 time steps and cost $\sim$86.8 min. Second, the same simulation was executed on another workstation with 6 NVIDIA RTX 2080 Ti GPUs and one Intel(R) Xeon(R) Silver 4110 CPU @ 2.10GHz. It spent 86.5 sec to predict the velocities of all 100,000 particles at each time step, and the entire simulation ran for 40 time steps and cost $\sim$57.6 min. 

Here, to use multiple GPUs, we evenly split the graph into $n_G$ subgraphs, where $n_G$ denotes the number of GPUs, and distributed each subgraph into a GPU. Each subgraph has all $ 100,000 $ vertices but only the edges and faces pointing to a subset of vertices (consisting of $ 100,000/n_G $ vertices). For example, if $n_G = 2$, the subgraph in the first GPU only includes the edges and faces targeting to the first 50,000 vertices (particles), and the subgraph in the second GPU accounts for the remaining edges and faces. As such, the HIGNN requires only one all-to-all communication of the particles' coordinates (vertices' feature) when evaluating the particles' velocities. All the following calculation and aggregation of edge and face information for different target vertices can be performed independently and simultaneously in each GPU. Thus, the cost for communication between GPUs is minor. Through this case, we assess the computational efficiency of the HIGNN for simulating a large-scale particulate suspension and also demonstrate the ease of allocating multiple GPUs for executing the HIGNN inference with parallel efficiency.  


\section{Discussion}\label{sec:conclusion}
We have introduced the HIGNN framework to enable fast simulations of particulate suspensions. By extending edge convolutional GNN and introducing faces in addition to vertices and edges in a graph, we can retain up to three-body HI in the surrogate modeling of the mobility tensor. By further introducing higher-order connectivity into the graph, the HIGNN framework is extendable to include any $m$-body HI ($m>3$) for achieving high-order accuracy. Training the HIGNN only requires the data for three particles sampled with varying spatial configurations in the domain of interest. In general, if the desired accuracy requires the surrogate modeling to retain up to $m$-body HI, the training would need the data for $m$ particles sampled with different spatial configurations. Thus, the training cost can be maintained very low. Once the HIGNN is trained, it permits fast predictions of particles' velocities and is transferable to suspensions of different numbers or concentrations of particles in the same domain and to any types and magnitudes of external forces. We have demonstrated that the HIGNN is accurate in capturing both the far-field HI and near-field lubrication effects. And the accuracy can be preserved as the number of particles increases.

When simulating large-scale suspensions, the HIGNN framework permits to call multiple GPUs for parallel computing. The workload can be evenly distributed into each GPU with only one all-to-all communication between GPUs required. All the convolution operations and other calculations can be executed in each GPU independently and simultaneously. This feature may not be seen in other methodologies; for example, in accelerated SD \cite{Ouaknin2021ParaASD,WANG2016ASD2} and fast SD \cite{Fiore2019FSD}, each single iteration step of the linear solver (for inversion of matrix) calls for one all-to-all communication between processors to evaluate the far-field part of HI, resulting in many all-to-all communications (proportional to the number of iteration steps) required for computing the particles' velocities. With increasing numbers of processors, more significant portions of its execution spent in communications and synchronizations could result in a bottleneck slowing down parallel efficiency.


For non-spherical particles, $\textbf{M}$ is a function of both $\textbf{X}$ and $\boldsymbol{\theta}$, the particles' centroid positions and orientations. In turn, the surrogate models for each $m$-body HI should be functions of $(\textbf{X},\boldsymbol{\theta})$. The data structure represented by a graph permits to include general properties of particles in the feature vector of each vertex as the input of GNN, including $\textbf{X}_i$ and $\boldsymbol{\theta}_i$. Therefore, the HIGNN framework introduced in this work has the potential to be generalized to particles of anisotropic shapes.


\section*{Declaration of Competing Interest}

The authors declare that they have no known competing financial interests or personal relationships that could have appeared to influence the work reported in this paper.

\section*{Acknowledgement}

Z. M. and W. P. were supported by the Defense Established Program to Stimulate Competitive Research (DEPSCoR) Grant No. FA9550-20-1-0072. Z. Y. was supported by the University of Wisconsin - Madison Office of the Vice Chancellor for Research and Graduate Education with funding from the Wisconsin Alumni Research Foundation. 

\appendix
\section{Derivation of Eq. \eqref{Eq:theo_basis_trun_final}} 
\label{sec:append}
The grand mobility $\mathbf{M}$ can be decomposed into:
\begin{equation}\label{Eq:M}
\mathbf{M}
= 
\begin{bmatrix}
    \mathbf{M}^{TT} & \mathbf{M}^{TR} \\
    \mathbf{M}^{RT} & \mathbf{M}^{RR} 
\end{bmatrix}
\end{equation}
where $\mathbf{M}^{TT}$ is the translational mobility tensor, $\mathbf{M}^{RR}$ is the rotational mobility tensor, and the matrices $\mathbf{M}^{TR}$ and $\mathbf{M}^{RT}$ couples the translational and rotational motions. Mazur and Saarloos \cite{Mazur1982ManysphereHI} have provided general expressions for each type mobility tensor for spherical particles in unbounded domain, based on the method of induced forces \cite{ALBANO1975InducedForce1, MAZUR1978InducedForce2} with the structure of an infinite series of reflections from particles, as: 
\begin{subequations}\label{Eq:M_series}
\begin{equation}\label{Eq:M_TT_series}
\begin{split}
    \mathbf{M}^{TT}_{ij} = & (6\pi \mu a)^{-1} [\mathbf{1} \delta_{ij} + \mathbf{A}^{(1,1)}_{ij} 
    +  \sum\limits_{\gamma = 1}^\infty \sum\limits_{m_1 = 2}^\infty{}^{'} \sum\limits_{m_2 = 2}^{\infty}{}^{'}  \cdots \sum\limits_{m_\gamma = 2}^\infty{}^{'} \sum\limits_{\substack{j_1 = 1\\j_1\neq i}}^N \sum\limits_{\substack{j_2 = 1\\j_2\neq j_1}}^N \cdots  \\
    \times & \sum\limits_{\substack{j_\gamma = 1\\j_\gamma\neq j_{\gamma-1}, j}}^N \mathbf{A}^{(1,m_1)}_{ij_1} \odot \mathbf{B}^{(m_1, m_1)^{-1}} \odot \mathbf{A}^{(m_1,m_2)}_{j_1 j_2}  
    \odot  \cdots \odot \mathbf{B}^{(m_\gamma, m_\gamma)^{-1}} \odot \mathbf{A}^{(m_\gamma,1)}_{j_\gamma j} ] \;, 
\end{split}
\end{equation}

\begin{equation}\label{Eq:M_RR_series}
\begin{split}
    \mathbf{M}^{RR}_{ij} = & (8\pi \mu a^3)^{-1} [\mathbf{1} \delta_{ij} - \frac{1}{3} \boldsymbol{\epsilon} : \mathbf{A}^{(2\text{a} ,2\text{a})}_{ij} :\boldsymbol{\epsilon}
    -  \frac{1}{3} \sum\limits_{\gamma = 1}^\infty \sum\limits_{m_1 = 2}^\infty{}^{'} \sum\limits_{m_2 = 2}^{\infty}{}^{'}  \cdots \sum\limits_{m_\gamma = 2}^\infty{}^{'} \sum\limits_{\substack{j_1 = 1\\j_1\neq i}}^N \sum\limits_{\substack{j_2 = 1\\j_2\neq j_1}}^N \cdots  \\
    \times & \sum\limits_{\substack{j_\gamma = 1\\j_\gamma\neq j_{\gamma-1}, j}}^N \boldsymbol{\epsilon} : \mathbf{A}^{(2\text{a},m_1)}_{ij_1} \odot \mathbf{B}^{(m_1, m_1)^{-1}}  
    \odot  \mathbf{A}^{(m_1,m_2)}_{j_1 j_2}    \odot \cdots \odot \mathbf{B}^{(m_\gamma, m_\gamma)^{-1}} \odot \mathbf{A}^{(m_\gamma,2\text{a})}_{j_\gamma j} :\boldsymbol{\epsilon} ] \;,
\end{split}
\end{equation}

\begin{equation}\label{Eq:M_RT_series}
\begin{split}
    \mathbf{M}^{RT}_{ij} = & (12\pi \mu a^2)^{-1} [ \boldsymbol{\epsilon} : \mathbf{A}^{(2\text{a} , 1)}_{ij} 
    + \sum\limits_{\gamma = 1}^\infty \sum\limits_{m_1 = 2}^\infty{}^{'} \sum\limits_{m_2 = 2}^{\infty}{}^{'}  \cdots \sum\limits_{m_\gamma = 2}^\infty{}^{'} \sum\limits_{\substack{j_1 = 1\\j_1\neq i}}^N \sum\limits_{\substack{j_2 = 1\\j_2\neq j_1}}^N \cdots \\
    \times & \sum\limits_{\substack{j_\gamma = 1\\j_\gamma\neq j_{\gamma-1}, j}}^N \boldsymbol{\epsilon} : \mathbf{A}^{(2\text{a},m_1)}_{ij_1} \odot \mathbf{B}^{(m_1, m_1)^{-1}} 
    \odot  \mathbf{A}^{(m_1,m_2)}_{j_1 j_2}    \odot \cdots \odot \mathbf{B}^{(m_\gamma, m_\gamma)^{-1}} \odot \mathbf{A}^{(m_\gamma, 1)}_{j_\gamma j} ] \;, 
\end{split}
\end{equation}

\begin{equation}\label{Eq:M_TR_series}
\begin{split}
    \mathbf{M}^{TR}_{ij} = & (12\pi \mu a^2)^{-1} [ \mathbf{A}^{(1, 2\text{a} )}_{ij} :\boldsymbol{\epsilon} 
    +  \sum\limits_{\gamma = 1}^\infty \sum\limits_{m_1 = 2}^\infty{}^{'} \sum\limits_{m_2 = 2}^{\infty}{}^{'}  \cdots \sum\limits_{m_\gamma = 2}^\infty{}^{'} \sum\limits_{\substack{j_1 = 1\\j_1\neq i}}^N \sum\limits_{\substack{j_2 = 1\\j_2\neq j_1}}^N \cdots \\
    \times & \sum\limits_{\substack{j_\gamma = 1\\j_\gamma\neq j_{\gamma-1}, j}}^N  \mathbf{A}^{(1 , m_1)}_{ij_1} \odot \mathbf{B}^{(m_1, m_1)^{-1}} 
    \odot \mathbf{A}^{(m_1,m_2)}_{j_1 j_2}    \odot \cdots \odot \mathbf{B}^{(m_\gamma, m_\gamma)^{-1}} \odot \mathbf{A}^{(m_\gamma, 2\text{a})}_{j_\gamma j} : \boldsymbol{\epsilon} ] \;,
\end{split}
\end{equation}
\end{subequations}
where $\mathbf{M}_{ij} \in \mathbb{R}^{3 \times 3} $ is the submatrix of grand mobility $\mathbf{M} $; $ \delta_{ij}$ is the Kronecker delta function; and $ \boldsymbol{\epsilon} $ is the Levi-Civita tensor. Here, $ \mathbf{A}^{(n, m)}_{ij} $, named as connector, is a tensor of rank $n + m$ and characterizes the HI between a force multipole of order $n$ on particle $i$ and a multipole of order $m$ on particle $j$. The tensor $\mathbf{B}^{(m, m)^{-1}}$ is the generalized inverse of a tensor $ \mathbf{B}^{(m, m)} $ of rank $2m$. The notation $ \mathbf{A}^{(n , m)}_{ij} \odot \mathbf{B}^{(m, m)^{-1}} $ denotes a full contraction over the last $m$ indices of the first tensor and the first $m$ indices of the second tensor. The connectors $ \mathbf{A}^{(n, 2 )}_{ij} $ and $\mathbf{A}^{( 2 , m)}_{ij}$ have been decomposed as: 
\begin{equation}
    \mathbf{A}^{(n, 2 )}_{ij} = \mathbf{A}^{(n, 2\text{a})}_{ij} + \mathbf{A}^{(n, 2\text{s})}_{ij}, \ \ \mathbf{A}^{(2, m)}_{ij} = \mathbf{A}^{(2\text{a}, m)}_{ij} + \mathbf{A}^{(2\text{s}, m)}_{ij} \;, 
\end{equation}
where the superscripts $2\text{a}$ and $2\text{s}$ denote the asymmetric and traceless symmetric parts, respectively. The notation $\sum\limits_{m = 2}^\infty{}^{'}$ denotes a summation over all integers $m \geq 2$, with the proviso that for $m = 2$ only the connectors $\mathbf{A}^{(n, 2\text{s})}_{ij}$ are included in the summation, e.g.
\begin{equation}
    \sum\limits_{m = 2}^\infty{}^{'} \mathbf{A}^{(1, m)}_{ij} \odot \mathbf{B}^{(m, m)^{-1}} \odot \mathbf{A}^{(m,1)}_{j i} = \mathbf{A}^{(1, 2\text{s})}_{ij} \odot \mathbf{B}^{(2\text{s}, 2\text{s})^{-1}} \odot \mathbf{A}^{(2\text{s}, 1)}_{j i} +  \sum\limits_{m = 3}^\infty \mathbf{A}^{(1, m)}_{ij} \odot \mathbf{B}^{(m, m)^{-1}} \odot \mathbf{A}^{(m,1)}_{j i} \;.
\end{equation}

In this work, we focus on translational motions of spherical particles. Thus, the following derivations are for the two mobility tensors contributing to the translational velocity, i.e., $\mathbf{M}^{TT}$ and $\mathbf{M}^{TR}$. To explain the derivation process, we take $\mathbf{M}^{TT}_{ij}$ as an example. In Eq.~\eqref{Eq:M_TT_series}, the first term $ \mathbf{1}\delta_{ij}$ is from the Stokes law, which connects the velocity of particle $i$ to the force exerted on it. The second term $ \mathbf{A}^{(1,1)}_{ij} $ corresponds to the most dominant two-body interaction, i.e., the interaction from the force (the first-order multipole) on particle $j$ to the velocity of particle $i$. In the last term, each summation represents an effect from particle $j$ to $i$ due to the reflection of induced force multipoles, which might pass other particles or only reflect between $i$ and $j$. Here, to be consistent with the main text, we classify the terms with only reflections between $i$ and $j$ as two-body interactions, and the terms with the induced force multipoles on other particles as many-body interactions. For instance, $\mathbf{A}^{(1, 2\text{s})}_{ij} \odot \mathbf{B}^{(2\text{s}, 2\text{s})^{-1}} \odot \mathbf{A}^{(2\text{s}, 1)}_{j i}$ is a two-body interaction term, $\mathbf{A}^{(1, 2\text{s})}_{ik} \odot \mathbf{B}^{(2\text{s}, 2\text{s})^{-1}} \odot \mathbf{A}^{(2\text{s}, 1)}_{k j}$ is a three-body interaction term, and so on so forth. Each $\mathbf{A}^{(n, m)}_{ij}$ $(i\neq j)$ is a function of $ \mathbf{r}_{ij} := \mathbf{X}_j - \mathbf{X}_i $ of order $ (a/|\mathbf{r}_{ij}|)^{n+m-1} $ \cite{Mazur1982ManysphereHI}. 

Thus, we can separate each $m$-body interaction term. By retaining only the two-body and three-body interaction terms and separately considering $i = j$ or $i \neq j$, we can rewrite Eqs.~\eqref{Eq:M_TT_series} and \eqref{Eq:M_TR_series} as:
\begin{subequations}\label{Eq:Mii_Mij}
\begin{equation}\label{Eq:M_TT_ii}
\begin{split}
    \mathbf{M}^{TT}_{ii} = & \frac{1}{6\pi \mu a} [\mathbf{1}
    + \sum\limits_{\substack{ {l = 1} \\ l\neq i}}^N \sum\limits_{\gamma = 1}^\infty \sum\limits_{m_1 = 2}^\infty{}^{'}  \cdots \sum\limits_{m_\gamma = 2}^\infty{}^{'} \sum\limits_{\substack{j_1, j_2, \dots, j_\gamma \in \{ i, l \}\\j_1\neq i, j_2 \neq j_1, \dots, j_\gamma \neq i}}  \mathbf{A}^{(1,m_1)}_{ij_1} \odot \mathbf{B}^{(m_1, m_1)^{-1}}
    \odot  \cdots  \odot \mathbf{A}^{(m_\gamma,1)}_{j_\gamma i} \\
    + & \sum\limits_{\substack{ {l = 1} \\ l\neq i}}^N \sum\limits_{\substack{k = 1\\k\neq i \text{ or } l }}^N \sum\limits_{\gamma = 1}^\infty \sum\limits_{m_1 = 2}^\infty{}^{'}   \cdots \sum\limits_{m_\gamma = 2}^\infty{}^{'} \sum\limits_{\substack{j_1, j_2, \dots, j_\gamma \in \{ i, l, k \}\\j_1\neq i, j_2 \neq j_1, \dots, j_\gamma \neq i}} \mathbf{A}^{(1,m_1)}_{ij_1} \odot \mathbf{B}^{(m_1, m_1)^{-1}} \odot  \cdots  \odot \mathbf{A}^{(m_\gamma,1)}_{j_\gamma i} ] + O(r^{-10}) \mathbf{J} \;, 
\end{split}
\end{equation}

\begin{equation}\label{Eq:M_TT_ij}
\begin{split}
    \mathbf{M}^{TT}_{ij} = & \frac{1}{6\pi \mu a} [ \mathbf{A}^{(1,1)}_{ij} 
    + \sum\limits_{\gamma = 1}^\infty \sum\limits_{m_1 = 2}^\infty{}^{'}  \cdots \sum\limits_{m_\gamma = 2}^\infty{}^{'} \sum\limits_{\substack{j_1, j_2, \dots, j_\gamma \in \{ i, j \}\\j_1\neq i, j_2 \neq j_1, \dots, j_\gamma \neq i}}  \mathbf{A}^{(1,m_1)}_{ij_1} \odot \mathbf{B}^{(m_1, m_1)^{-1}}
    \odot  \cdots  \odot \mathbf{A}^{(m_\gamma,1)}_{j_\gamma j} \\
    + & \sum\limits_{\substack{k = 1\\k\neq i \text{ or } j }}^N \sum\limits_{\gamma = 1}^\infty \sum\limits_{m_1 = 2}^\infty{}^{'}   \cdots \sum\limits_{m_\gamma = 2}^\infty{}^{'} \sum\limits_{\substack{j_1, j_2, \dots, j_\gamma \in \{ i, j, k \}\\j_1\neq i, j_2 \neq j_1, \dots, j_\gamma \neq j}} \mathbf{A}^{(1,m_1)}_{ij_1} \odot \mathbf{B}^{(m_1, m_1)^{-1}} \odot  \cdots  \odot \mathbf{A}^{(m_\gamma,1)}_{j_\gamma j} ] + O(r^{-7}) \mathbf{J} \;, 
\end{split}
\end{equation}

\begin{equation}\label{Eq:M_TR_ii}
\begin{split}
    \mathbf{M}^{TR}_{ii} = & \frac{1}{12\pi \mu a^2} [
    \sum\limits_{\substack{ {l = 1} \\ l\neq i}}^N \sum\limits_{\gamma = 1}^\infty \sum\limits_{m_1 = 2}^\infty{}^{'}  \cdots \sum\limits_{m_\gamma = 2}^\infty{}^{'} \sum\limits_{\substack{j_1, j_2, \dots, j_\gamma \in \{ i, l \}\\j_1\neq i, j_2 \neq j_1, \dots, j_\gamma \neq i}} \mathbf{A}^{(1,m_1)}_{ij_1} \odot \mathbf{B}^{(m_1, m_1)^{-1}}
    \odot  \cdots  \odot \mathbf{A}^{(m_\gamma, 2\text{a})}_{j_\gamma i} : \boldsymbol{\epsilon} \\
    + & \sum\limits_{\substack{ {l = 1} \\ l\neq i}}^N \sum\limits_{\substack{k = 1\\k\neq i \text{ or } l }}^N \sum\limits_{\gamma = 1}^\infty \sum\limits_{m_1 = 2}^\infty{}^{'}   \cdots \sum\limits_{m_\gamma = 2}^\infty{}^{'} \sum\limits_{\substack{j_1, j_2, \dots, j_\gamma \in \{ i, l, k \}\\j_1\neq i, j_2 \neq j_1, \dots, j_\gamma \neq i}} \mathbf{A}^{(1,m_1)}_{ij_1} \odot \mathbf{B}^{(m_1, m_1)^{-1}} \odot  \cdots  \odot \mathbf{A}^{(m_\gamma, 2\text{a})}_{j_\gamma i} : \boldsymbol{\epsilon} ] + O(r^{-11}) \mathbf{J} \;, 
\end{split}
\end{equation}

\begin{equation}\label{Eq:M_TR_ij}
\begin{split}
    \mathbf{M}^{TR}_{ij} = & \frac{1}{12\pi \mu a^2} [ \mathbf{A}^{(1, 2\text{a} )}_{ij} :\boldsymbol{\epsilon} 
    + \sum\limits_{\gamma = 1}^\infty \sum\limits_{m_1 = 2}^\infty{}^{'}  \cdots \sum\limits_{m_\gamma = 2}^\infty{}^{'} \sum\limits_{\substack{j_1, j_2, \dots, j_\gamma \in \{ i, j \}\\j_1\neq i, j_2 \neq j_1, \dots, j_\gamma \neq i}}  \mathbf{A}^{(1,m_1)}_{ij_1} \odot \mathbf{B}^{(m_1, m_1)^{-1}}
    \odot  \cdots  \odot \mathbf{A}^{(m_\gamma, 2\text{a} )}_{j_\gamma j} :\boldsymbol{\epsilon}  \\
    + & \sum\limits_{\substack{k = 1\\k\neq i \text{ or } j }}^N \sum\limits_{\gamma = 1}^\infty \sum\limits_{m_1 = 2}^\infty{}^{'}   \cdots \sum\limits_{m_\gamma = 2}^\infty{}^{'} \sum\limits_{\substack{j_1, j_2, \dots, j_\gamma \in \{ i, j, k \}\\j_1\neq i, j_2 \neq j_1, \dots, j_\gamma \neq j}} \mathbf{A}^{(1,m_1)}_{ij_1} \odot \mathbf{B}^{(m_1, m_1)^{-1}} \odot  \cdots  \odot \mathbf{A}^{(m_\gamma, 2\text{a} )}_{j_\gamma j} :\boldsymbol{\epsilon} ] + O(r^{-8}) \mathbf{J} \;, 
\end{split}
\end{equation}
\end{subequations}
where $\mathbf{J} \in \mathbb{R}^{3 \times 3} $ is the all-ones matrix. The approximation error of Eq.~\eqref{Eq:Mii_Mij} is from the truncated many-body interactions, whose order is determined by the leading four-body interaction terms. For example, the most dominant four-body interaction term in Eq.~\eqref{Eq:M_TT_ij} is $  \mathbf{A}^{(1,2\text{s})}_{ij_1} \odot \mathbf{B}^{(2\text{s}, 2\text{s})^{-1}} \mathbf{A}^{(2\text{s},2\text{s})}_{j_1 j_2} \odot \mathbf{B}^{(2\text{s}, 2\text{s})^{-1}} \mathbf{A}^{(2\text{s},1)}_{j_2 j} $ with the order $O(r^{-7})$, where $r$ is the characteristic distance between particles. 

Noting that each $m$-body interaction depends on the positions $\mathbf{X}$ of associated particles and hence can be denoted as a nonlinear function $\boldsymbol{\alpha}_m(\mathbf{X})$. Thereby, Eq.~\eqref{Eq:Mii_Mij} can be rewritten as:
\begin{subequations}\label{Eq:M_sim}
\begin{equation}\label{Eq:M_TT_sim}
\mathbf{M}^{TT}_{ij} = \left\{
\begin{matrix}
\begin{aligned}
    & \boldsymbol{\alpha}^{TT}_1 + \sum\limits_{\substack{ {l = 1} \\ l\neq i}}^N \left[ \boldsymbol{\alpha}^{TT,(s)}_2(\mathbf{X}_i, \mathbf{X}_l) + \sum\limits_{\substack{k = 1\\k\neq i \text{ or } l }}^N \boldsymbol{\alpha}^{TT,(s)}_3(\mathbf{X}_i, \mathbf{X}_k, \mathbf{X}_l) \right] + O(r^{-10}) \mathbf{J}
\end{aligned}
& \ \ i = j \\
 & \\
\begin{aligned}
    &  \boldsymbol{\alpha}^{TT,(t)}_2(\mathbf{X}_i, \mathbf{X}_j) +  \sum\limits_{\substack{k = 1\\k\neq i \text{ or } j }}^N  \boldsymbol{\alpha}^{TT,(t)}_3(\mathbf{X}_i, \mathbf{X}_k, \mathbf{X}_j) + O(r^{-7}) \mathbf{J}
\end{aligned}
&  \ \  i \neq j
\end{matrix}
 \right. \;,
\end{equation}

\begin{equation}\label{Eq:M_TR_sim}
\mathbf{M}^{TR}_{ij} = \left\{
\begin{matrix}
\begin{aligned}
    & \boldsymbol{\alpha}^{TR}_1 + \sum\limits_{\substack{ {l = 1} \\ l\neq i}}^N \left[ \boldsymbol{\alpha}^{TR,(s)}_2(\mathbf{X}_i, \mathbf{X}_l) + \sum\limits_{\substack{k = 1\\k\neq i \text{ or } l }}^N \boldsymbol{\alpha}^{TR,(s)}_3(\mathbf{X}_i, \mathbf{X}_k, \mathbf{X}_l) \right]  +  O(r^{-11})\mathbf{J}
\end{aligned}
& \ \ i = j \\
 & \\
\begin{aligned}
    &  \boldsymbol{\alpha}^{TR,(t)}_2(\mathbf{X}_i, \mathbf{X}_j) +   \sum\limits_{\substack{k = 1\\k\neq i \text{ or } j }}^N  \boldsymbol{\alpha}^{TR,(t)}_3(\mathbf{X}_i, \mathbf{X}_k, \mathbf{X}_j) +  O(r^{-8}) \mathbf{J}
\end{aligned}
&  \ \  i \neq j
\end{matrix}
 \right. \;.
\end{equation}
\end{subequations}
Here, the superscripts $(s)$ and $(t)$ are used to indicate that the contribution to the mobility is associated with the force exerted on particle $i$ itself or another particle $j$, respectively; and the subscripts $1$, $2$, and $3$ denote the one-body, two-body, and three-body interactions, respectively. While $\boldsymbol{\alpha}^{TT}_1 = (6\pi \mu a)^{-1} \mathbf{1}$ and $\boldsymbol{\alpha}^{TR}_1 = \mathbf{0}$, all $\boldsymbol{\alpha}_2$ and $\boldsymbol{\alpha}_3$ contain infinite series and do not have analytical forms. 

So far, the discussion is based on unbounded domains. However, the forms given in Eq. \eqref{Eq:M_sim} are also applicable to particulate suspensions in periodic or bounded domains, while for different domains, $\boldsymbol{\alpha}_1$, $\boldsymbol{\alpha}_2$, and $\boldsymbol{\alpha}_3$ would be different. In particular, the constant $\boldsymbol{\alpha}_1$ should be generalized to a function $\boldsymbol{\alpha}_1 (\mathbf{X}_i)$ noting that in bounded domains, it can depend on the particle's position. 

From Eq. \eqref{Eq:M_sim} and neglecting higher order terms, the particle's translational velocity can be approximated by:
\begin{equation}\label{Eq:theo_basis}
\begin{split}
    \mathbf{U}_i   \approx & \boldsymbol{\alpha}^{TT}_1 (\mathbf{X}_i) \cdot \mathbf{F}_i + \sum\limits_{\substack{{j = 1} \\ j\neq i}}^N \left[\boldsymbol{\alpha}_2^{TT,(s)} (\mathbf{X}_i, \mathbf{X}_j) + \sum\limits_{\substack{ k=1 \\k\neq i \text{ or } j }}^N \boldsymbol{\alpha}_3^{ TT,(s)} (\mathbf{X}_i, \mathbf{X}_k, \mathbf{X}_j )\right] \cdot \mathbf{F}_i \\
    + & \sum\limits_{\substack{{j = 1} \\ j\neq i}}^N \left[\boldsymbol{\alpha}_2^{TT, (t)} (\mathbf{X}_i, \mathbf{X}_j) + \sum\limits_{\substack{ k=1 \\ k\neq i \text{ or } j }}^N \boldsymbol{\alpha}_3^{TT,(t)} (\mathbf{X}_i, \mathbf{X}_k, \mathbf{X}_j )\right] \cdot \mathbf{F}_j \\ 
    + & \boldsymbol{\alpha}^{TR}_1 (\mathbf{X}_i) \cdot \mathbf{T}_i + \sum\limits_{\substack{{j = 1} \\ j\neq i}}^N \left[\boldsymbol{\alpha}_2^{TR,(s)} (\mathbf{X}_i, \mathbf{X}_j) + \sum\limits_{\substack{ k=1 \\k\neq i \text{ or } j }}^N \boldsymbol{\alpha}_3^{TR,(s)} (\mathbf{X}_i, \mathbf{X}_k, \mathbf{X}_j )\right] \cdot \mathbf{T}_i  \\
    + & \sum\limits_{\substack{{j = 1} \\ j\neq i}}^N \left[\boldsymbol{\alpha}_2^{TR,(t)} (\mathbf{X}_i, \mathbf{X}_j) + \sum\limits_{\substack{ k=1 \\ k\neq i \text{ or } j }}^N \boldsymbol{\alpha}_3^{TR,(t)} (\mathbf{X}_i, \mathbf{X}_k, \mathbf{X}_j )\right] \cdot \mathbf{T}_j \;. 
\end{split}
\end{equation}
It is convenient to combine the force and torque into one vector, i.e. $\mathbf{F}_i = [\mathbf{F}_{i}^T , \mathbf{T}_{i}^T ]^T \in \mathbb{R}^{6 \times 1} $, to give:
\begin{equation}\label{Eq:theo_basis1}
\begin{split}
    \mathbf{U}_i   \approx & \boldsymbol{\alpha}_1 (\mathbf{X}_i) \cdot \mathbf{F}_i + \sum\limits_{\substack{{j = 1} \\ j\neq i}}^N \left[\boldsymbol{\alpha}_2^{(s)} (\mathbf{X}_i, \mathbf{X}_j) + \sum\limits_{\substack{ k=1 \\k\neq i \text{ or } j }}^N \boldsymbol{\alpha}_3^{ (s)} (\mathbf{X}_i, \mathbf{X}_k, \mathbf{X}_j )\right] \cdot \mathbf{F}_i \\
    & +  \sum\limits_{\substack{{j = 1} \\ j\neq i}}^N \left[\boldsymbol{\alpha}_2^{(t)} (\mathbf{X}_i, \mathbf{X}_j) + \sum\limits_{\substack{ k=1 \\ k\neq i \text{ or } j }}^N \boldsymbol{\alpha}_3^{(t)} (\mathbf{X}_i, \mathbf{X}_k, \mathbf{X}_j )\right] \cdot \mathbf{F}_j \;, 
\end{split}
\end{equation}
where $\boldsymbol{\alpha}_2^{(s)} = [\boldsymbol{\alpha}_2^{TT,(s)}, \boldsymbol{\alpha}_2^{TR,(s)} ] \in \mathbb{R}^{3 \times 6} $, same to different subscript (1, 2, or 3) and superscript $(s)$ or $(t)$. Note that in Eq. \eqref{Eq:theo_basis1}, the approximation retains up to three-body contributions in both short-range \textit{and} long-range HIs. 

We further analyze the orders of each contribution in Eq. \eqref{Eq:theo_basis1}. For the two-body interactions, while the most dominant term in $\boldsymbol{\alpha}^{(t)}_2 (\mathbf{X}_i, \mathbf{X}_j)$ is $\mathbf{A}^{(1,1)}_{ij}$ with the order of $O(| \mathbf{r}_{ij} |^{-1})$, the most dominant term in $ \boldsymbol{\alpha}^{(s)}_2 (\mathbf{X}_i, \mathbf{X}_j) $ is $\mathbf{A}^{(1,2\text{s})}_{ij} \odot \mathbf{B}^{(2\text{s}, 2\text{s})^{-1}} \odot \mathbf{A}^{(2\text{s},1)}_{j i}$ with the order of $ O(| \mathbf{r}_{ij} |^{-4}) $. For the three-body interactions, while $ \boldsymbol{\alpha}_3^{(t)} (\mathbf{X}_i, \mathbf{X}_k, \mathbf{X}_j ) $ has the dominant term $\mathbf{A}^{(1,2\text{s})}_{ik} \odot \mathbf{B}^{(2\text{s}, 2\text{s})^{-1}} \odot \mathbf{A}^{(2\text{s},1)}_{kj}$ with the order of $ O(| \mathbf{r}_{ik} |^{-2}| \mathbf{r}_{jk} |^{-2}) $,  $\boldsymbol{\alpha}_3^{(s)} (\mathbf{X}_i, \mathbf{X}_k, \mathbf{X}_j ) $ has the dominant term $\mathbf{A}^{(1,2\text{s})}_{ik} \odot \mathbf{B}^{(2\text{s}, 2\text{s})^{-1}} \odot \mathbf{A}^{(2\text{s},2\text{s})}_{kj} \odot \mathbf{B}^{(2\text{s}, 2\text{s})^{-1}} \odot \mathbf{A}^{(2\text{s},1)}_{ji}$ with the order of $ O(| \mathbf{r}_{ik} |^{-2}| \mathbf{r}_{jk} |^{-3} | \mathbf{r}_{ji} |^{-2}) $. Assuming $r$ the characteristic distance between particles, the orders of $\boldsymbol{\alpha}^{(t)}_2$, $\boldsymbol{\alpha}^{(s)}_2$, $\boldsymbol{\alpha}^{(t)}_3$, and $\boldsymbol{\alpha}^{(s)}_3$ are hence $O(r^{-1}) $, $O(r^{-4}) $, $O(r^{-4}) $, and $O(r^{-7}) $, respectively. Based on this analysis, we can introduce a cutoff distance, $R_\text{cut}$, for the terms with the orders of $O(r^{-4})$ and above and hence approximate the velocity as:
\begin{equation}\label{Eq:theo_basis_trun}
\begin{split}
    \mathbf{U}_i  & \approx \boldsymbol{\alpha}_1 (\mathbf{X}_i) \cdot \mathbf{F}_i  + \sum\limits_{\substack{ j\neq i \\ j \in \mathcal{N}(i) }}^N \left[\boldsymbol{\alpha}_2^{(s)} (\mathbf{X}_i, \mathbf{X}_j) + \sum\limits_{\substack{ k\neq i \text{ or } j \\i,j \in \mathcal{N}(k) }}^N \boldsymbol{\alpha}_3^{(s)} (\mathbf{X}_i, \mathbf{X}_k, \mathbf{X}_j )\right] \cdot \mathbf{F}_i \\
    & + \sum\limits_{\substack{{j = 1} \\ j\neq i}}^N \left[\boldsymbol{\alpha}_2^{(t)} (\mathbf{X}_i, \mathbf{X}_j) + \sum\limits_{\substack{ k\neq i \text{ or } j \\i,j \in \mathcal{N}(k) }}^N \boldsymbol{\alpha}_3^{(t)} (\mathbf{X}_i, \mathbf{X}_k, \mathbf{X}_j )\right] \cdot \mathbf{F}_j \;,
\end{split}
\end{equation}
where $ \mathcal{N}( * )$ denotes the neighbors of particle $ * $ within the cutoff distance $R_\text{cut}$. 

Finally, if torques can be omitted,  Eq.~\eqref{Eq:theo_basis_trun} can be simplified as:
\begin{equation}\label{Eq:theo_basis_trun_final1}
\begin{split}
    \mathbf{U}_i  & \approx \boldsymbol{\alpha}_1 (\mathbf{X}_i) \cdot \mathbf{F}_i + \sum\limits_{\substack{{j = 1} \\ j\neq i}}^N \left[\boldsymbol{\alpha}_2^{(s)} (\mathbf{X}_i, \mathbf{X}_j) \cdot \mathbf{F}_i  + \boldsymbol{\alpha}_2^{(t)} (\mathbf{X}_i, \mathbf{X}_j) \cdot \mathbf{F}_j\right]    + \sum\limits_{\substack{ j, k: j \neq i,k \neq i \\i,j \in \mathcal{N}(k) }}^N  \left[ \boldsymbol{\alpha}_3^{(s)} (\mathbf{X}_i, \mathbf{X}_k, \mathbf{X}_j ) \cdot \mathbf{F}_i +  \boldsymbol{\alpha}_3^{(t)} (\mathbf{X}_i, \mathbf{X}_k, \mathbf{X}_j ) \cdot \mathbf{F}_j \right] \\
    & =  \boldsymbol{\alpha}_1 (\mathbf{X}_i) \cdot \mathbf{F}_i + \sum\limits_{\substack{{j = 1} \\ j\neq i}}^N \boldsymbol{\alpha}_2 (\mathbf{X}_i, \mathbf{X}_j) \cdot  \mathbf{F}_{i,j}   + \sum\limits_{\substack{ j, k: j \neq i,k \neq i \\i,j \in \mathcal{N}(k) }}^N \boldsymbol{\alpha}_3 (\mathbf{X}_i, \mathbf{X}_k, \mathbf{X}_j ) \cdot \mathbf{F}_{i,j} \;,
\end{split}
\end{equation}
where $\boldsymbol{\alpha}_2^{(s)} (\mathbf{X}_i, \mathbf{X}_j) =\mathbf{0}$ for $j \not \in \mathcal{N}(i)$; $\boldsymbol{\alpha}_2 = [ \boldsymbol{\alpha}_2^{(s)},  \boldsymbol{\alpha}_2^{(t)}] \in \mathbb{R}^{3 \times 6} $; $\boldsymbol{\alpha}_3 = [ \boldsymbol{\alpha}_3^{(s)},  \boldsymbol{\alpha}_3^{(t)}] \in \mathbb{R}^{3 \times 6} $; and $\mathbf{F}_{i,j} = [\mathbf{F}_{i}^T , \mathbf{F}_{j}^T ]^T \in \mathbb{R}^{6 \times 1}$. 

\bibliographystyle{elsarticle-num}
\bibliography{ref}
\end{document}